\documentclass{article}

\usepackage[final]{corl_2021} 

\usepackage{caption}
\usepackage{subfig}
\usepackage{wrapfig}

\usepackage{floatrow}
\newfloatcommand{capbtabbox}{table}[][\FBwidth]

\usepackage{algorithmicx}
\usepackage{algorithm}
\usepackage{algpseudocode}

\usepackage{amsmath}
\usepackage{amssymb}
\usepackage{amsfonts}
\usepackage{amsthm}
\usepackage{amsbsy}
\usepackage{mathrsfs}
\usepackage{mathtools}

\usepackage{booktabs}
\usepackage{array}
\usepackage{makecell}

\usepackage{multirow}

\newcommand*\diff{\mathop{}\!\mathrm{d}}

\usepackage{xcolor}
\definecolor{darkblue}{rgb}{0.0,0.0,0.3}
\definecolor{commentBlue}{rgb}{0.0,0.0,0.8}

\definecolor{blau}{RGB}{0, 69, 134}
\definecolor{orangeR}{RGB}{255,140,0}
\definecolor{gruen}{RGB}{0,150,0}
\definecolor{lila}{rgb}{0.49400,0.18400,0.55600}
\definecolor{rot}{rgb}{1,0.12500,0.009800}
\definecolor{hellblau}{RGB}{23, 190, 207}
\definecolor{rosa}{RGB}{207, 23, 190}

\def\generatePlots{0}

\usepackage{tikz}
\usepackage{pgfplots}
\usetikzlibrary{external}
\if\generatePlots1
\fi  

\pgfplotsset{compat=newest}
\usetikzlibrary{plotmarks}
\usepackage{grffile}
\usepgfplotslibrary{statistics}

\pgfplotsset{every axis/.append style={legend style={ font=\footnotesize, mark options={scale=0.5}, inner sep=0.0cm, row sep=-1pt } }}
\pgfplotsset{
	legend image code/.code={
		\draw[mark repeat=2,mark phase=2]
		plot coordinates {
			(0cm,0cm)
			(0.15cm,0cm)        
			(0.3cm,0cm)         
		};%
	}
}
\pgfplotsset{label style={font=\small},
	tick label style={font=\small} }

\usetikzlibrary{shapes, arrows, fit, calc, positioning, matrix}
\usetikzlibrary{decorations.markings}

\tikzset{font=\small}

\renewcommand{\baselinestretch}{0.94}

\title{Learning Models as Functionals of Signed-Distance Fields for Manipulation Planning}

\author{
	Danny Driess$^1$ \hspace{0.3cm} Jung-Su Ha$^1$ \hspace{0.3cm} Marc Toussaint$^1$ \hspace{0.3cm} Russ Tedrake$^2$ \\[0.1cm]
	$^1$TU Berlin \hspace{1cm} $^2$MIT
}

\begin{document}
\maketitle

\vspace{-0.9cm}
\begin{abstract}
	This work proposes an optimization-based manipulation planning framework where the objectives are learned functionals of signed-distance fields that represent objects in the scene.
	Most manipulation planning approaches rely on analytical models and carefully chosen abstractions/state-spaces to be effective.
	A central question is how models can be obtained from data that are not primarily accurate in their predictions, but, more importantly, enable efficient reasoning within a planning framework, while at the same time being closely coupled to perception spaces.
	We show that representing objects as signed-distance fields not only enables to learn and represent a variety of models with higher accuracy compared to point-cloud and occupancy measure representations, but also that SDF-based models are suitable for optimization-based planning. 
	To demonstrate the versatility of our approach, we learn both kinematic 
	and dynamic models to solve tasks that involve hanging mugs on hooks and pushing objects on a table.
	We can unify these quite different tasks within one framework, since SDFs are the common object representation.
	Video: \url{https://youtu.be/ga8Wlkss7co}
\end{abstract}


\vspace{-0.3cm}
\section{Introduction}
\vspace{-0.2cm}
Manipulation planning is challenging for multiple reasons.
On the one hand, planning robot motions to solve a task can be formulated as a decision problem over a high-dimensional, non-convex space, including discrete and continuous aspects.
Especially long-horizon tasks that consist of multiple manipulation steps have the property that motions have to be coordinated globally with the future goal.
This coupling of potentially all variables requires joint reasoning and makes the problem particularly challenging \citep{21-driess-ICRA}.
On the other hand, the problem solving capabilities of a planning framework is inherently dependent on its underlying models.
The field of Task and Motion Planning (TAMP) has made significant progress in solving challenging multi-step, long-horizon tasks \citep{garrett2021integrated}, ranging from ones that involve mainly kinematic models \citep{kaelbling2010hierarchical,srivastava14combined,dantam18ijrr,hartmann2020robust, driess2020deep} to dynamic tasks that require reasoning about forces, friction etc.\ based on more general dynamic equations \citep{mordatch2012discovery,posa2014direct,toussaint2018differentiable,hogan18reactive,doshi2020hybrid,20-toussaint-physicsLGP, 21-driess-IJRR}.
However, most TAMP approaches rely on carefully chosen abstractions and analytically defined models in order to be successful and efficient.
In particular, TAMP often makes simplifying assumptions on the possible geometries of objects it can deal with to define manipulation constraints in the first place.
It is unclear how these models can be grounded from sensor information.

To overcome these issues, a natural idea is to replace the analytic models in TAMP frameworks with learned ones.
Recent advances in deep learning have enabled to learn predictive forward models even in high-dimensional observation spaces like images.
The typical objective for learning a forward model is its predictive accuracy.
However, having an accurate model does not necessarily imply that a planning framework can utilize it efficiently.
While having an accurate forward prediction model might be sufficient for short-horizon tasks, especially for long-horizon tasks, learned models can exhibit too high combinatorics for sampling or non-informative gradients for achieving future goals.

This paper aims to address these challenges by learning models that can be used effectively by a planning framework while at the same time using a general object representation more closely related to senors spaces.
To realize this, we present an optimization-based TAMP framework where the objectives are learned \emph{functionals} of signed-distance fields (SDFs).
The SDFs represent each object in the scene separately, while the functionals defined on top of them induce constraints on possible, physically plausible interactions between the objects within a trajectory optimization problem.
The task planning aspect is realized by (discrete) decisions that determine which of those functionals are active at which phase of the planning horizon.

We argue that representing objects as SDFs has multiple advantages.
First, an SDF can be seen as an intermediate representation between raw perception like point-clouds or images and full state information.
While not the focus of this work, many methods have been developed to learn and obtain SDFs from, e.g., image observations of the scene.
Further, SDFs can represent arbitrary, non-convex geometries, which is beneficial, since manipulation problems and physical phenomena often depend on the geometry of the interacting objects.
Finally, we show that SDFs are particularly suited for learning and representing models that can later be used within a planning framework effectively.
Since our models are functionals of the SDFs, the constraints can take the information about whole objects into account to reason about their geometry and therefore especially the \emph{interaction} between objects.
Compared to a representation that only describes the surface of an object like point clouds or occupancy measures, a signed-distance field also provides information about the object at distance.
As we experimentally show, this not only leads to models that perform better in their prediction accuracy compared to models learned on top of point-cloud or occupancy object representations, but SDFs also enable the functionals/learned models to have more useful gradients for planning.

In the experiments, we demonstrate the versatility of our approach by tackling two completely different tasks within one framework:
On the one hand, a kinematic task where the goal is to hang mugs of different shapes on hooks of different shapes.
On the other hand, a pushing scenario where boxes and L-shaped objects should be pushed to different goal regions by pushers of different sizes.
In the first case, the model predicts whether the static interaction between SDFs leads to manipulation success,
whereas in the latter case, the model predicts the forward dynamics in SDF space based on a history of SDF interactions of two objects.
We show that our framework can be used to plan motions that involve multiple push phases.
To summarize our main contributions, we propose
\begin{itemize}
    \vspace{-0.2cm}
	\item To learn a novel class of kinematic and dynamic models as functionals of SDFs,
	\item A manipulation planning framework that utilizes these learned functionals as constraints,
	\item Comparison to other object representations showing the advantages of the SDFs.
\end{itemize}

\section{Related Work}\label{sec:relatedWork}
\vspace{-0.2cm}
\subsection{Signed Distance Fields as Object Representation}
\vspace{-0.2cm}
Representing objects or scenes as implicit surfaces \citep{chen2019learning, liu2019learning, OccupancyNetworks} or SDFs \citep{Park_2019_CVPR, sitzmann2019metasdf, Atzmon_2020_CVPR, jiang2020local, SDFCollision, Fuhrmann2003DistanceFF} is an active research topic, due to aforementioned advantages like learning shape completion, non-convex shapes etc.
Our focus is not to obtain SDFs from observations in the first place.
Conversely, we are interested in what can be done with SDFs in the context of model learning and manipulation planning.
There are some works that utilize SDFs within trajectory optimization \citep{hauser1, zhang2021semi,pfrommer2020contactnets}, but without learning or integration into a TAMP framework.
While some recent approaches \citep{breyer2020volumetric,jiang2021synergies,van2020learning} have suggested that grasping of diverse objects can be addressed using implicit functions, we present a manipulation framework that utilizes SDFs for learning and formulating more general models.

\subsection{Perceptual Models}
\vspace{-0.2cm}
There is great interest in learning predictive models in perception spaces, especially applied to the problem of pushing.
So-called visual foresight approaches \citep{visualforesight, kandukuri2021learning, xu2019densephysnet} aim to predict the evolution of the scene in image space.
Our SDF dynamics model is also closely related to perception spaces, but, in comparison, is naturally differentiable.
\citet{xu2020learning} use a voxelized SDF-based representation of the whole scene to predict the motion of an object when an action is applied.
Our approach is more structured in the sense that we do not predict the scene flow for actions applied on a single object, but the dynamics of interacting of objects.
In \citep{manuelli2020keypoints}, the pushing dynamics in keypoints extracted from visual object observations is learned. However, their focus is to utilize the learned model to stabilize a trajectory with MPC. We focus on planning a complex pushing trajectory and not stabilization during execution.
SE3 networks \citep{byravan2017se3} learn a forward model that predicts a rigid transformation of an observed point cloud given actions. However, they need ground-truth transformations at training time (we only need SDF observations).
Where most of these approaches differ from our approach is that they assume the model to be a function of the observation of a single object or the scene and an action as input.
Therefore, these approaches are mostly limited to the same pusher geometry and make the assumption that actions can readily be applied to the object. Our model handles the interaction between objects of different shapes and can plan the contact establishment phase as well.
Transporter networks \citep{zeng2020transporter} or deep visual reasoning \citep{driess2020deep} predict manipulation sequences from image spaces. However, no dynamic models are considered in these approaches.

\subsection{Manipulation Planning (with learned models)}
\vspace{-0.2cm}
In \citep{simeonov2020learning, mukherjee2020sim}, a manipulation framework based on point cloud observations and manipulation primitives is proposed. Our method plans the complete motions based on learned dynamic models.
\citet{sutanto2020learning} is related to our formulation in the sense that they learn manifolds that are used as constraints in sequential manipulation problems. However, there are no dynamic models or dependencies on the geometry of the involved objects in the learned constraints.
\citet{you2021omnihang} address a hanging task similar to our mug hanging experiment on a more diverse set of object categories. They use a point-cloud-based input representation to predict a hanging pose. Therefore, they need a special neural network for collision avoidance (similar to \citep{danielczuk2020object}), while our SDF based representation can handle collisions directly.
Further, we learn a manifold of solutions instead of predicting a single hanging configuration.
To summarize, what makes our approach unique is that we propose to use SDFs as a common object representation that is closely connected to perception to learn a variety of models that are able to take the interaction of objects into account and can be integrated in an optimization-based motion planning framework due to their differentiability.

\section{Background on Signed-Distance Fields (SDFs)}\label{sec:notation}
\vspace{-0.2cm}
Let $\Omega \subset \mathbb{R}^3$ be an object in the 3D Euclidean space.
A function $\phi : \mathbb{R}^3 \rightarrow \mathbb{R}$, $\phi\in\Phi$ with $\phi(x) = -d(x, \partial \Omega)$ for $x\in\Omega$ and $\phi(x) = d(x, \partial \Omega)$ for $x\in \mathbb{R}^3\backslash \Omega$
is called a signed-distance field of $\Omega$ in $\mathbb{R}^3$.
Here, $d(x, \partial \Omega) = \inf_{x^\prime\in\partial \Omega} \left\|x-x^\prime\right\|_2$ and $\partial \Omega$ the boundary of $\Omega$.
We assume $\phi$ to be differentiable almost everywhere in $\mathbb{R}^3$.
The way $\phi$ is defined ensures that inside the object, $\phi$ attains negative values, on the boundary zeros, and outside positive ones. We denote with the set $\Phi$ the space of all functions $\phi$ that are SDFs for some object.
%
\vspace{-0.3cm}
\paragraph{Rigid Transformations of SDFs}
A central concept in this work is to rigidly transform SDFs in space.
This can be realized by transforming the input where the SDF is queried.
To simplify the notation, we define a \emph{rigid transformation}, parameterized by $q\in\mathbb{R}^7$ (translation + quaternion),
\begin{align}
T(q)[\phi](\cdot) := \phi\left(R(q)^T\left(~\cdot~ - r(q)\right)\right) \label{eq:rigidTransformation}
\end{align}
of an SDF $\phi$, where $R(q)\in\mathbb{R}^{3\times 3}$ is a rotation matrix and $r(q)\in\mathbb{R}^3$ the translation vector.

\section{Manipulation Planning with Signed-Distance Functionals}\label{sec:planning}
\vspace{-0.2cm}
The core idea of this work is to represent each object $i$ in the scene as a signed-distance field $\phi^i$ in order to learn predictive models as functionals $H$ of these SDFs.
Based on the learned functionals, we formulate a trajectory optimization problem where the decision variable is a trajectory of rigid transformations applied on the initial SDFs as they have been observed in the initial scene.

More specifically, through interaction with the environment, we aim to learn functionals of the form $H : \Phi \times \cdots \times \Phi \rightarrow \mathbb{R}$
that map multiple SDFs of multiple, possible different objects at possibly different consecutive times to a real number.
These are trained in a way that a value of zero implies that the SDFs as input are compatible with what has been learned through interaction with the environment. Otherwise, they should attain a positive value, hence functionals $H$ discriminate correct from incorrect dynamics or desired from undesired manipulations.

The learned functionals then define constraints for the (hybrid) trajectory optimization problem
\begin{subequations}\label{eq:opt}
	\begin{align}
	\min_{\substack{q_{0:KT}, q_t\in\mathbb{R}^{7 \cdot n_O} \\ K\in\mathbb{N}, ~s_{1:K} }} \sum_{t=1}^{KT} c\big(q_{t-l:t}, s_{k(t)}\big) ~~~~~~~~~~~~~~~~~~~~~~~~~~~~~~~~~~~~~~~&\\
	\text{s.t.}~~~~\forall_{H \in \mathbb{H}(s_{k(t)})} ~:~ H\left(\big(T(q^i_t)[\phi^i]\big)_{(t,i)\in \mathcal{I}_H(s_{k(t)})}\right) &= 0\\
	s_{1:K} \in \mathbb{S}(S), ~~q_0 &= 0.
	\end{align}
\end{subequations}
The discrete variable $s_k$ determines which functionals $H$ from the set $\mathbb{H}(s_{k(t)})$ are active at which of the $K\in\mathbb{N}$ phases of the motion ($k(t) = \lfloor t/T \rfloor$).
This number of phases is part of the decision problem.
The trajectory $q_{0:KT}$ of rigid transformations is discretized in time into $T\in\mathbb{N}$ steps per phase.
If $n_O$ is the number of objects in the scene $S$, then $q_t\in\mathbb{R}^{7\cdot n_O}$, leading to $7\cdot KTn_O$ continuous variables.
Further, $s_k$ selects through the set $\mathcal{I}_H(s_{k(t)})$ the time slice and object index tuples $(t,i)$ that determine the SDFs $\phi^i$, which have been transformed through $q_t^i$, at the times $t$ of the trajectory on which the functional constraints $H$ depends on.
This problem formulation is inspired by LGP \citep{toussaint2018differentiable}, but the constraints are replaced by learned functionals of SDFs.
The set $\mathbb{S}(S)$ contains all valid sequences $s_{1:K}$ of such discrete variables for the scene $S$.
The goal of the manipulation problem is specified through $\mathbb{S}(S)$ by $s_K$ selecting a desired goal functional constraint that has to be fulfilled at the end $q_{KT}$ of the trajectory.
Solving \eqref{eq:opt} therefore involves a tree search over nodes $s_{1:K}$ such that the continuous optimization problem implied by the choice of $s_{1:K}$ at a node of the tree is feasible.
The role of $q$ in the optimization problem is not absolute object poses, but rather rigid transformations applied to the SDFs $\phi^i$ that represent the configurations of the objects as observed in the scene initially. 
With the term $c$, we can include regularizing motion costs.
As will be described in sec.~\ref{sec:DSDFuncs:dynamic}, the forward dynamic model we learn for pushing implies a constraint on the evolution of one object based on the motion of another object.
Therefore, we only add motion costs to those degrees of freedom that can be interpreted as being controlled, meaning the motion of the other object. From the perspective of \eqref{eq:opt}, there is no explicit notion of controlled actions.
%

\section{Deep Signed-Distance Functionals}\label{sec:DSDFuncs}
\vspace{-0.2cm}
This section presents two main types of models we propose.
First, a way of learning forward dynamic models that predict the dynamics in SDF space based on the interaction between objects.
Second, a kinematic success model that determines whether a static configuration of interacting SDFs leads to manipulation success.
All functionals we consider are of the form $H : \Phi \times \cdots \times \Phi \rightarrow \mathbb{R}$, i.e.\ they only take the SDFs of interacting objects as input, there is no explicit notion of position, orientation, action etc.
Therefore, the functionals can be used at arbitrary locations in space.
\vspace{-0.3cm}
\paragraph{Bounding-Box}
To define most of the following functionals and those in sec.~\ref{sec:TaskFuncs}, we utilize a set $\mathcal{X}$ with the property $\Omega \subset \mathcal{X} \subset \mathbb{R}^3$ for all objects $\Omega$ that are involved.
This set should be large enough to cover the relevant workspace of the manipulation problem where the interaction between the objects should occur.
A more detailed discussion about the role of $\mathcal{X}$ can be found in sec.~\ref{sec:DSDFuncs:neuralNet}.

\subsection{Forward Dynamic Models}\label{sec:DSDFuncs:dynamic}
\vspace{-0.2cm}
Generally, a forward model predicts future states/observations of a system given the current or additionally a history of states/observations.
In the context of objects being represented solely as SDFs, we propose to learn a forward model $F : \Phi \times \cdots \times \Phi \rightarrow \Phi$ that predicts the SDF of an object $\phi^1_t$ at time step $t$ based on a history of SDF observations of the object $\phi^1_{t-l:t-1}$ until time $t-1$ and the motion of another object $\phi^2_{t-l:t}$ until time $t$.
This means $F$ as
\begin{align}
\phi^1_t(\cdot) = F\big[\phi^1_{t-l:t-1}, \phi^2_{t-l:t}\big](\cdot)\label{eq:FDirect}
\end{align}
is an SDF itself that can be queried in $\mathbb{R}^3$.
Interactions between more than two objects are possible, but we focus on pair-interactions in the present work.
If $l=1$, $F$ is a quasi-static model.
Internally, $F$ can be defined to either directly predict the SDF $\phi^1_t$ as in \eqref{eq:FDirect} or the flow
\begin{align}
\phi^1_t(\cdot) = \phi^1_{t-1}(\cdot) + F_\text{flow}\big[\phi^1_{t-l:t-1}, \phi^2_{t-l:t}\big](\cdot)\label{eq:FFlow}
\end{align}
from $\phi^1_{t-1}$ to $\phi^1_t$.
In both cases, the functional $H$ for planning is then naturally defined as
\begin{align}
H_F\big(\phi^1_{t-l:t}, \phi^2_{t-l:t}\big) = \int_\mathcal{X} \Big(\phi^1_{t}(x) - F\big[\phi^1_{t-l:t-1}, \phi^2_{t-l:t}\big](x)\Big)^2 \diff x. \label{eq:FFunctional}
\end{align}
For a perfect model $F$, this functional $H_F$ attains a zero value if and only if the evolution of $\phi^1_{t-l:t}$ and $\phi^2_{t-l:t}$ is compatible with the underlying physical process in the space $\mathcal{X}$.
Therefore, the loss function to train $F$ is also \eqref{eq:FFunctional} for a dataset $D = \left\{\left(\phi^1_{0:l}, \phi^2_{0:l}\right)_i\right\}_{i=1}^n$ of such consecutive SDF motions of the two objects.
Since $F$ takes as input the complete SDFs of the objects and not just values like the distance between objects and their contact point locations, it can learn to reason not only about these quantities, but also the contact geometry, relative object movements, center of mass and inertial parameters (assuming an equal density of the objects), all of which are necessary quantities to represent the dynamics.
This way, $F$ inherently takes the geometry of the objects into account.
Note that usually, forward models are understood in terms of a function that maps the current state (history) and an (abstract) action $a$ to the next state.
For SDFs, this would mean a model of the form $\phi^1_t = F[\phi^1_{t-l:t-1}, a_{t-1}]$.
In our case, however, there is no notion of an abstract action, instead, our formulation learns a generic model of the interaction between two objects, where the motion of one object ($\phi^2$) influences the other ($\phi^1$).
Therefore, while the transformation applied to $\phi^2$ can be interpreted as an action, the model has no action as input and hence can deal with different geometries of $\phi^2$, which is not possible in case of an abstract action without $\phi^2$ also being an input.

\subsection{Kinematic Success Models}\label{sec:DSDFuncs:kinematic}
\vspace{-0.2cm}
Many tasks in manipulation planning can be specified in terms of static success models instead of a full forward dynamics model.
We call a model that predicts whether a configuration of potentially multiple SDFs at the same time slice leads to manipulation success a kinematic success model.
%
%
Assume through interaction with the environment, a dataset $D = \left\{(\phi^j)_{j\in\mathcal{I}_i}, y^i\right\}_{i=1}^n$ of SDFs representing $|\mathcal{I}_i|$ many objects has been obtained with $y^i = 1$ indicating that the configuration of SDFs leads to manipulation success, $y^i = 0$ to failure. 
Then learning $H$ is similar to a classification problem, where $H\left((\phi^i)_{i\in\mathcal{I}}\right) = 0$ implies success prediction. 
This way, $H$ can model a manifold of feasible configurations and not only a single solution.
See sec.~\ref{sec:appendix:lossKinematicSuccess} for details (loss function etc.).

\subsection{Learning Functionals with Neural Networks}\label{sec:DSDFuncs:neuralNet}
\vspace{-0.2cm}
So far, we have not discussed how functionals of the form $H:\Phi\times \cdots \times \Phi \rightarrow\mathbb{R}$ can be learned or even queried in the first place with usual function approximators like neural networks, since, in general, the neural network would have to take functions as infinite dimensional objects as input.
To approximate this, we choose in this work the straight-forward approach by evaluating $\phi\in\Phi$ on a discretized version of the set $\mathcal{X}$, denoted by $\mathcal{X}_h$.
As discussed previously, the set $\mathcal{X}$ should cover the relevant region of the workspace where the interaction between the objects takes place.
We specifically do not assume $\mathcal{X}$ to be aligned or perfectly centered with the objects that are involved.
%
This way, the dynamics model from sec.~\ref{sec:DSDFuncs:dynamic} can be realized by
\begin{align}
F\big[\phi^1_{t-l:t-1}, \phi^2_{t-l:t}\big](x) \approx F_\theta(\phi^1_{t-l:t-1}(\mathcal{X}_h), \phi^2_{t-l:t}(\mathcal{X}_h), x)
\end{align} 
with $F_\theta$ being usual neural network architectures.
During training, the integral in \eqref{eq:FFunctional} is approximated over the same discretized $\mathcal{X}_h$ for simplicity.
Hence, the dataset to train $F$ can contain the SDF observations at the grid points of $\mathcal{X}_h$ only.
However, $F_\theta$ still approximates an SDF which can be queried at arbitrary $x\in\mathbb{R}^3$ and does not only predict the values on the grid points.
%
For general functionals $H$, the evaluation is analogous, i.e.\ $H\!\left(\left(\phi^i\right)_{i\in\mathcal{I}}\right) \approx H_\theta\left(\left(\phi^i(\mathcal{X}_h)\right)_{i\in\mathcal{I}}\right)$.
Technically, $\mathcal{X}_h\subset \mathbb{R}^{d\times h\times w}$ is a regular grid which allows us to encode $\phi(\mathcal{X}_h)$ using 2D or 3D convolutions.
In contrast to an occupancy grid, the evaluation of $\phi(\mathcal{X}_h)$ contains more information about the object than whether there is an object at the grid point or not.
Note that the differentiabilty of $H\!\left(\left(T(q^i)[\phi^i(\mathcal{X}_h)]\right)_{i\in \mathcal{I}}\right)$ with respect to $q^i$ is maintained, which is another advantage of representing such models as functionals of SDF functions evaluated on a grid instead of static values on a grid.
During training, it is sufficient to only have the SDF values evaluated on a gird, no other information like actions or velocity/pose estimations are needed.

\section{Task Constraint Functionals}\label{sec:TaskFuncs}
\vspace{-0.2cm}
Here we present analytical functionals of SDFs that are useful to specify goals of a manipulation problem or other task aspects.
These functionals are general as a direct consequence of our object representations being SDFs. Therefore, there is no advantage or need to learn these given the SDFs.

\subsection{Pair-Collision between Objects}\label{sec:TaskFuncs:collision}
\vspace{-0.2cm}
Collision avoidance is an inherent part of many task specifications.
Given two SDFs $\phi_1, \phi_2$, we can measure whether they are in collision via their overlap integral
\begin{align}
H_\text{coll}(\phi_1, \phi_2) = \int_\mathcal{X} \left[\phi_1(x) < 0\right]\left[\phi_2(x) < 0\right] \diff x. \label{eq:pairCollision}
\end{align}
The indicator bracket $[\cdot]$ means $[P] = 1$ if $P$ is true, otherwise $[P] = 0$.
The integral in \eqref{eq:pairCollision} integrates over the space where both SDFs are negative at the same time, which is only the case if the two objects overlap, hence are in collision.
The gradients of \eqref{eq:pairCollision} are smoothed using the sigmoid function $\sigma(z) = \frac{1}{1 + \exp(-z)}$, i.e.\ $\left[\phi_{1,2}(x) < 0\right] = \sigma\left(-a\phi_{1,2}(x)\right)$ with a parameter $a > 0$.

\subsection{Goal Region}\label{sec:TaskFuncs:goalRegion}
\vspace{-0.2cm}
If part of the task specification is that an object $\phi_1$ is fully contained inside the boundary of another object $\phi_g$, called the goal region, then a similar integral as for the pair-collision can be utilized
\begin{align}
H_\text{g}(\phi_1, \phi_g) = \int_\mathcal{X} \left[\phi_1(x) < 0\right]\left[\phi_g(x) > 0\right] \diff x \approx \int_\mathcal{X} \sigma\left(-a\phi_1(x)\right)\sigma\left(a\phi_g(x)\right) \diff x.\label{eq:goalRegion}
\end{align}
Here, points outside of the goal region that are inside the object count towards the integral.

\subsection{Establishing Contact between Objects}\label{sec:TaskFuncs:PoC}
\vspace{-0.2cm}
Establishing and maintaining contact between objects is central for many manipulation tasks.
One way to model that the distance between two objects $\phi_1$ and $\phi_2$ should be zero is via the functional
\begin{align}
H_\text{PoC}(\phi_1, \phi_2) = \min_{p\in\mathcal{X}} |\phi_1(p)| + |\phi_2(p)|. \label{eq:PoCOpt}
\end{align}
%

\section{Experiments}\label{sec:exp}
\vspace{-0.2cm}
\subsection{Mug-Hanging: Kinematic Success Model}\label{sec:exp:mugHanging}
\vspace{-0.2cm}
In this experiment, we want to find rigid transformations applied on observed mugs of different shapes in a scene to hang them stably on hooks of different types.
The functional $H_\text{hang}$ is therefore a kinematic success model that takes the SDFs of the mug and the hook as input.
To generate data to learn $H_\text{hang}$, we randomly sample scenes of different mug and hook shapes (1600 scenes for training, 400 for testing and 150 for evaluation).
See Fig.~\ref{fig:mugs} for examples of mugs and hooks in the evaluation data.
Then we sample for each scene in the training and test data the position and orientation of the mug uniformly in the bounding box $\mathcal{X}$ until at least one successful configuration has been obtained where the mug does not fall on the ground when being dropped from the sampled configuration while at the same time not being in collision with the hook.
We use Bullet \citep{coumans2021} to simulate the dropping. 
In total, 20 configurations per scene are generated.
Since sampling a successful configuration is a rare event, for the majority of the scenes, only one successful and 19 failure configurations are contained in the training and test data, making learning challenging.
Another challenge of this task is that the model has to reason about both the hook and mug geometry jointly. 
Formulating an analytical model, e.g.\ on a mesh-based object representation, to model this constraint is non-trivial.

\subsubsection{Performance with Optimization}
\begin{figure}
	\TopFloatBoxes
\begin{floatrow}
\ffigbox[5.2cm]{%
    \centering
    \includegraphics[trim={6cm 8cm 6cm 3cm}, clip, height=1.6cm]{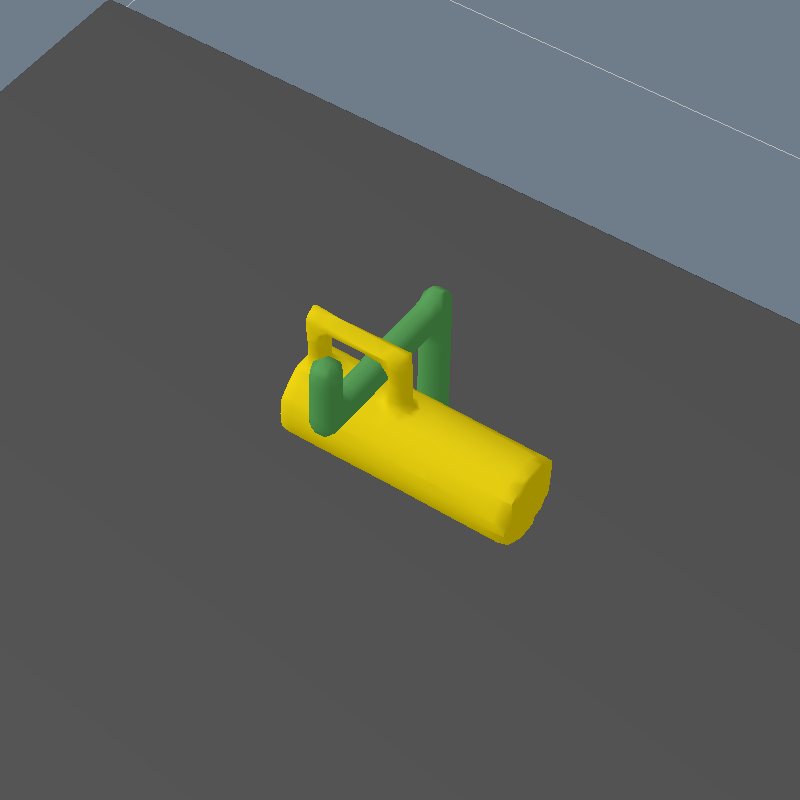}
    \includegraphics[trim={6cm 8cm 6cm 3cm}, clip, height=1.6cm]{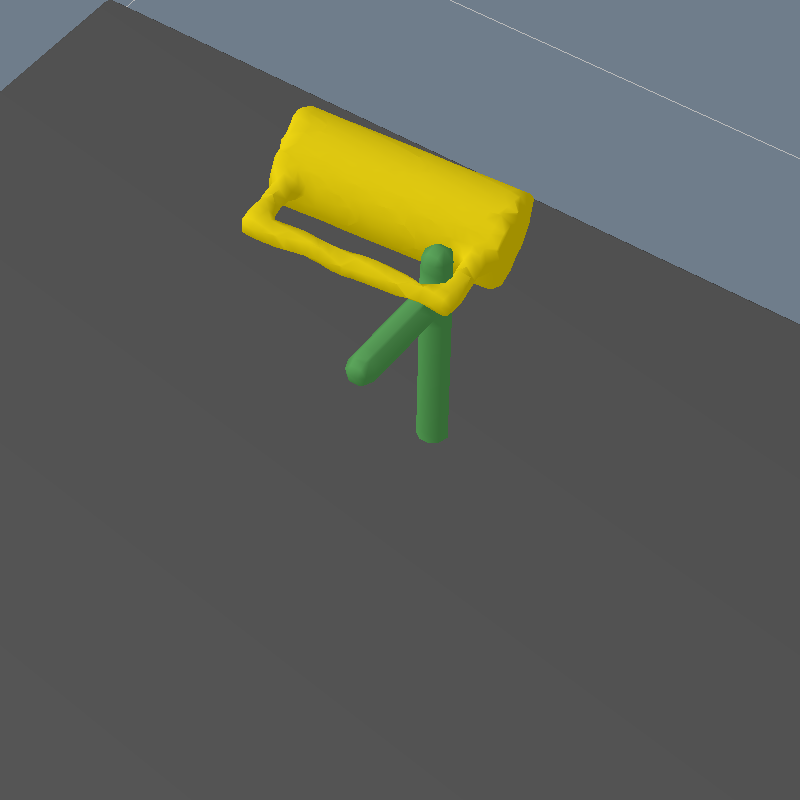}
    \includegraphics[trim={6cm 8cm 6cm 3cm}, clip, height=1.6cm]{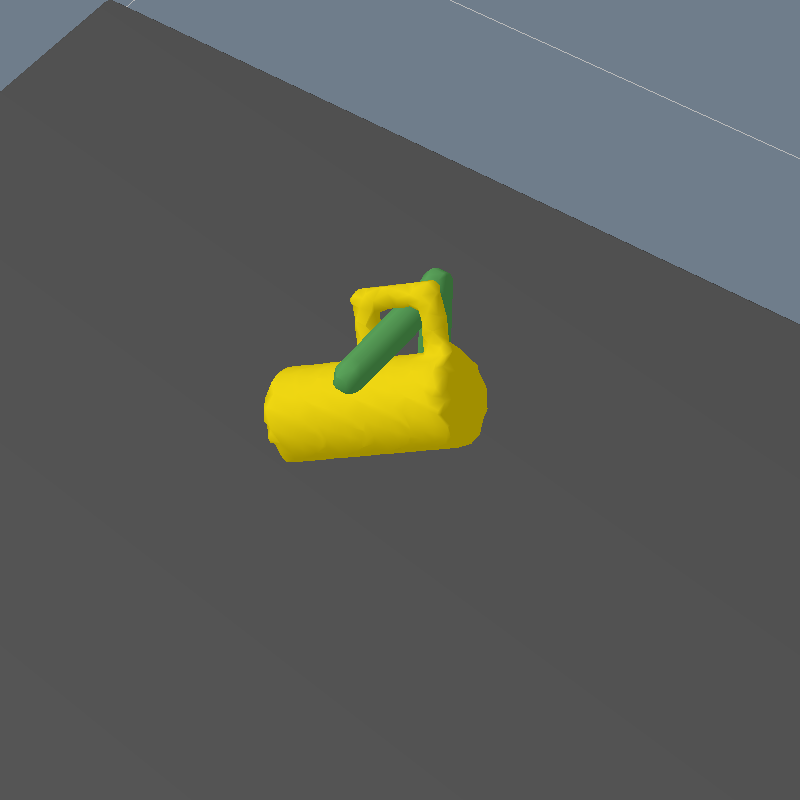}\\
    \includegraphics[trim={6cm 8cm 6cm 3cm}, clip, height=1.6cm]{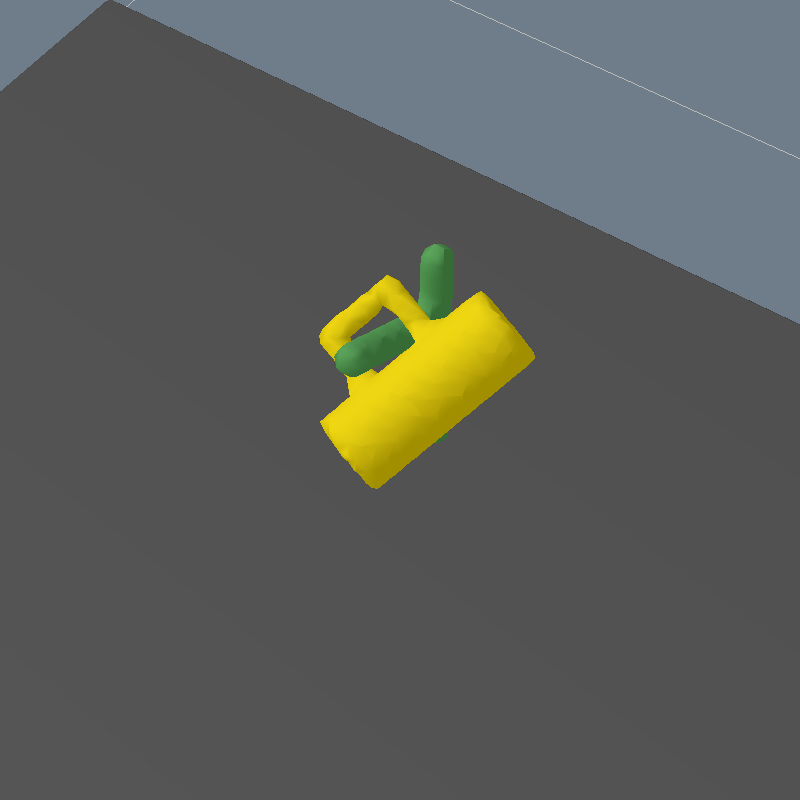}
    \includegraphics[trim={6cm 8cm 6cm 3cm}, clip, height=1.6cm]{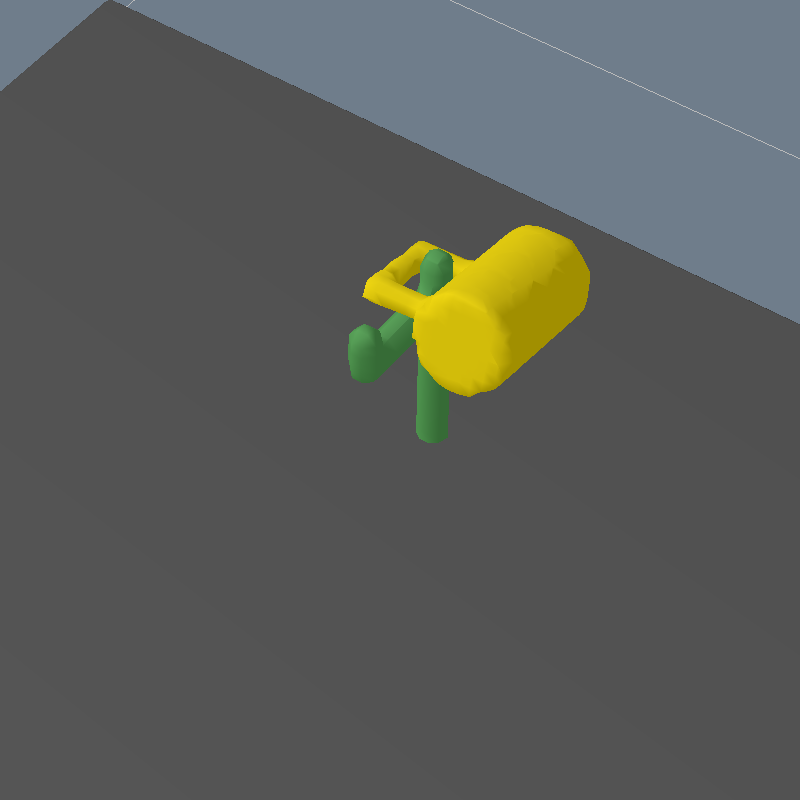}
    \includegraphics[trim={6cm 8cm 6cm 3cm}, clip, height=1.6cm]{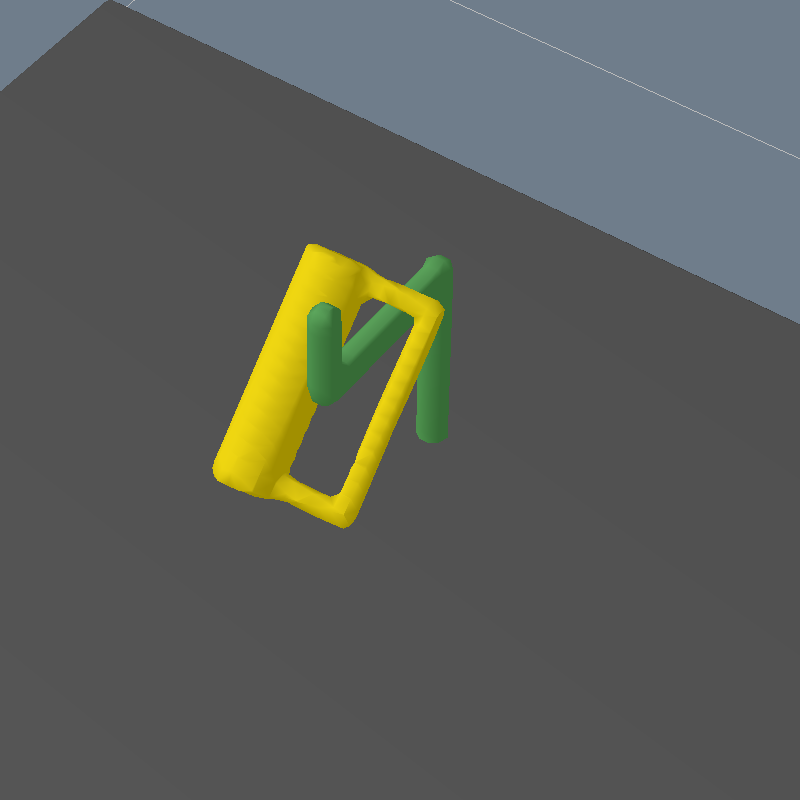}
}{%
\caption{\small Found solution configurations by the optimizer using the learned model.}
\label{fig:mugsSolutions}
}
\capbtabbox{%
    \centering
	\small
	\setlength{\tabcolsep}{3pt}
	\begin{tabular}{cccc}
		\hline
		& solution & success & \textbf{total scenes} \\
		& found & rate & \textbf{solved} \\
		\hline
		\textbf{SDF opt. + sampling} & 98.7\% & 88.5\% & \textbf{87.3\%} \\
		SDF opt. only & 51.3\% & 93.5\% & 47.3\% \\
		SDF sampling only $\kappa_1$ & 83.8\% & 82.3\% & 68.9\% \\
		\hline
		occupancy $\psi$, sampling $\kappa_3$ & 100.0\% & 34.0\% & 34.0\%\\
		\hline
		pointnet++, sampling $\kappa_1$ & 100.0\% & 82.0\% & 52.0\%\\
		\hline
	\end{tabular}
}{%
\vspace{-0.15cm}
\caption{\small Success rates of mug hanging experiment. Total scenes solved means the percentage of scenes in the evaluation dataset for which a solution was found that is not in collision and is stable when dropped. Only best results for each object representation shown. Full results see Tab.~\ref{tab:mugsComparison} and sec.~\ref{sec:appending:comparisonObjectRepresentations:mugs}.}
\label{tab:mugs}
}
\vspace{-0.5cm}
\end{floatrow}
\end{figure}

\vspace{-0.2cm}
Fig.~\ref{fig:mugsSolutions} shows solution configurations found by our model $H_\text{hang}$ as an optimization objective.
Interestingly, the solutions not always contain the intuitive solution, but also ones where other parts of the hook are being utilized (middle column in Fig.~\ref{fig:mugsSolutions}).
The optimization problem \eqref{eq:opt} to solve this mug hanging problem has two objectives, the learned kinematic success functional $H_\text{hang}$ and the pair-collision $H_\text{coll}$ from sec.~\ref{sec:TaskFuncs:collision}.
While in principle $H_\text{hang}$ also learns to avoid collisions, we found that the robustness in avoiding collisions increases when including $H_\text{coll}$.
The learned functional $H_\text{hang}$ is, in general, non-convex in the rigid transformation $q$ of the mug.
Therefore, we observed that using gradient based optimization is not sufficient for the optimizer to find a feasible solution, i.e.\ where $H_\text{hang}$ predicts zero,
in every instance.
To overcome this issue, we restart the optimization procedure up to 20 times with a randomly sampled initial guess of the mug in $\mathcal{X}$.
Fig.~\ref{fig:mugsEvolution} shows an example of a sampled initial configuration from which the optimizer is started (left), then the optimized configuration (middle) and finally, the configuration after simulation.
Tab.~\ref{tab:mugs} shows the success rates on the evaluation scenes.
As one can see, for the proposed approach where objects are represented as SDFs using optimization and sampling, in 98.7\% of the evaluation scenes, a solution is found where $H_\text{hang}$ predicts success and no collision is violated (first column).
Out of these, 88.5\% are stable configurations (checked by simulation) and the optimized configuration of the mug is collision free with the hook, leading to 87.3\% total solved scenes (last column).
When the optimization is run only once (second row), then only in 51.3\% of the cases it converges to a feasible solution.

\subsubsection{Comparison to Sampling, Point-Cloud and Occupancy Measure Representations}
\vspace{-0.2cm}
This section shows that learning a kinematic success model based on the proposed SDF object representation outperforms other representations (point-clouds and occupancy measures) and further highlights the advantage of the models learned with SDFs providing useful gradients by comparing it to sampling without optimization.
For full results, refer to sec.~\ref{sec:appendix:comparisonObjectRepresentations} and Tab.~\ref{tab:mugsComparison}.
The sampling approach draws relative transformations of the mugs uniformly in $\mathcal{X}$ until the evaluation with the learned $H_\text{hang}$ and the collision functional $H_\text{coll}$ predicts a successful and collision free configuration with a threshold $\kappa$.
As one can see in Tab.~\ref{tab:mugs} and Tab.~\ref{tab:mugsComparison}, our proposed approach has a significantly higher performance (87.3\%) compared to the best threshold for pure sampling with SDF (68.9\%) and the best of the other object representations (34\% with occupancy measure, 52\% with point-cloud).
%
%

\subsection{Pushing Objects on a Table: Dynamic Model}\label{sec:exp:pushing}
\begin{figure}
	\centering
	\includegraphics[trim={1cm 1cm 2cm 2cm}, clip, width=1.6cm]{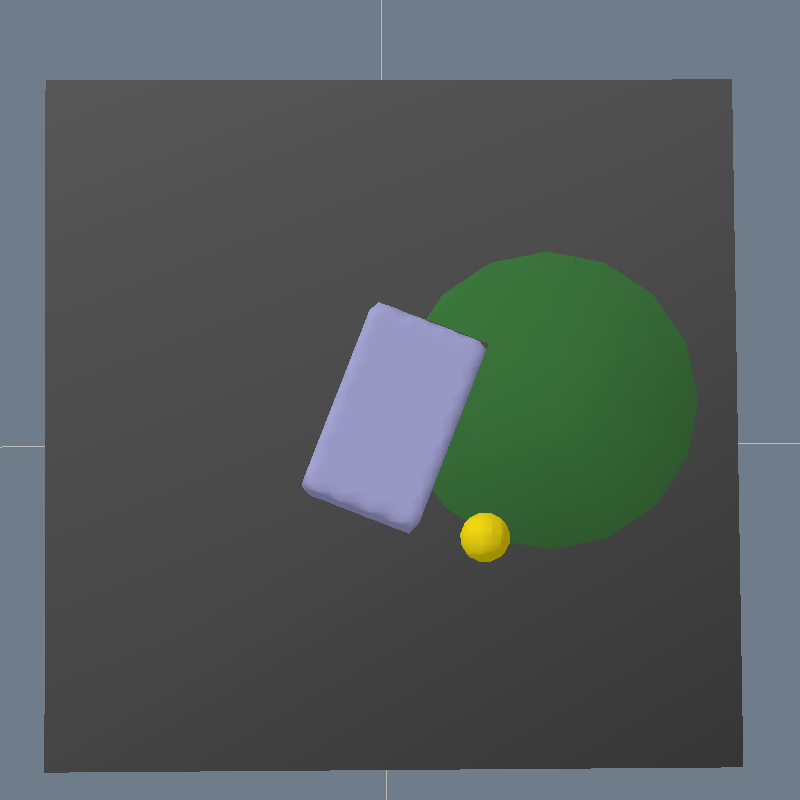}
	\includegraphics[trim={1cm 1cm 2cm 2cm}, clip, width=1.6cm]{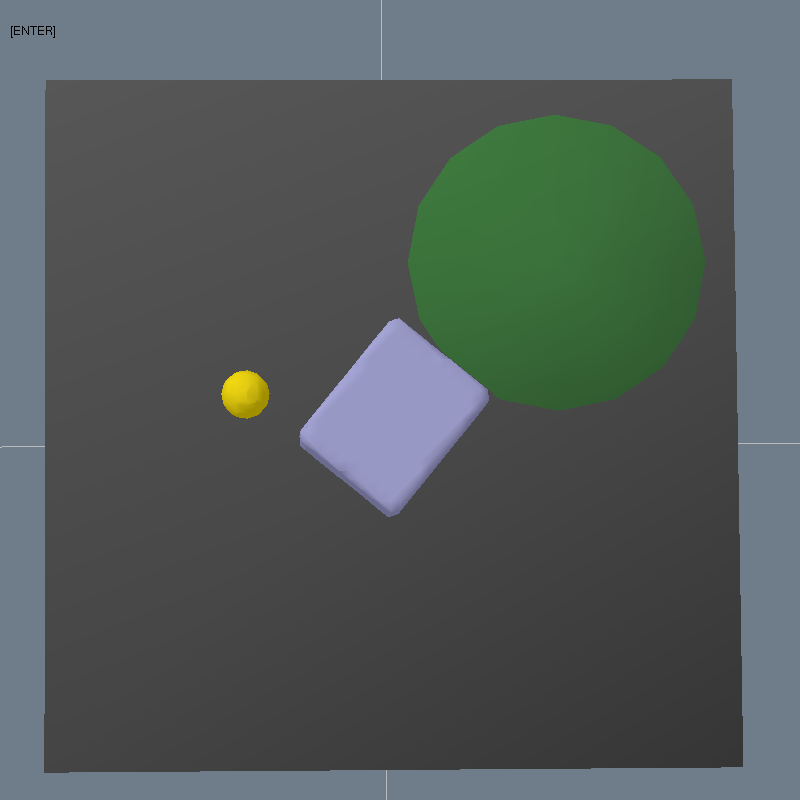}
	\includegraphics[trim={1cm 1cm 2cm 2cm}, clip, width=1.6cm]{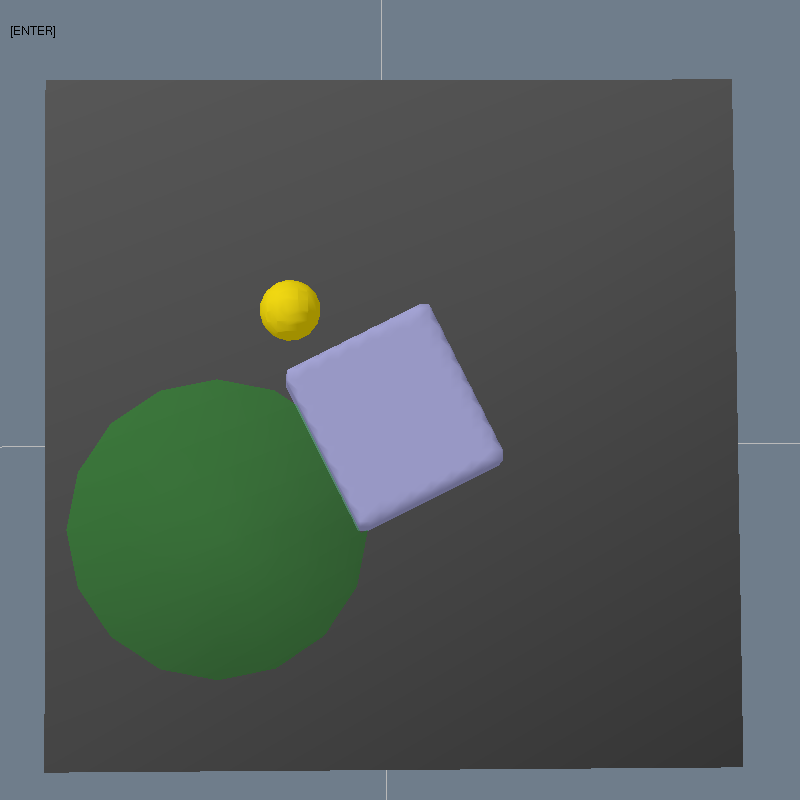}
	\includegraphics[trim={1cm 1cm 2cm 2cm}, clip, width=1.6cm]{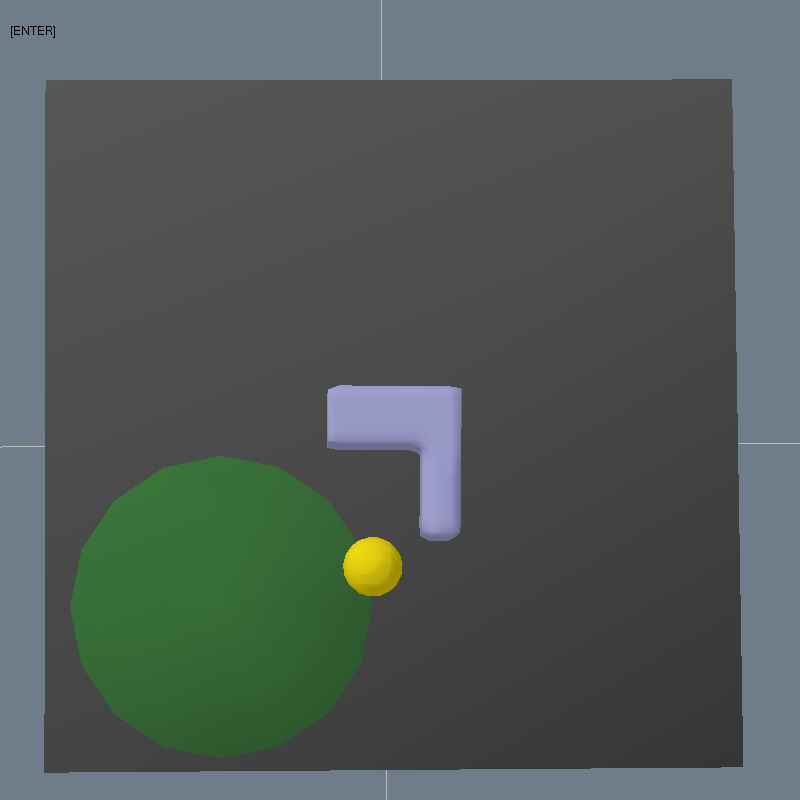}
	\includegraphics[trim={1cm 1cm 2cm 2cm}, clip, width=1.6cm]{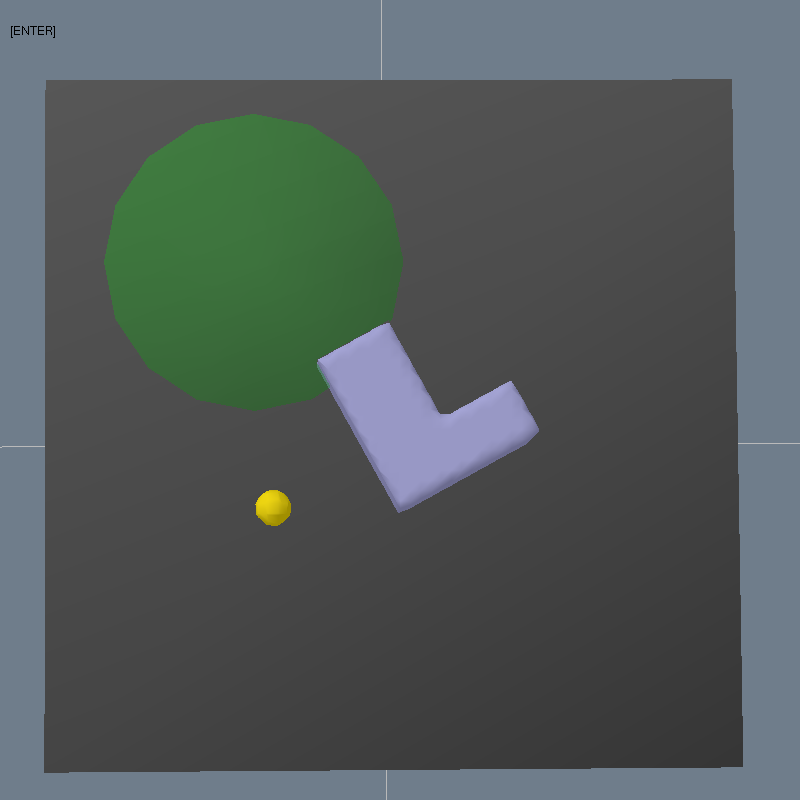}
	\includegraphics[trim={1cm 1cm 2cm 2cm}, clip, width=1.6cm]{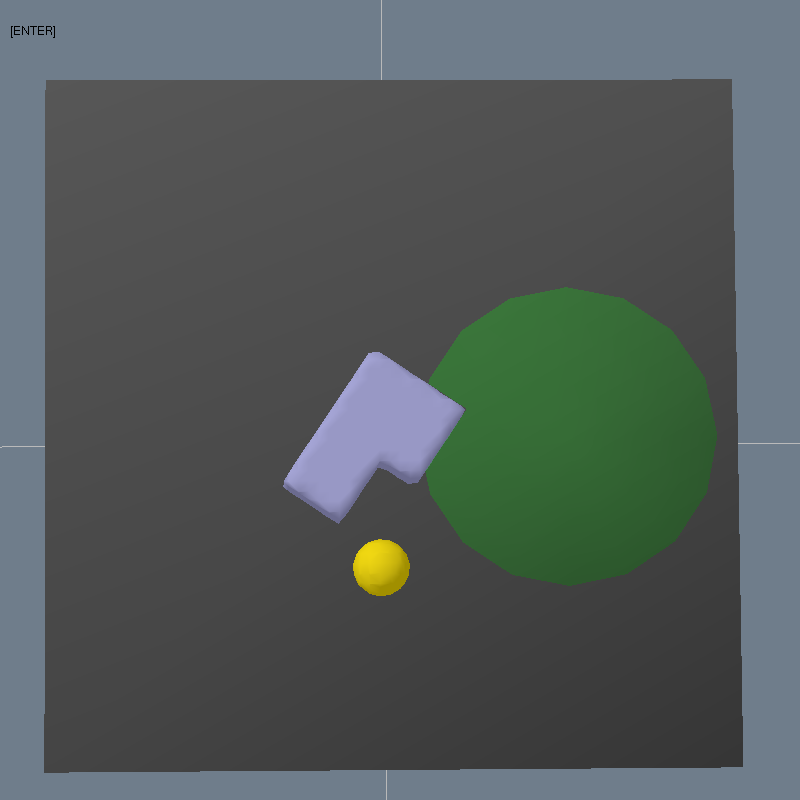}
	\includegraphics[trim={10cm 9cm 9cm 10cm}, clip, width=1.6cm]{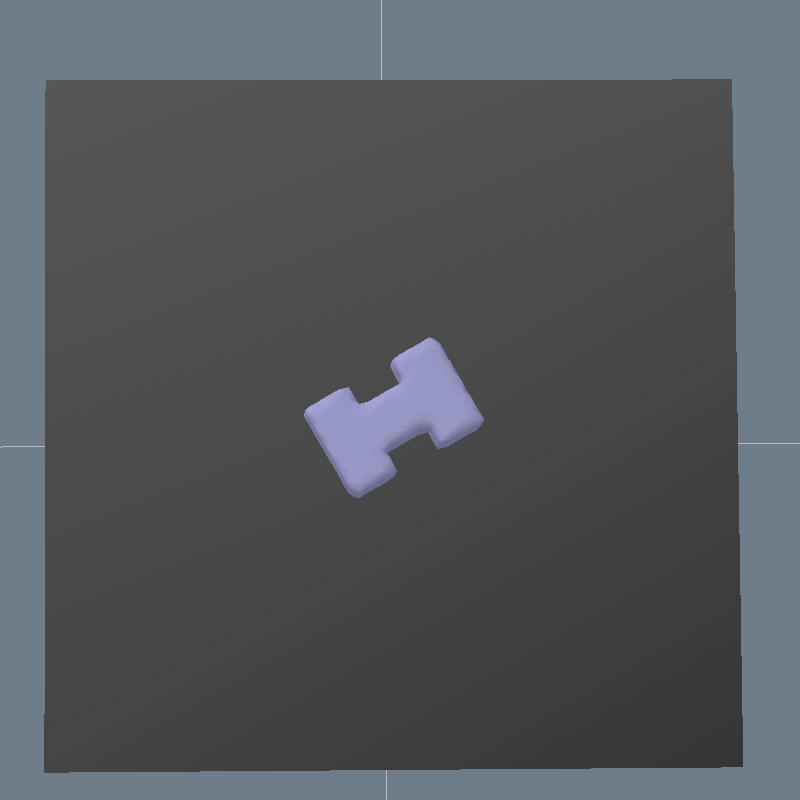}
	\includegraphics[trim={10cm 9cm 9cm 10cm}, clip, width=1.6cm]{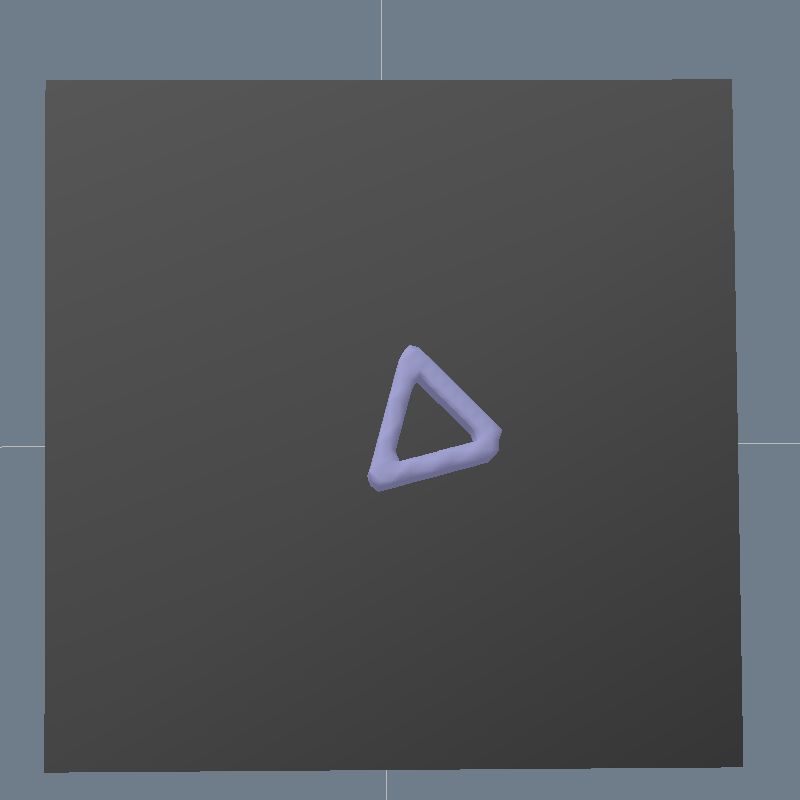}
	\vspace{-0.2cm}
	\caption{\small Different pushing scenarios in evaluation dataset. Yellow is the pusher, green the goal region and light blue the object. The two images from the right show out-of-distribution generalization shapes.\vspace{-0.5cm}}
	\label{fig:pushingScenarios}
\end{figure}

\vspace{-0.2cm}
In this experiment, we consider the task of pushing boxes and L-shaped objects of different dimensions with a spherical pusher of different radii into a goal region $\phi_g$ on a table.
Fig.~\ref{fig:pushingScenarios} visualizes typical objects, pushers and goal regions.
The goal region has to be large enough that all possible objects fit.
We again use Bullet as a simulator to generate data to train a dynamics model of the from described in sec.~\ref{sec:DSDFuncs:dynamic} with $l=1$, i.e.\ $H_F$ is a function of four SDFs $\phi^1_t, \phi^1_{t-1}$ (object) and $\phi^2_t, \phi^2_{t-1}$ (pusher).
In total, 14975 different scenes (including shapes and initial configuration) are sampled where random push actions biased roughly towards the object center are applied until the object leaves the table.
%
Since the dynamics and interaction of the objects in this scenario can be described in the 2D plane, the 3D signed distance functions of the objects are evaluated in the 2D set $\mathcal{X}_h \in \mathbb{R}^{140\times 140}$ only.
Therefore, the model $F$ predicts the dynamics of $\phi^1$ in this 2D projection.

\subsubsection{Forward Prediction Error}\label{sec:exp:pushing:evalPredError}
\vspace{-0.2cm}
\begin{wraptable}[5]{r}{6cm}
	\centering
	\vspace{-0.9cm}
	\small
	\setlength{\tabcolsep}{3pt}
	\caption{\small RMSE [mm] on evaluation dataset.}
	\begin{tabular}{ccc}
		\hline
		& contact phase & no contact phase\\
		\hline
		$F_\text{flow}$ & 3.4 $\pm$ 1.6 & 1.4 $\pm$ 1.8\\
		$F$ & 5.8 $\pm$ 1.7 & 5.2 $\pm$ 1.6\\
		$\phi_t^1 = \phi_{t-1}^1$ & 10.8 $\pm$ 3.4 & 0\\
		\hline
	\end{tabular}
	\vspace{-0.3cm}
	\label{tab:pushingPredError}
\end{wraptable}

Tab.~\ref{tab:pushingPredError} shows the one-step prediction error on the evaluation dataset for the flow model $F_\text{flow}$ \eqref{eq:FFlow} and the direct SDF prediction $F$ \eqref{eq:FDirect}.
The way we utilize the model within the trajectory optimization problem never asks for predictions more than one step into the future.
We train one single dynamics model for both box and L-shaped objects and different pushers.
The prediction error is the RMSE of predicting the correct SDF values in $\mathcal{X}_h$.
The last row shows the error if the model would simply predict the next state as the last state of the object.
As one can see, $F_\text{flow}$ achieves a lower error than $F$. This is due to $F$ having to predict the complete SDF, while $F_\text{flow}$ only the flow.
In phases of the motion where there is no contact between the object and the pusher, both models $F_\text{flow}$ and $F$ have to learn that the object should not move (and $F$ has to predict the complete SDF in this case as well), which is also non-trivial, but they accomplish this with low error.

\subsubsection{Comparison to other Object Representations (Point-Cloud and Occupancy Measure)}
In sec.~\ref{sec:appendix:comparisonObjectRepresentations:resultsPushingPrediction} and sec.~\ref{sec:appendix:comparisonObjectRepresentations:resultsPushingPredictionQSpace} we present and explain a comparison of the forward prediction error between models learned with object representations being SDFs, occupancy measures and point-clouds.
As shown in Tab.~\ref{tab:pushingPredErrorComparison} and Tab.~\ref{tab:pushingPredictionError_Q_comparison}, models learned with the SDF representation outperform models based on point-clouds and occupancy measures in their predictive performance.
Further, in Tab.~\ref{tab:pushingPredictionError_Q_comparison} and sec.~\ref{sec:appendix:comparisonObjectRepresentations:imageSDF}, we also show that one can learn image conditioned SDFs and dynamic models on top of the learned SDF simultaneously with no noticeable performance degradation.

\subsubsection{Planning with the Learned Model and Execution Performance}\label{sec:exp:pushing:evalPlanning}
\begin{figure}
	\TopFloatBoxes
	\begin{floatrow}
		\ffigbox{%
			\if\generatePlots1
				\input{plots/pushPerformance}
			\else
				\includegraphics{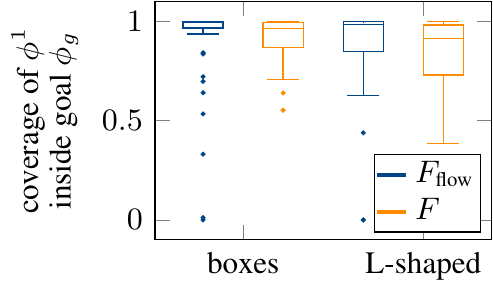}
			\fi
		}{%
			\vspace{-0.3cm}
			\caption{\small Pushing performance on evaluation scenes in terms of the amount of $\phi^1$ that is inside of the goal region at the end of the execution. A value of 1 means that the object is fully contained in the goal region.}
			\label{fig:pushingPerformance}
		}
		\ffigbox{%
			\includegraphics[trim={3cm 3cm 3cm 3cm}, clip, width=2.2cm]{images/pushing/threeObjectsPush_1}
			\includegraphics[trim={0cm 0cm 0cm 0cm}, clip, width=2.2cm]{images/pushing/twoRobotsTouch_1}
			\includegraphics[trim={0cm 0cm 0cm 0cm}, clip, width=2.2cm]{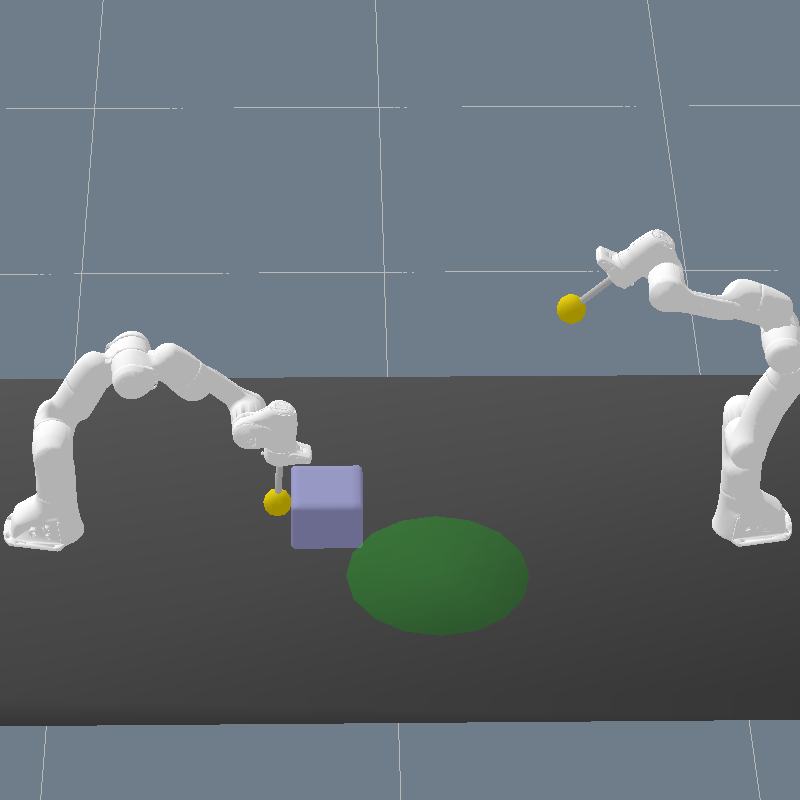}
		}{%
			\vspace{-0.5cm}
			\caption{\small Generalization experiments for pushing scenario. Left: interaction between three objects to solve the task. Middle: the goal is that the right robot arm touches the object. Right: the two robots have to collaborate to push the object into the goal region, which would not be possible with one arm alone.}
			\label{fig:generalization}
		}
		\vspace{-0.5cm}
	\end{floatrow}
\end{figure}

\vspace{-0.2cm}
Having learned the pushing dynamics prediction model, we now utilize it within \eqref{eq:opt} to solve the task of pushing the object into the goal region.
There are four constraints. First, the dynamics model $H_F$ and, second, the goal region $H_\text{g}$.
While this seems to be enough to specify the problem fully, we add two additional constraints, $H_\text{coll}$ and $H_\text{PoC}$.
The discrete variable $s_k$ of \eqref{eq:opt} decides whether there are one or two push phases.
Only in a push phase, $H_\text{PoC}$ is active.
$H_\text{coll}$ is always active.
Similarly to the mug hanging experiment, local minima are a core issue as well.
Therefore, we initialize the pusher position at phase 1 or 2 on a set of 4 different points around the object.
These 4 points around the object are always the same in all scenarios, no matter of the size, shape or orientation of the object.
Compared to other approaches where the action space has to be chosen much more carefully, we believe that this is a rather weak prior.
The initialization also does not start from contact with the object or similar, because our problem contains the challenge of contact establishment and possible breakage to push from a different side to achieve the goal.
Therefore, due to this initialization, there are 20 different optimization problems we solve for each scene (4 for one pushing phase, $4^2$ for two pushing phases).
To evaluate the performance, we execute the planning result with the least constraint violation and cost \emph{open-loop} in the simulator.
Despite the fact that pushing is unstable over long-horizons, our proposed approach achieves a high performance.
As shown in Fig.~\ref{fig:pushingPerformance}, which plots the amount of the object that is inside of the goal region at the end of the execution, using the learned $F_\text{flow}$, the approach achieves 99.7\% (median) coverage of the object inside the goal region on box pushing and 98.4\% (median) on the L-shaped objects (50 evaluation scenes each).
Please note that, although the goal region for small objects seems large, the optimizer usually moves the object until it is just barely inside the goal region and not any further.
Therefore, even very small deviations during the open-loop execution already lead to some parts sticking out.
For the larger objects in the evaluation scenes, the goal region is barely large enough.
With the direct $F$, the performance is a bit worse, but still high (median 96.6\% for boxes, 91.5\% for L-shaped objects).

Sec.~\ref{sec:appendix:analyticPushing} demonstrates that our proposed SDF framework outperforms an approach where the optimization problem is formulated for objects represented as meshes and the analytic dynamic model from \cite{20-toussaint-physicsLGP} for the pushing dynamics. Refer to Fig.~\ref{fig:appendix:compSDFMesh} for quantitative results.
Finally, in sec.~\ref{sec:appendix:planningPushing} and Fig.~\ref{fig:appendix:simpleScenario} we investigate the importance of models learned on top of SDF representations providing useful gradients for planning in comparison to models with point-clouds and occupancy measures.

\subsubsection{Ablation Study}\label{sec:exp:pushing:ablation}
\vspace{-0.2cm}
Sec.~\ref{sec:appendix:ablationPushing} presents an ablation study regarding the importance of the additional objectives $H_\text{coll}$ and $H_\text{PoC}$ in the optimization problem.
Results in Fig.~\ref{fig:pushingAblation} show that while it is possible to solve the scenes without them, the performance greatly increases if they are part of the problem formulation.

\subsubsection{Generalization to Out-of-Distribution Shapes, Multiple Objects, and Robots}
\vspace{-0.2cm}
In sec.~\ref{sec:appendix:pushingOutOfDistribution}, we demonstrate that the framework and the learned model generalizes to shapes beyond the training distribution.
See Fig.~\ref{fig:pushingScenarios} and Fig.~\ref{fig:exp:pushing:generalizationShapes} for those shapes.
Quantitative results presented in Tab.~\ref{tab:shapeGeneralizationPushingPredictionError_Q_comparison} and Fig.~\ref{fig:appendix:pushingGeneralizationShapesPerformance} indicate that the model achieves both high prediction accuracy and high performance when used for planning.
Furthermore, as seen in Tab.~\ref{tab:shapeGeneralizationPushingPredictionError_Q_comparison}, a model learned with SDFs generalizes significantly better than with point-clouds, also relative to the results obtained on-distribution.
We further show in sec.~\ref{sec:appendix:obstacles} and sec.~\ref{sec:appendix:threeObjects} that the framework is capable of generalizing to scenes that contain obstacles (Fig.~\ref{fig:appendix:obstacle}) and a scenario where three objects interact in order to solve the task (Fig.~\ref{fig:appendix:threeObjects} and Fig.~\ref{fig:generalization}).
Finally, in sec.~\ref{sec:appendix:robots} we demonstrate multiple scenarios where the learned pushing dynamics model is embedded into a scene that contains robots.
All these generalization experiments require no change in the methodology or to learn a new model, showing the generality and versatility of our proposed framework.
For more details, refer to the respective sections in the appendix.

\section{Conclusion}\label{sec:conclusion}
\vspace{-0.2cm}
In this work, we have shown that the constraints of a trajectory optimization problem for solving manipulation problems can be formulated in terms of learned functionals of SDFs only.
SDFs can serve as a \emph{common} object representation across completely different tasks.
The functionals can naturally model the interaction between objects of arbitrary shapes and can be learned directly from SDF observations, which is closely connected to perception.
We have shown that learning models on top of SDFs outperform other object representations like point-clouds and occupancy measures both in terms of prediction accuracy and the ability to plan.
The greatest challenge of our framework are local minima of the resulting trajectory optimization problem.
While sampling strategies for initial guesses can mitigate this to some extend, it is an issue, which is not unique to our approach, but many nonlinear trajectory optimization formulations.
While we have considered rigid objects in this work only, we believe that the proposed approach can be extended to deformables as well.

\newpage
\renewcommand{\baselinestretch}{1.0}
\normalsize

\acknowledgments{
Danny Driess thanks the International Max-Planck Research School for Intelligent Systems (IMPRS-IS) for the support.
This research has been supported by the German Research Foundation (DFG) under Germany’s Excellence Strategy -- EXC 2002/1–390523135 ``Science of Intelligence''.
The authors thank the anonymous reviewers for their comments.}

\bibliography{corl}  

\begin{thebibliography}{45}
\providecommand{\natexlab}[1]{#1}
\providecommand{\url}[1]{\texttt{#1}}
\expandafter\ifx\csname urlstyle\endcsname\relax
  \providecommand{\doi}[1]{doi: #1}\else
  \providecommand{\doi}{doi: \begingroup \urlstyle{rm}\Url}\fi

\bibitem[Driess et~al.(2021)Driess, Ha, Tedrake, and Toussaint]{21-driess-ICRA}
D.~Driess, J.-S. Ha, R.~Tedrake, and M.~Toussaint.
\newblock Learning geometric reasoning and control for long-horizon tasks from
  visual input.
\newblock In \emph{Proc{.} of the IEEE Int{.} Conf{.} on Robotics and
  Automation (ICRA)}, 2021.

\bibitem[Garrett et~al.(2021)Garrett, Chitnis, Holladay, Kim, Silver,
  Kaelbling, and Lozano-P{\'e}rez]{garrett2021integrated}
C.~R. Garrett, R.~Chitnis, R.~Holladay, B.~Kim, T.~Silver, L.~P. Kaelbling, and
  T.~Lozano-P{\'e}rez.
\newblock Integrated task and motion planning.
\newblock \emph{Annual review of control, robotics, and autonomous systems},
  2021.

\bibitem[Kaelbling and Lozano-P{\'e}rez(2011)]{kaelbling2010hierarchical}
L.~P. Kaelbling and T.~Lozano-P{\'e}rez.
\newblock Hierarchical planning in the now.
\newblock In \emph{Proc. of the IEEE International Conference on Robotics and
  Automation (ICRA)}, 2011.

\bibitem[Srivastava et~al.(2014)Srivastava, Fang, Riano, Chitnis, Russell, and
  Abbeel]{srivastava14combined}
S.~Srivastava, E.~Fang, L.~Riano, R.~Chitnis, S.~J. Russell, and P.~Abbeel.
\newblock Combined task and motion planning through an extensible
  planner-independent interface layer.
\newblock In \emph{Proc. of the Int. Conf. on Robotics and Automation (ICRA)},
  2014.

\bibitem[Dantam et~al.(2018)Dantam, Kingston, Chaudhuri, and
  Kavraki]{dantam18ijrr}
N.~T. Dantam, Z.~K. Kingston, S.~Chaudhuri, and L.~E. Kavraki.
\newblock An incremental constraint-based framework for task and motion
  planning.
\newblock \emph{International Journal on Robotics Research}, 2018.

\bibitem[Hartmann et~al.(2020)Hartmann, Oguz, Driess, Toussaint, and
  Menges]{hartmann2020robust}
V.~N. Hartmann, O.~S. Oguz, D.~Driess, M.~Toussaint, and A.~Menges.
\newblock Robust task and motion planning for long-horizon architectural
  construction planning.
\newblock arXiv:2003.07754, 2020.

\bibitem[Driess et~al.(2020)Driess, Ha, and Toussaint]{driess2020deep}
D.~Driess, J.-S. Ha, and M.~Toussaint.
\newblock Deep visual reasoning: Learning to predict action sequences for task
  and motion planning from an initial scene image.
\newblock \emph{arXiv:2006.05398}, 2020.

\bibitem[Mordatch et~al.(2012)Mordatch, Todorov, and
  Popovi{\'c}]{mordatch2012discovery}
I.~Mordatch, E.~Todorov, and Z.~Popovi{\'c}.
\newblock Discovery of complex behaviors through contact-invariant
  optimization.
\newblock \emph{ACM Transactions on Graphics (TOG)}, 31\penalty0 (4):\penalty0
  1--8, 2012.

\bibitem[Posa et~al.(2014)Posa, Cantu, and Tedrake]{posa2014direct}
M.~Posa, C.~Cantu, and R.~Tedrake.
\newblock A direct method for trajectory optimization of rigid bodies through
  contact.
\newblock \emph{The International Journal of Robotics Research}, 33\penalty0
  (1):\penalty0 69--81, 2014.

\bibitem[Toussaint et~al.(2018)Toussaint, Allen, Smith, and
  Tenenbaum]{toussaint2018differentiable}
M.~Toussaint, K.~R. Allen, K.~A. Smith, and J.~B. Tenenbaum.
\newblock Differentiable physics and stable modes for tool-use and manipulation
  planning.
\newblock In \emph{Robotics: Science and Systems}, 2018.

\bibitem[Hogan et~al.(2018)Hogan, Grau, and Rodriguez]{hogan18reactive}
F.~R. Hogan, E.~R. Grau, and A.~Rodriguez.
\newblock Reactive planar manipulation with convex hybrid {MPC}.
\newblock In \emph{Int. Conf. on Robotics and Automation (ICRA)}, 2018.

\bibitem[Doshi et~al.(2020)Doshi, Hogan, and Rodriguez]{doshi2020hybrid}
N.~Doshi, F.~R. Hogan, and A.~Rodriguez.
\newblock Hybrid differential dynamic programming for planar manipulation
  primitive.
\newblock In \emph{Int. Conf. on Robotics and Automation (ICRA)}, 2020.

\bibitem[Toussaint et~al.(2020)Toussaint, Ha, and
  Driess]{20-toussaint-physicsLGP}
M.~Toussaint, J.-S. Ha, and D.~Driess.
\newblock Describing physics for physical reasoning: Force-based sequential
  manipulation planning.
\newblock \emph{IEEE Robotics and Automation Letters}, 2020.

\bibitem[Driess et~al.(to appear)Driess, Ha, and Toussaint]{21-driess-IJRR}
D.~Driess, J.-S. Ha, and M.~Toussaint.
\newblock Learning to solve sequential physical reasoning problems from a scene
  image.
\newblock \emph{The International Journal of Robotics Research}, to appear.

\bibitem[Chen and Zhang(2019)]{chen2019learning}
Z.~Chen and H.~Zhang.
\newblock Learning implicit fields for generative shape modeling.
\newblock In \emph{Proceedings of the IEEE/CVF Conference on Computer Vision
  and Pattern Recognition}, 2019.

\bibitem[Liu et~al.(2019)Liu, Saito, Chen, and Li]{liu2019learning}
S.~Liu, S.~Saito, W.~Chen, and H.~Li.
\newblock Learning to infer implicit surfaces without 3d supervision.
\newblock \emph{arXiv preprint arXiv:1911.00767}, 2019.

\bibitem[Mescheder et~al.(2019)Mescheder, Oechsle, Niemeyer, Nowozin, and
  Geiger]{OccupancyNetworks}
L.~Mescheder, M.~Oechsle, M.~Niemeyer, S.~Nowozin, and A.~Geiger.
\newblock Occupancy networks: Learning 3d reconstruction in function space.
\newblock In \emph{Proceedings IEEE Conf. on Computer Vision and Pattern
  Recognition (CVPR)}, 2019.

\bibitem[Park et~al.(2019)Park, Florence, Straub, Newcombe, and
  Lovegrove]{Park_2019_CVPR}
J.~J. Park, P.~Florence, J.~Straub, R.~Newcombe, and S.~Lovegrove.
\newblock Deepsdf: Learning continuous signed distance functions for shape
  representation.
\newblock In \emph{The IEEE Conference on Computer Vision and Pattern
  Recognition (CVPR)}, June 2019.

\bibitem[Sitzmann et~al.(2020)Sitzmann, Chan, Tucker, Snavely, and
  Wetzstein]{sitzmann2019metasdf}
V.~Sitzmann, E.~R. Chan, R.~Tucker, N.~Snavely, and G.~Wetzstein.
\newblock Metasdf: Meta-learning signed distance functions.
\newblock In \emph{arXiv}, 2020.

\bibitem[Atzmon and Lipman(2020)]{Atzmon_2020_CVPR}
M.~Atzmon and Y.~Lipman.
\newblock Sal: Sign agnostic learning of shapes from raw data.
\newblock In \emph{IEEE/CVF Conference on Computer Vision and Pattern
  Recognition (CVPR)}, June 2020.

\bibitem[Jiang et~al.(2020)Jiang, Sud, Makadia, Huang, Nie{\ss}ner, Funkhouser,
  et~al.]{jiang2020local}
C.~Jiang, A.~Sud, A.~Makadia, J.~Huang, M.~Nie{\ss}ner, T.~Funkhouser, et~al.
\newblock Local implicit grid representations for 3d scenes.
\newblock In \emph{Proceedings of the IEEE/CVF Conference on Computer Vision
  and Pattern Recognition}, pages 6001--6010, 2020.

\bibitem[Macklin et~al.(2020)Macklin, Erleben, Müller, Chentanez, Jeschke, and
  Corse]{SDFCollision}
M.~Macklin, K.~Erleben, M.~Müller, N.~Chentanez, S.~Jeschke, and Z.~Corse.
\newblock Local optimization for robust signed distance field collision.
\newblock \emph{Proceedings of the ACM on Computer Graphics and Interactive
  Techniques}, 2020.

\bibitem[Fuhrmann and Sobottka(2003)]{Fuhrmann2003DistanceFF}
A.~Fuhrmann and G.~Sobottka.
\newblock Distance fields for rapid collision detection in physically based
  modeling.
\newblock 2003.

\bibitem[Hauser(2018)]{hauser1}
K.~Hauser.
\newblock Semi-infinite programming for trajectory optimization with non-convex
  obstacles.
\newblock \emph{The International Journal of Robotics Research}, 2018.

\bibitem[Zhang and Hauser(2021)]{zhang2021semi}
M.~Zhang and K.~Hauser.
\newblock Semi-infinite programming with complementarity constraints for pose
  optimization with pervasive contact.
\newblock In \emph{IEEE International Conference on Robotics and Automation},
  2021.

\bibitem[Pfrommer et~al.(2020)Pfrommer, Halm, and
  Posa]{pfrommer2020contactnets}
S.~Pfrommer, M.~Halm, and M.~Posa.
\newblock Contactnets: Learning of discontinuous contact dynamics with smooth,
  implicit representations.
\newblock \emph{Conference on Robot Learning}, 2020.

\bibitem[Breyer et~al.(2020)Breyer, Chung, Ott, Roland, and
  Juan]{breyer2020volumetric}
M.~Breyer, J.~J. Chung, L.~Ott, S.~Roland, and N.~Juan.
\newblock Volumetric grasping network: Real-time 6 dof grasp detection in
  clutter.
\newblock In \emph{Conference on Robot Learning}, 2020.

\bibitem[Jiang et~al.(2021)Jiang, Zhu, Svetlik, Fang, and
  Zhu]{jiang2021synergies}
Z.~Jiang, Y.~Zhu, M.~Svetlik, K.~Fang, and Y.~Zhu.
\newblock Synergies between affordance and geometry: 6-dof grasp detection via
  implicit representations.
\newblock \emph{arXiv preprint arXiv:2104.01542}, 2021.

\bibitem[Van~der Merwe et~al.(2020)Van~der Merwe, Lu, Sundaralingam, Matak, and
  Hermans]{van2020learning}
M.~Van~der Merwe, Q.~Lu, B.~Sundaralingam, M.~Matak, and T.~Hermans.
\newblock Learning continuous 3d reconstructions for geometrically aware
  grasping.
\newblock In \emph{Int. Conf. on Robotics and Automation (ICRA)}, 2020.

\bibitem[Ebert et~al.(2018)Ebert, Finn, Dasari, Xie, Lee, and
  Levine]{visualforesight}
F.~Ebert, C.~Finn, S.~Dasari, A.~Xie, A.~Lee, and S.~Levine.
\newblock Visual foresight: Model-based deep reinforcement learning for
  vision-based robotic control.
\newblock \emph{arXiv:1812.00568}, 2018.

\bibitem[Kandukuri et~al.(2021)Kandukuri, Achterhold, Moeller, and
  Stueckler]{kandukuri2021learning}
R.~Kandukuri, J.~Achterhold, M.~Moeller, and J.~Stueckler.
\newblock Learning to identify physical parameters from video using
  differentiable physics.
\newblock In \emph{Pattern Recognition: 42nd DAGM German Conference}, 2021.

\bibitem[Xu et~al.(2019)Xu, Wu, Zeng, Tenenbaum, and Song]{xu2019densephysnet}
Z.~Xu, J.~Wu, A.~Zeng, J.~B. Tenenbaum, and S.~Song.
\newblock Densephysnet: Learning dense physical object representations via
  multi-step dynamic interactions.
\newblock \emph{arXiv preprint arXiv:1906.03853}, 2019.

\bibitem[Xu et~al.(2020)Xu, He, Wu, and Song]{xu2020learning}
Z.~Xu, Z.~He, J.~Wu, and S.~Song.
\newblock Learning 3d dynamic scene representations for robot manipulation.
\newblock In \emph{Conference on Robotic Learning (CoRL)}, 2020.

\bibitem[Manuelli et~al.(2020)Manuelli, Li, Florence, and
  Tedrake]{manuelli2020keypoints}
L.~Manuelli, Y.~Li, P.~Florence, and R.~Tedrake.
\newblock Keypoints into the future: Self-supervised correspondence in
  model-based reinforcement learning.
\newblock In \emph{Conference on Robotic Learning (CoRL)}, 2020.

\bibitem[Byravan and Fox(2017)]{byravan2017se3}
A.~Byravan and D.~Fox.
\newblock Se3-nets: Learning rigid body motion using deep neural networks.
\newblock In \emph{IEEE International Conference on Robotics and Automation
  (ICRA)}, pages 173--180, 2017.

\bibitem[Zeng et~al.(2020)Zeng, Florence, Tompson, Welker, Chien, Attarian,
  Armstrong, Krasin, Duong, Sindhwani, and Lee]{zeng2020transporter}
A.~Zeng, P.~Florence, J.~Tompson, S.~Welker, J.~Chien, M.~Attarian,
  T.~Armstrong, I.~Krasin, D.~Duong, V.~Sindhwani, and J.~Lee.
\newblock Transporter networks: Rearranging the visual world for robotic
  manipulation.
\newblock \emph{Conference on Robot Learning (CoRL)}, 2020.

\bibitem[Simeonov et~al.(2020)Simeonov, Du, Kim, Hogan, Agrawal, and
  Rodriguez]{simeonov2020learning}
A.~Simeonov, Y.~Du, B.~Kim, F.~R. Hogan, P.~Agrawal, and A.~Rodriguez.
\newblock Learning to plan with pointcloud affordances for general-purpose
  dexterous manipulation.
\newblock \emph{Conference on Robot Learning}, 2020.

\bibitem[Mukherjee et~al.(2020)Mukherjee, Paxton, Mousavian, Fishman,
  Likhachev, and Fox]{mukherjee2020sim}
S.~Mukherjee, C.~Paxton, A.~Mousavian, A.~Fishman, M.~Likhachev, and D.~Fox.
\newblock Sim-to-real task planning and execution from perception via
  reactivity and recovery.
\newblock \emph{arXiv preprint arXiv:2011.08694}, 2020.

\bibitem[Sutanto et~al.(2020)Sutanto, Fern{\'a}ndez, Englert, Ramachandran, and
  Sukhatme]{sutanto2020learning}
G.~Sutanto, I.~M.~R. Fern{\'a}ndez, P.~Englert, R.~K. Ramachandran, and G.~S.
  Sukhatme.
\newblock Learning equality constraints for motion planning on manifolds.
\newblock \emph{arXiv preprint arXiv:2009.11852}, 2020.

\bibitem[You et~al.(2021)You, Shao, Migimatsu, and Bohg]{you2021omnihang}
Y.~You, L.~Shao, T.~Migimatsu, and J.~Bohg.
\newblock Omnihang: Learning to hang arbitrary objects using contact point
  correspondences and neural collision estimation.
\newblock In \emph{2021 IEEE International Conference on Robotics and
  Automation (ICRA)}. IEEE, 2021.

\bibitem[Danielczuk et~al.(2020)Danielczuk, Mousavian, Eppner, and
  Fox]{danielczuk2020object}
M.~Danielczuk, A.~Mousavian, C.~Eppner, and D.~Fox.
\newblock Object rearrangement using learned implicit collision functions.
\newblock \emph{arXiv:2011.10726}, 2020.

\bibitem[Coumans and Bai(2016--2021)]{coumans2021}
E.~Coumans and Y.~Bai.
\newblock Pybullet, a python module for physics simulation for games, robotics
  and machine learning, 2016--2021.

\bibitem[Yan(2019)]{PytorchPointnetPointnet2}
X.~Yan.
\newblock Pointnet/pointnet++ pytorch.
\newblock \emph{\url{https://github.com/yanx27/Pointnet_Pointnet2_pytorch}},
  2019.

\bibitem[Pfaff et~al.(2020)Pfaff, Fortunato, Sanchez-Gonzalez, and
  Battaglia]{pfaff2020learning}
T.~Pfaff, M.~Fortunato, A.~Sanchez-Gonzalez, and P.~W. Battaglia.
\newblock Learning mesh-based simulation with graph networks.
\newblock \emph{arXiv preprint arXiv:2010.03409}, 2020.

\bibitem[Driess et~al.(2019)Driess, Schmitt, and Toussaint]{19-driess-IROS}
D.~Driess, S.~Schmitt, and M.~Toussaint.
\newblock Active inverse model learning with error and reachable set estimates.
\newblock In \emph{Proc{.} of the IEEE Int{.} Conf{.} on Intelligent Robots and
  Systems (IROS)}, 2019.

\end{thebibliography}

\newpage
\appendix

\renewcommand{\baselinestretch}{1.0}
\normalsize

\section{Comparisons to Point-Cloud and Occupancy Measure Object Representations as well as SDFs Learned from Images}
\label{sec:appendix:comparisonObjectRepresentations}
The purpose of this section is to investigate the importance of the object representations being signed-distance functions for both the mug-hanging and the pushing scenario.
In order to do so, we consider two other object representations, namely occupancy measures, which are functions that indicate whether there is an object at a certain location or not, and point-clouds which represent the surface of an object as a set of points in 3D space.
Moreover, we also show that one can learn an image-conditioned SDF and the dynamics model simultaneously for the pushing scenario while maintaining high performance.

\subsection{Occupancy Measure Object Representation}
\label{sec:appendix:comparisonObjectRepresentations:occupancy}
An occpuancy measure $\psi : \mathbb{R}^3 \rightarrow \left\{-1, 1\right\}$ is a function that is defined such that $\psi(x)$ is $-1$ if $x\in\mathbb{R}^3$ is inside the object $\psi$ represents and $+1$ if $x$ is outside the object (for the boundary of the object there are multiple conventions).
The function $\psi$ can therefore be interpreted as an indicator whether there is an object at $x$ or not.

Formally, we can utilize a signed-distance function $\phi$ to define the corresponding occupancy measure $\psi$ via
\begin{align}
	\psi(x) = \text{sign}(\phi(x)). \label{eq:occupancySign}
\end{align}
To make $\psi$ differentiable, one can use
\begin{align}
	\psi_\text{s}(x) = \tanh(b~\phi(x)) \label{eq:occupancyTanh}
\end{align}
for $b>0$.
The smaller $b$, the smoother the boundary and hence the gradients become more well-behaved.
On the other hand, if $b$ is too small, then the objects do not have well-defined boundaries anymore, which is an issue, since we want to model both collision avoidance and contact establishment at the same time.
Further, if $b$ is very small, then $\psi$ locally around the surface of the object maintains the information of the SDF $\phi$ it was constructed with.
We therefore test values of $b = 100$ and $b = 1000$.

Compared to occupancy values on a static grid, $\psi$ contains more information, since it is a function of $x\in\mathbb{R}^3$, which means that it can also be transformed rigidly in space using \eqref{eq:rigidTransformation} as for $\phi$.
For static occupancy grids, such transformations are also possible, but they require a post-processing step to recreate a valid occupancy grid after the transformation.

\subsection{Point-Cloud Object Representation}\label{sec:appendix:comparisonObjectRepresentations:PC}
As another object representation, we consider point-clouds, which are a set of 3D points on the surface of the object.
Formally, we can define a point-cloud $\mathcal{P}$ for an object via the corresponding signed-distance function $\phi$ of the object as
\begin{align}
	\mathcal{P} = \left\{x_1, \ldots, x_N \in\mathbb{R}^3 : \forall_{i=1,\ldots, N}~ \phi(x_i) = 0 \right\}.
\end{align}
In order to make the comparison fair, the number $N$ of points is chosen such that the resolution of the SDF and the point-cloud is similar.
More precisely, we first run the marching cube algorithm on the SDF evaluated on the same $\mathcal{X}_h$ as in the other experiments and then extract $\mathcal{P}$ as the vertex points of the obtained mesh.

As described in sec.~\ref{sec:DSDFuncs:neuralNet}, \ref{sec:appendix:hangingNetworkArchitecture} and \ref{sec:appendix:pushingNetworkArchitecture}, we can encode the SDF as input to the functionals $H$ via grid evaluation of the SDF function, followed by a convolutional neural network encoder architecture.
For point-clouds, this is not directly possible.
Therefore, we utilize a so-called pointnet architecture to encode the input point-clouds into a feature vector from which the dynamics model and the kinematic success model is defined.
We use the pointnet and pointnet++ implementation from \cite{PytorchPointnetPointnet2} for the point-cloud encoding.

\subsection{Dynamics Model via Predicting Rigid Transformations}\label{sec:appendix:dynamicsModelRigidTransformation}
Our proposed forward dynamics model \eqref{eq:FDirect} and \eqref{eq:FFlow}, described in sec.~\ref{sec:DSDFuncs:dynamic}, predicts the SDF \emph{function} $\phi^1_t$ of object 1 at time $t$ as a function of the history of SDF observations of the object $\phi^1_{t-l:t-1}$ until time $t-1$ and the motion of another object $\phi^2_{t-l:t}$ until time t.
The model does not predict the values on a static grid, but as a function that can be queried in $\mathbb{R}^3$.

For point-clouds, there is no notion of a function that could be predicted.
Instead, a forward dynamics model in point-cloud space means to predict the positions of the points at the next timestep.
To achieve this, we learn a model
\begin{align}
	\Delta q^1 &= F_q\big(\mathcal{P}^1_{t-1}, \mathcal{P}^2_{t-1}, \mathcal{P}^2_t\big)
\end{align}
that predicts the rigid transformation $\Delta q^1$ (in case of the pushing scenario $\Delta q^1 \in\mathbb{R}^3$, $x,y$-translation and $\alpha$ rotation) of the point-cloud $\mathcal{P}^1_{t-1}$ of object 1 at time $t-1$ as a function of the point-clouds $\mathcal{P}^2_{t-1}$ and $\mathcal{P}^2_t$ of object 2 such that
\begin{align}
	\mathcal{P}^1_{t} = R\big(\Delta q^1\big)\mathcal{P}^1_{t-1} + r\big(\Delta q^1\big).
\end{align}
This means whereas for training the model based on SDFs and occupancy measures it was sufficient to have a dataset of just SDF/occupancy observations, the training process for the point-cloud model requires those relative transformations of the point-cloud either as ground-truth data or via a (non-trivial) point-cloud registration step.
We assume to have access to ground-truth rigid transformations $\Delta q^1$ to train the point-cloud model.

In order to compare the performance of this point-cloud model with the SDF and occupancy measure approach, we also train models
\begin{align}
	\Delta q^1 &= F_q\big(\phi^1_{t-1}, \phi^2_{t-1}, \phi^2_t\big)\\
	\Delta q^1 &= F_q\big(\psi^1_{t-1}, \psi^2_{t-1}, \psi^2_t\big)
\end{align}
for both the SDF and occupancy measure that predict the rigid transformations of these object representations (as functions).
In this case
\begin{align}
	\phi^1_t = T\big(\Delta q^1\big)\big[\phi^1_{t-1}\big]\\
	\psi^1_t = T\big(\Delta q^1\big)\big[\psi^1_{t-1}\big]
\end{align}
with the transformation $T$ as defined in \eqref{eq:rigidTransformation}.

As a side note, while we have focused on rigid objects in this work, our approach of predicting a function would directly be applicable to the deformable case for SDFs and occupancy measures, which would not directly be possible with a point-cloud model that predicts the rigid transformation of the points. 

\subsection{Learning Image Conditioned SDF and Dynamics Model Simultaneously for Pushing Scenario}
\label{sec:appendix:comparisonObjectRepresentations:imageSDF}
The experiments in this work assume to have access to the SDFs of the involved objects.
Here, we show for the pushing scenario that it is possible to train a neural network that predicts the SDF from an image observation of the corresponding object and the dynamics model based on top of the learned SDF simultaneously with very little performance degradation compared to having the ground-truth SDF available.
For this, we assume to have a segmented image $I\in\mathbb{R}^{h_I\times w_I}$ of each object.
Then, we learn one function for all objects such that
\begin{align}
	\phi(\cdot, I)\in\Phi
\end{align}
is the SDF corresponding to the object masked in the image $I$.
We assume an overhead camera perspective of the table such that the images $I$ are masked depth images of the objects from above.
We learn a single $\phi$ for both the pusher and the object that is being pushed.
Also note that the masked image is not transformed into the origin or some other canonical frame, i.e.\ the image shows the configuration (position and rotation) of the object as it is in the scene.
Therefore, no pose estimation step or similar is necessary, $\phi(\cdot, I)$ is the SDF of the object in the configuration as in the scene.

For quantitative results of this experiment, see Tab.~\ref{tab:pushingPredictionError_Q_comparison} (second row) and sec.~\ref{sec:appendix:comparisonObjectRepresentations:resultsPushingPrediction} below.

\subsection{Comparison Results for Mug-Hanging}\label{sec:appending:comparisonObjectRepresentations:mugs}

Tab.~\ref{tab:mugsComparison} shows the success rates in terms of the percentage of successfully solved scenes for the mug hanging experiment for the object representations being SDFs (proposed approach) and occupancy measures/point-clouds as baselines.
We investigate various different approaches, namely optimization + sampling, optimization only, sampling only with different feasibility thresholds $\kappa$, and two different pointnet architectures.

\begin{table}[!b]
	\centering
	\begin{tabular}{cccccc}
		\hline
		&& solution & simulation & collision & \textbf{total scenes} \\
		&& found & success & free & \textbf{solved} \\
		\hline
		\multirow{5}{1.5cm}{signed- distance $\phi$} & \textbf{opt. + sampling} & 98.7\% & 88.5\% & 100.0\% & \textbf{87.3\%} \\
		&opt. only & 51.3\% & 93.5\% & 98.6\% & 47.3\% \\
		&sampling only $\kappa_1$ & 83.8\% & 82.3\% & 100.0\% & 68.9\% \\
		&sampling only $\kappa_2$ & 100\% & 35.6\% & 100.0\% & 35.6\% \\
		&sampling only $\kappa_3$ & 26.0\% & 87.5\% & 100.0\% & 22.8\% \\ 
		\hline
		\multirow{4}{1.5cm}{occupancy measure $\psi$} &$\psi_s$ + opt.\ + sampling & 34.0\% & 51.0\% & 100.0\% & 17.3\% \\
		&$\psi$ + sampling $\kappa_1$ & 100\% & 17.3\% & 100\% & 17.3\% \\
		&$\psi$ + sampling $\kappa_2$ & 100\% & 22.0\% & 100\% & 22.0\% \\
		&$\psi$ + sampling $\kappa_3$ & 100\% & 34.0\% & 100\% & 34.0\% \\
		\hline
		\multirow{8}{1.8cm}{point-cloud pointnet} & opt.\ + sampling & 100.0\% & 79.3\% & 14.3\% & 11.3\% \\
		& opt.\ + sampling + SDF collision & 99.3\% & 67.1\% & 100.0\% & 66.7\% \\
		& sampling $\kappa_1$, no collision & 100.0\% & 48.0\% & 23.6\% & 11.3\% \\
		& sampling $\kappa_2$, no collision & 100.0\% & 36.0\% & 53.7\% & 19.3\% \\
		& sampling $\kappa_3$, no collision & 100.0\% & 78.0\% & 32.5\% & 25.3\% \\
		& sampling $\kappa_1$ + SDF collision & 99.3\% & 35.6\% & 100.0\% & 35.3\% \\
		& sampling $\kappa_2$ + SDF collision & 100.0\% & 21.3\% & 100.0\% & 21.3\% \\
		& sampling $\kappa_3$ + SDF collision & 73.3\% & 70.9\% & 100.0\% & 52.0\% \\
		\hline
		\multirow{8}{1.8cm}{point-cloud pointnet++} & opt.\ + sampling & 0\% & -- & -- & 0\% \\
		& opt.\ + sampling + SDF collision & 0\% & -- & -- & 0\% \\
		& sampling $\kappa_1$, no collision & 100.0\% & 82.0\% & 63.4\% & 52.0\% \\
		& sampling $\kappa_2$, no collision & 100.0\% & 36.0\% & 53.7\% & 19.3\% \\
		& sampling $\kappa_3$, no collision & 0\% & -- & -- & 0\% \\
		& sampling $\kappa_1$ + SDF collision & 0\% & -- & -- & 0\% \\
		& sampling $\kappa_2$ + SDF collision & 100.0\% & 24.0\% & 100.0\% & 24.0\% \\
		& sampling $\kappa_3$ + SDF collision & 0\% & -- & -- & 0\% \\
		\hline
	\end{tabular}
	\caption{Comparison of success rates for mug hanging experiment between an object representation based on signed-distance functions $\phi$ (proposed approach), occupancy measures $\psi$ (baseline) and point-clouds (baseline). Total scenes solved (last column) means the percentage of scenes in the evaluation dataset for which a solution was found where the learned model predicts success (first column) that is actually stable when dropped in the simulator from the optimized solution configuration (second column) and is not in collision before dropping (third column).}
	\label{tab:mugsComparison}
\end{table}

The results in Tab.~\ref{tab:mugsComparison} have to be interpreted in the following way.
For each method (object representation, solution method, thresholds), we report in the first column the percentage of scenes for which the method (optimization + sampling, optimization only, sampling with different thresholds) found a solution within the maximum allowed functional evaluations.
To allow for a fair comparison, both the optimization problem (including its up to 20 restarts) and the sampling procedure are allowed to use the exact same total maximum number of functional evaluations (20,000).
Then the second column shows the percentage of the found solutions where the solution configuration leads to a stable hanging of the mug on the hook when dropped in the simulator.
Since the simulator also simulates if the solution configuration of the mug is in collision with the hook, we report in the third column the percentage of the successfully simulated configurations that were initially not in collision with the hook, because they cannot be counted as success.
Taking all these into account, the last column shows the total percentage of the scenes in the evaluation dataset for which a valid, i.e.\ collision free, and stable solution was found.

Note that due to the way it was generated, the test and training data distribution has a significantly different ratio of success/failure examples than when sampling the model until it predicts success. 
Therefore, when using the same feasibility threshold $\kappa_2$ as for evaluating the test data, the model can be overly optimistic.
Hence, we investigate different feasibility thresholds $\kappa_1 = 0.15$, $\kappa_2 = 1.5$, $\kappa_3 = 0.015$, i.e.\ if $H_\text{hang}(\phi^1, \phi^2) < \kappa_i$, then the model predicts success.

For the SDF model and sampling only, using this threshold $\kappa_2$, in 100\% of the cases a solution is found, but only 35.6\% of those are actually successful in simulation.
For a 10-times smaller threshold $\kappa_1$, the success rate is high again (92.3\%), but only in 83.8\% of the scenes a feasible solution is sampled within the computational budget, leading to a total of 68.9\% solved scenes with sampling compared to 87.3\% with optimization and sampling, which shows the advantages of our models based on SDFs being differentiable with informative gradients.

As one can see in Tab.~\ref{tab:mugsComparison}, the performance using the occupancy measure $\psi$ or $\psi_s$ is significantly worse than with $\phi$.
The best result with the occupancy measure was 34.0\%, compared to 87.3\% with our proposed approach.
Especially with sampling, the model based on $\psi$ very often mistakenly predicts that a sampled configuration will lead to success.
With optimization ($\psi_s$ + opt.\ + sampling), the success rate for a found solution is higher (51.0\%). 
However, in only 34.0\% of the scenes, the solver converges to a configuration within 20 restarts where the learned model with $\psi_s$ predicts success.
Therefore, the total percentage of successfully solved scenes in this case is only 17.3\% with $\psi_s$ compared to 87.3\% with $\phi$.
One reason for the optimizer finding a solution in only 34.0\% of the scenes with the model based on $\psi_s$ is the fact that, although $\psi_s$ is differentiable, its gradients are mostly zero.
For $\psi_s$, we chose $b = 1000$.

For the point-cloud representation, there is no easy way to define the collision constraint $H_\text{coll}$, in contrast to both SDF and occupancy measure representations, where this can be realized by \eqref{eq:pairCollision}.
In \cite{you2021omnihang}, where they also consider hanging objects with a point-cloud based object representation, they have to learn a separate collision predictor.
We therefore additionally evaluate the point-cloud representation with the collision constraint \eqref{eq:pairCollision} defined based on the same SDF of the object as for the evaluation of the SDF approach.

The best achieved result with pointnet is 66.7\% of total solved scenes using optimization and the (differentiable) SDF collision constraint.
However, without this additional knowledge of the SDF representation for differentiable collision checking when optimizing with the learned pointnet model, the best achieved result is 25.3\%.
For pointnet++, the best achieved percentage of totally solved scenes is 52.0\% with sampling.

The results for the occupancy measure have been obtained by keeping everything the same as in sec.~\ref{sec:exp:mugHanging}, i.e.\ same network architecture, same training, test and evaluation datasets, same thresholds, sampling strategies (with same random seeds) etc., except for $\phi$ being replaced by $\psi$ and $\psi_s$ when learning and evaluating $H_\text{hang}$.
Furthermore, the collision constraint $\mathcal{H}_\text{coll}$ is also evaluated with $\psi$ or $\psi_s$, respectively.
For the point-cloud representation, we had to tune the hyperparameters of the pointnet++ architecture much more than we did tune the hyperparameters of our proposed approach (with the default parameters of pointnet++, we achieved a success rate of 0\%).
The gradients that a model learned with pointnet++ provides were not suitable for optimization at all (found solutions 0\%).
The training, test and evaluation datasets are also the same for the point-cloud architectures with the point-clouds obtained from the SDFs as described in sec.~\ref{sec:appendix:comparisonObjectRepresentations:PC}.
The thresholds and sampling strategies are also the same (with same random seeds).

\subsection{Comparison Results for Pushing Objects on a Table}\label{{sec:appending:comparisonObjectRepresentations:pushing}}
\subsubsection{Forward Prediction Error in Observation Space}\label{sec:appendix:comparisonObjectRepresentations:resultsPushingPrediction}
Tab.~\ref{tab:pushingPredErrorComparison} shows the mean RMSE of the one-step predictions on the evaluation dataset (different random seed as for the training and test dataset) for both the proposed approach where the forward model is learned as a functional of SDFs $\phi$ and occupancy measures $\psi$ as a baseline.
More precisely, the error is calculated very similar to the dynamics model functional \eqref{eq:FFunctional} as follows
\begin{align}
	\frac{1}{n}\sum_{i=1}^{n}\sqrt{\frac{1}{\left|\mathcal{X}_h\right|}\sum_{x\in\mathcal{X}_h}^{\left|\mathcal{X}_h\right|}\Big(\phi^1_{1,i}(x) - F\big[\phi^1_{0, i}(\mathcal{X}_h), \phi^2_{0:1,i}(\mathcal{X}_h)\big](x)\Big)^2 }
\end{align}
with $\phi^k_{t,i}(x)$ meaning the SDF value of object $k$ at time $t$ for scene $i$ at $x\in\mathcal{X}_h$ with $\mathcal{X}_h$ the discrete grid.
The index $t$ is $t=0$ or $t=1$ only here, since we are only interested in the one-step predictions, which is the data the dataset contains.
We investigate this prediction error for $\psi$ as defined in \eqref{eq:occupancySign} and its smooth version $\psi_s$ defined in \eqref{eq:occupancyTanh} for different values of $b$.
The absolute numbers of these errors in Tab.~\ref{tab:pushingPredErrorComparison} are not directly comparable between $\psi$ and $\phi$, since the predictions/ground-truth values have different magnitudes.
However, we show for each case the error if the model simply predicts the last observed SDF/occupancy measure, i.e.\ either $\phi^1_t = \phi^1_{t-1}$ or $\psi^1_t = \psi^1_{t-1}$.
Comparing a model based on SDFs vs occupancy measures, one can see that with SDFs, the prediction error relative to predicting the last observation as the future is significantly lower than with the occupancy measure.
These results have been obtained by keeping everything the same as in sec.~\ref{sec:exp:pushing}, except for the field values to train/test and evaluate the model being replaced by $\psi^k(\mathcal{X}_h)$ instead of $\phi^k(\mathcal{X}_h)$.

\begin{table}
	\caption{Comparison of mean RMSE [mm] on evaluation dataset for pushing scenario between an object representation based on signed-distance functions $\phi$ (proposed approach) and occupancy measures $\psi$ (baseline) with different parameters $b$ for $\psi_s$ as defined in \eqref{eq:occupancyTanh}.}
	\begin{tabular}{cccc}
		\hline
		&& contact phase & no contact phase\\
		\hline
		\multirow{3}{1.5cm}{signed- distance $\phi$} & $F_\text{flow}$ & 3.4 $\pm$ 1.6 & 1.4 $\pm$ 1.8\\
		&$F$ & 5.8 $\pm$ 1.7 & 5.2 $\pm$ 1.6\\
		&$\phi_t^1 = \phi_{t-1}^1$ & 10.8 $\pm$ 3.4 & 0\\
		\hline
		\multirow{6}{1.5cm}{occupancy measure $\psi$} & $F_\text{flow}$, $\psi_s$ with $b = 100$ & 59.0 $\pm$ 16.6 & 10.0 $\pm$ 11.7 \\
		& $\psi_{s, t}^1  = \psi_{s, t-1}^1$ with $b=100$ & 83.6 $\pm$ 25.1 &  0\\¸
		& $F_\text{flow}$, $\psi_s$ with $b = 1000$ & 100.2 $\pm$ 19.5 & 13.1 $\pm$ 16.7\\
		& $\psi_{s, t}^1  = \psi_{s, t-1}^1$ with $b=1000$ & 127.1 $\pm$ 27.0 &  0 \\
		& $F_\text{flow}$, $\psi$ & 109.7 $\pm$ 18.3 & 13.5 $\pm$ 19.5 \\
		& $\psi_{t}^1  = \psi_{t-1}^1$ &  134.9 $\pm$ 25.7  &  0 \\
		\hline
	\end{tabular}
	\label{tab:pushingPredErrorComparison}
\end{table}

\subsubsection{Forward Prediction Error in Rigid Transformation Space}\label{sec:appendix:comparisonObjectRepresentations:resultsPushingPredictionQSpace}
As explained in sec.~\ref{sec:appendix:dynamicsModelRigidTransformation}, in order to compare models based on SDFs and occupancy measures with a model that has point-clouds as input, we show in Tab.~\ref{tab:pushingPredictionError_Q_comparison} the forward prediction error of the models predicting the rigid transformation $\Delta q^1$ of the object that is being pushed.

As one can see, the model learned based on SDFs outperforms the other representations.
The pointnet++ architecture achieves a similar error than the occupancy measure, pointnet has the highest error.

Furthermore, one can see in the second row of Tab.~\ref{tab:pushingPredictionError_Q_comparison} that the case $\phi(\cdot, I)$, i.e.\ where the SDFs are learned functions conditioned on image observations of the objects, does not degrade the performance.

The training, test and evaluation datasets are the same as in sec.~\ref{sec:appendix:comparisonObjectRepresentations:resultsPushingPrediction}, but contain the ground-truth rigid transformations additionally for training and error evaluation purposes.

\begin{table}
	\centering
	\begin{tabular}{ccccc}
		\hline
		& & $x$ [mm] & $y$ [mm] & $\alpha$ [$^\circ$]\\
		\hline
		\multirow{2}{1.5cm}{SDF} & $\phi$ & 2.1 $\pm$ 2.2 & 2.1 $\pm$ 1.3 & 0.63 $\pm$ 0.69\\
		& $\phi(\cdot;¸ I)$ (with image encoder) & 2.2 $\pm$ 2.3 & 2.1 $\pm$ 2.3 & 0.63 $\pm$ 0.63 \\
		\hline
		\multirow{3}{1.5cm}{occupancy measure} & $\psi_s$ with $b = 100$  & 2.7 $\pm$ 2.7 & 2.7 $\pm$ 2.7 & 0.76 $\pm$ 0.82 \\ 
		& $\psi_s$ with $b = 1000$ & 2.8 $\pm$ 2.7 & 2.7 $\pm$ 2.7 & 0.80 $\pm$ 0.87\\
		& $\psi$ & 2.8 $\pm$ 2.7 & 2.8 $\pm$ 2.7 & 0.81 $\pm$ 0.87 \\
		\hline
		\multirow{2}{1.5cm}{point-cloud} & pointnet & 3.3 $\pm$ 2.9 & 3.3 $\pm$ 3.0 & 0.88 $\pm$ 0.86 \\
		& pointnet++ & 2.7 $\pm$ 2.6 & 2.6 $\pm$ 2.5 & 0.79 $\pm$ 0.80 \\
		\hline
	\end{tabular}
	\caption{Comparison of one-step $\Delta q^1$ prediction error between models learned on SDF, occupancy measure and point-cloud object representations.}
	\label{tab:pushingPredictionError_Q_comparison}
\end{table}

\subsubsection{Planning}\label{sec:appendix:planningPushing}
Although being worse with respect to the one-step prediction error, the occupancy measure based model is qualitatively able to learn the dynamics of the pushing scenario.
However, when trying to utilize the occupancy measure based model within the optimization problem, we could not solve a single scene of the evaluation scene dataset.
This is caused by the fact that in the majority of the cases the optimization process got stuck in its initial condition directly, since the learned model as well as the other functionals based on the occupancy measure have zero or non-informative gradients, even in the smoothed version $\psi_s$.
Therefore, for using the learned model with a gradient-based planning framework, the information encoded in the SDF is crucial.
An SDF contains non-flat information about the object at distance to it, whereas the occupancy measure is flat at distance to the object.
Empirically, this property of the SDFs carries over to the models learned on top of them.

Similarly, planning with models learned on top of point-cloud encodings also could not solve a single scene of the evaluation scene dataset.

\begin{figure}
	\centering
	\subfloat[Initial configuration]{
		\includegraphics[trim={1cm 1cm 2cm 2cm}, clip, width=3.cm]{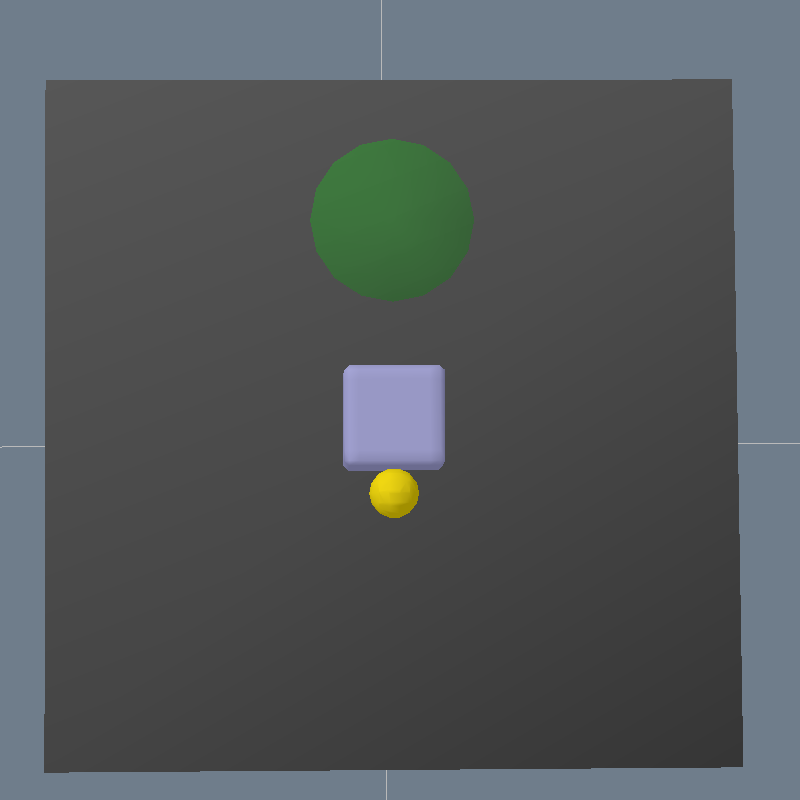}
		\label{fig:appendix:simpleScenario:initial}
	}\hfil
	\subfloat[Execution result with SDF]{
		\includegraphics[trim={1cm 1cm 2cm 2cm}, clip, width=3.cm]{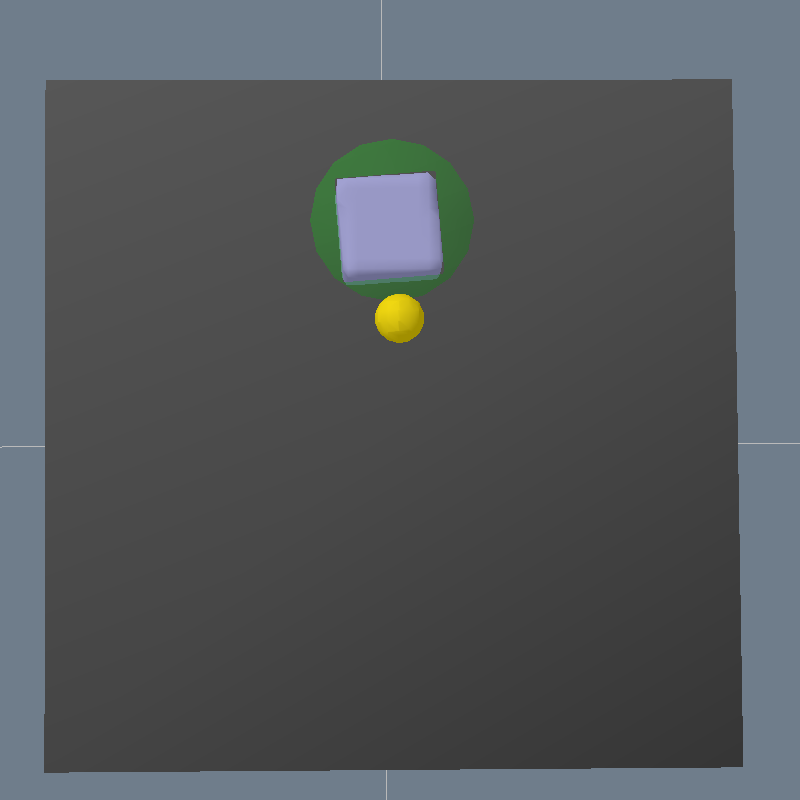}
		\label{fig:appendix:simpleScenario:SDF}
	}\hfil
	\subfloat[Execution result with point-cloud]{
		\includegraphics[trim={1cm 1cm 2cm 2cm}, clip, width=3.cm]{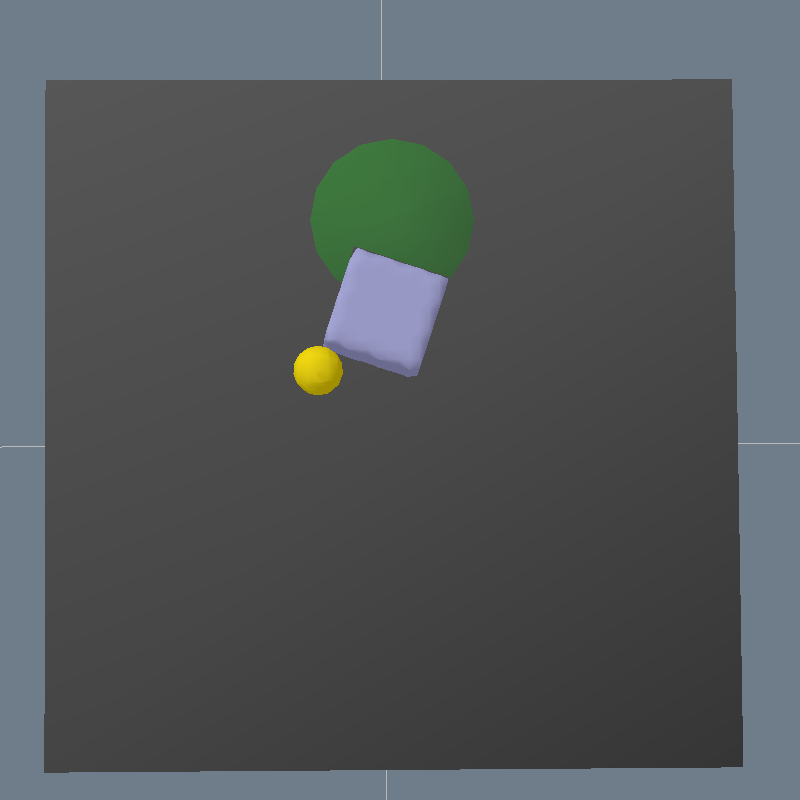}
		\label{fig:appendix:simpleScenario:PC}
	}\hfil
	\subfloat[Execution result with occupancy measure]{
		\includegraphics[trim={1cm 1cm 2cm 2cm}, clip, width=3.cm]{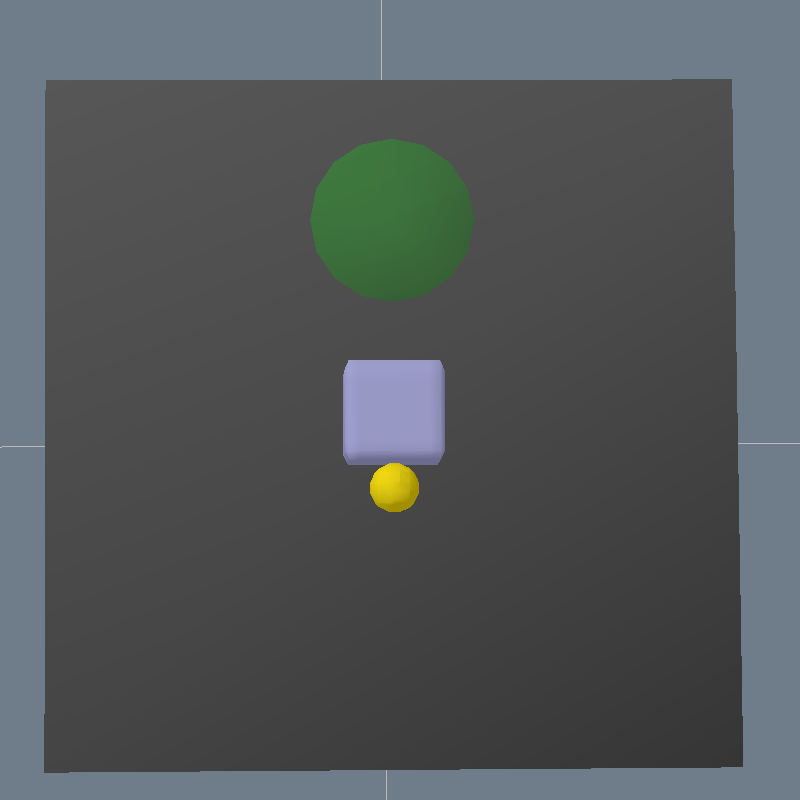}
		\label{fig:appendix:simpleScenario:occupancy}
	}
	\caption{Planning comparison for simple pushing scenario for models learned on top of SDF, occupancy measure and point-cloud object representations.}
	\label{fig:appendix:simpleScenario}
\end{figure}

To investigate this further, we consider a very simple scenario (see Fig.~\ref{fig:appendix:simpleScenario}) where the pusher already in the initial configuration has established contact with the object, as shown in Fig.~\ref{fig:appendix:simpleScenario:initial}. 
The object should just be pushed straight ahead to the goal.

Fig.~\ref{fig:appendix:simpleScenario:SDF}-\ref{fig:appendix:simpleScenario:occupancy} shows the execution result in the simulator of the trajectory the optimizer has converged to for this simple scenario.
Here, the only constraint of the optimization problem apart from the goal specification is $H_F$ (the dynamics model) and no other constraints like $H_\text{coll}$ or $H_\text{PoC}$.
As one can see, optimizing a trajectory with the occupancy measure based model does not move the pusher at all.
The point-cloud based model leads to some movement, but it is clearly unable to solve the task.
Using the SDF based model, planning is possible and hence solving this scene is no issue.

\subsection{Discussion}\label{sec:appendix:discussion}
For the mug-hanging scenario, the proposed approach of learning models based on SDF representations of the objects significantly outperforms the occupancy measure and point-cloud baselines.
The reasons for this are not only a higher prediction accuracy of the learned functional $H_\text{hang}$ with the SDF representation, but also that the model learned with SDFs provides more informative gradients for optimization.
Furthermore, an SDF representation allows to define the collision constraint (in a differentiable way) naturally.
For point-clouds, such collision constraints are not well-defined.
Generally, the results also highlight the importance of optimization (and sampling for initial guess), compared to sampling alone.

Regarding the pushing scenario, the SDF object representation enables to learn models that outperform all other considered object representations in terms of the forward prediction error.
Compared to a point-cloud representation, with SDFs (and occupancy measures) the learned dynamics model can directly predict in SDF (occupancy measure) space, requiring a dataset of such observations only, compared to ground-truth rigid object transformations as required for learning a model with point-clouds.
When it comes to planning, the (learned and analytic) functionals defined in terms of SDFs provide informative gradients, which is crucial for planning success.

This shows that model learning and subsequent planning has to be considered together.
Just from the fact that a model leads to acceptable prediction performance does not mean that it is useful at all for planning.

Further, we have also shown that one can learn an image conditioned SDF and the dynamics model based on the learned SDFs simultaneously with no noticeable performance degradation.

\section{Comparison to Analytic Mesh-Based Model for Pushing Scenario}\label{sec:appendix:analyticPushing}
In this experiment, we utilize the analytic model from \cite{20-toussaint-physicsLGP} that is based on mesh object representations to solve the same pushing scenarios as in Sec~\ref{sec:exp:pushing:evalPlanning}.
The problem formulation remains the same, i.e.\ we have the same constraints (collision avoidance and contact establishment), same initializations, same discrete decisions etc., with the only difference that the functionals are replaced by mesh-based representations and the learned SDF dynamics model with an analytic one.
Since it is unclear how to specify the same goal region constraint from sec.~\ref{sec:TaskFuncs:goalRegion} with a mesh, we utilize the same goal-region SDF constraint \eqref{eq:goalRegion} in this experiment, i.e.\ everything is based on meshes except the goal specification, which utilizes the same SDF as for the other experiments.
These choices allow for a fair comparison.

\begin{figure}
	\if\generatePlots1
		\input{plots/compMeshSDF.tex}
	\else
		\includegraphics{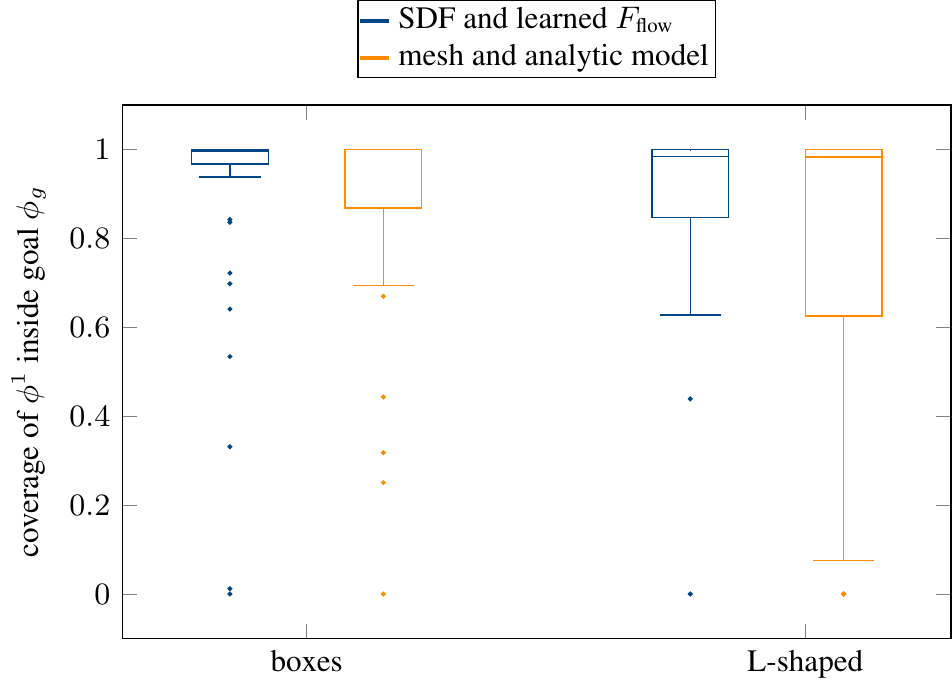}
	\fi
	\caption{Performance comparison between proposed SDF approach with learned models and analytic mesh-based models for pushing scenario on evaluation scenes. A value of 1 means that the object is fully contained in the goal region.}
	\label{fig:appendix:compSDFMesh}
\end{figure}

Fig.~\ref{fig:appendix:compSDFMesh} shows the performance in terms of the amount of object $\phi^1$ that is inside the goal region at the end of the execution both for the proposed SDF approach ($F_\text{flow}$) and the analytic mesh-based approach.
As one can see, especially for the L-shaped objects, our proposed method outperforms the mesh-based analytic variant significantly, but also for boxes the performance with the SDF models is better.

Possible reasons for this are mainly two fold.
On the one hand, the mismatch between the analytic model and the simulator leads to the object rotating more in the optimized trajectory than when executed in the simulator.
System identification (including potentially a more complex friction model, which makes planning much harder) for each object shape would be required to improve on this.
On the other hand, the optimized trajectories obtained with the mesh models very often push on a corner of the objects.
Such corner pushes are unstable when executed.
We have seen less such behavior of pushing at corners when using SDFs and the learned models on top of them.
A possible explanation for this is that the data generation for the learned model is performed via random pushes, which leads to only little data where there are pushes on corners.

As mentioned in sec.~\ref{sec:exp:pushing:evalPlanning} and sec.~\ref{sec:appendix:pushingInitialization}, directly solving \eqref{eq:opt} on the pushing scenario often leads the optimizer to converge to an infeasible local minimum if the pusher has to go around the object to achieve the goal.
This is not a fact that is caused by using learned models or SDF object representations.
The same holds true for the analytic model and mesh-based object representations.
Therefore, we also require the initializations described in sec.~\ref{sec:appendix:pushingInitialization} for this experiment with the analytic, mesh-based models.
Similar to the other experiments, these 4 rough initializations address the issue successfully.

Note that for the pushing scenario it is possible to write down an analytic physics-based model on the mesh representation that is suitable for planning.
However, for the hanging scenario, this is not directly possible.
In \cite{pfaff2020learning}, a mesh-based dynamic model is proposed, but there the focus is on learning a passive simulator and not learning models suitable for planning.

\section{Generalization Experiments for Pushing Scenario}
\subsection{Generalization to Out-of-Distribution Shapes}\label{sec:appendix:pushingOutOfDistribution}
The training data for the pushing scenario consists of boxes and L-shaped objects of different sizes.
The test and evaluation scenes use different random seeds, but sample the object shapes from the same distribution.
Since the learned model takes the actual geometry of the objects as input, it could, in principle, generalize to arbitrary shapes.
This experiment investigates whether this generalization works.

\begin{figure}
	\centering
	\includegraphics[trim={10cm 9cm 9cm 10cm}, clip, width=3.3cm]{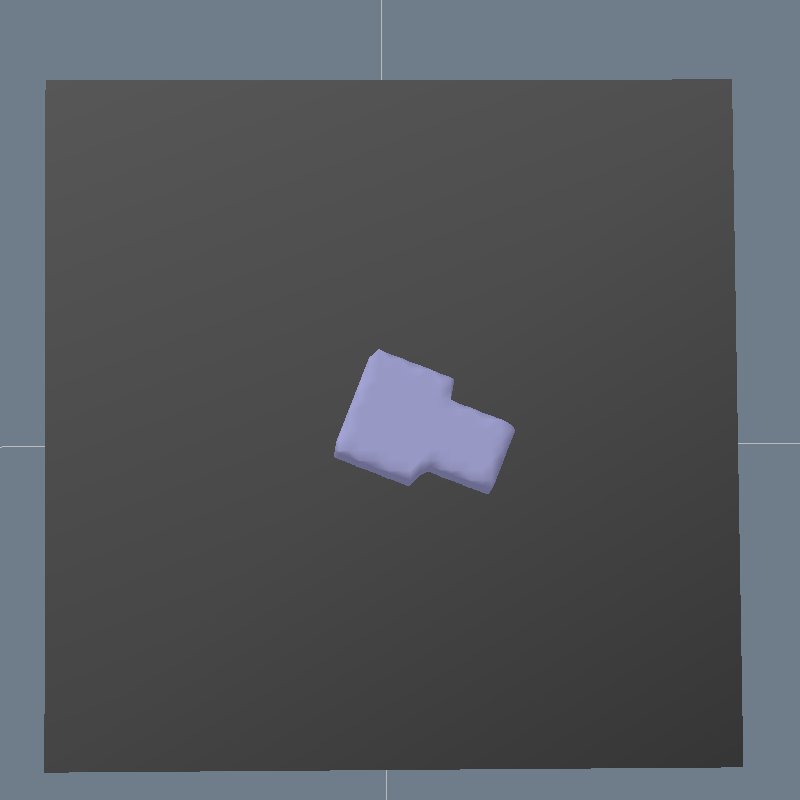}
	\includegraphics[trim={10cm 9cm 9cm 10cm}, clip, width=3.3cm]{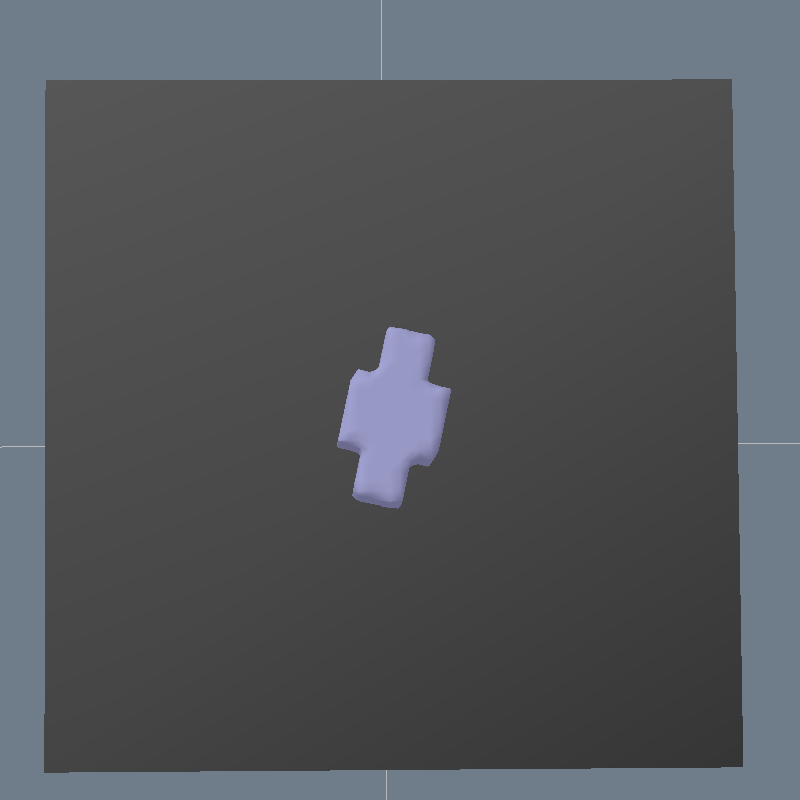}
	\includegraphics[trim={10cm 9cm 9cm 10cm}, clip, width=3.3cm]{images/pushing/g3}
	\includegraphics[trim={10cm 9cm 9cm 10cm}, clip, width=3.3cm]{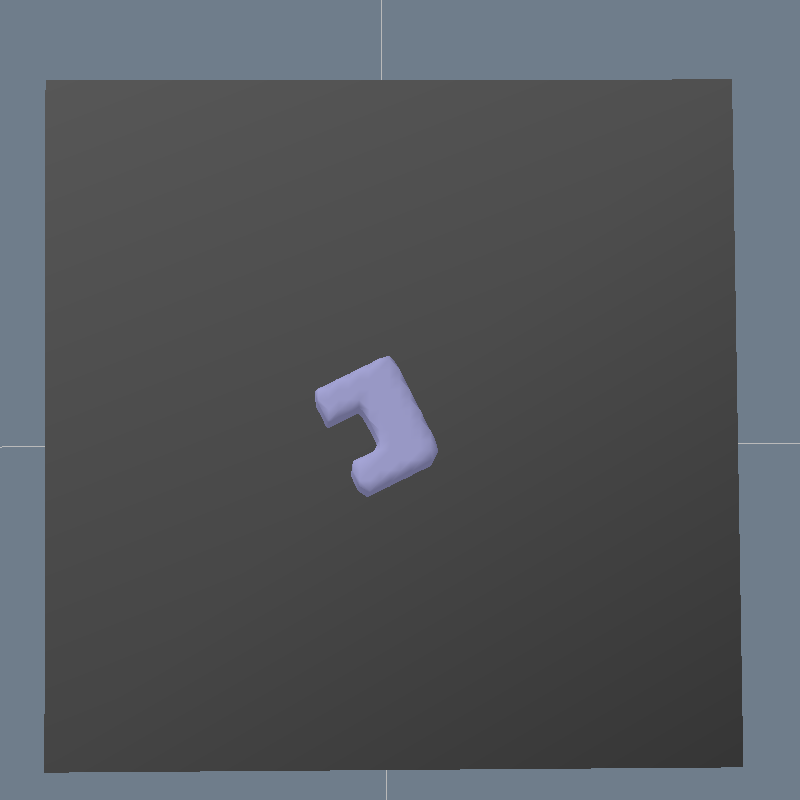}\\[0.1cm]
	\includegraphics[trim={8cm 9cm 10cm 9cm}, clip, width=3.3cm]{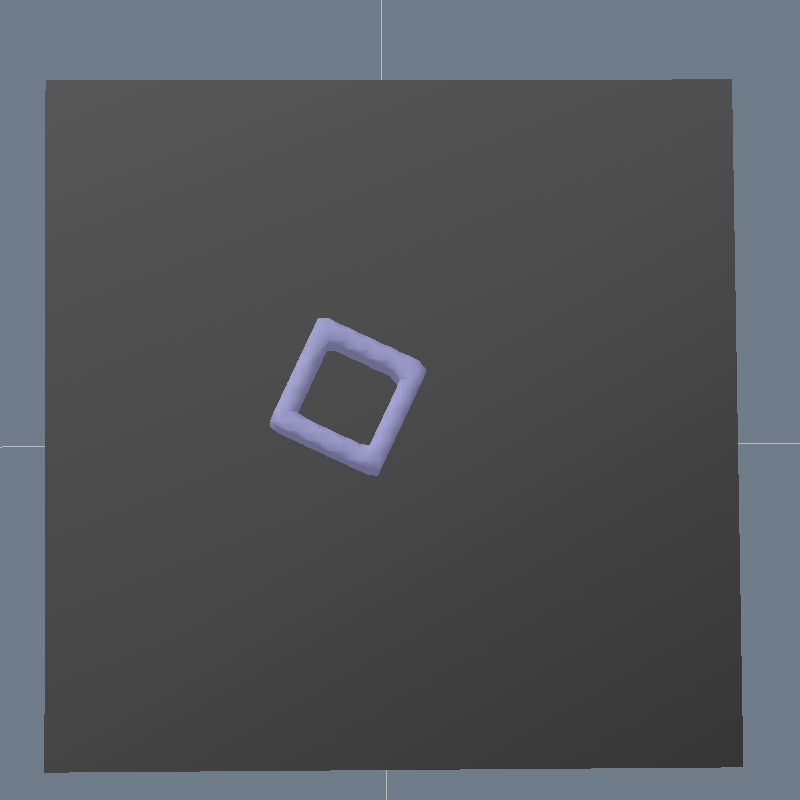}
	\includegraphics[trim={10cm 9cm 9cm 10cm}, clip, width=3.3cm]{images/pushing/g6}
	\includegraphics[trim={10cm 9cm 9cm 10cm}, clip, width=3.3cm]{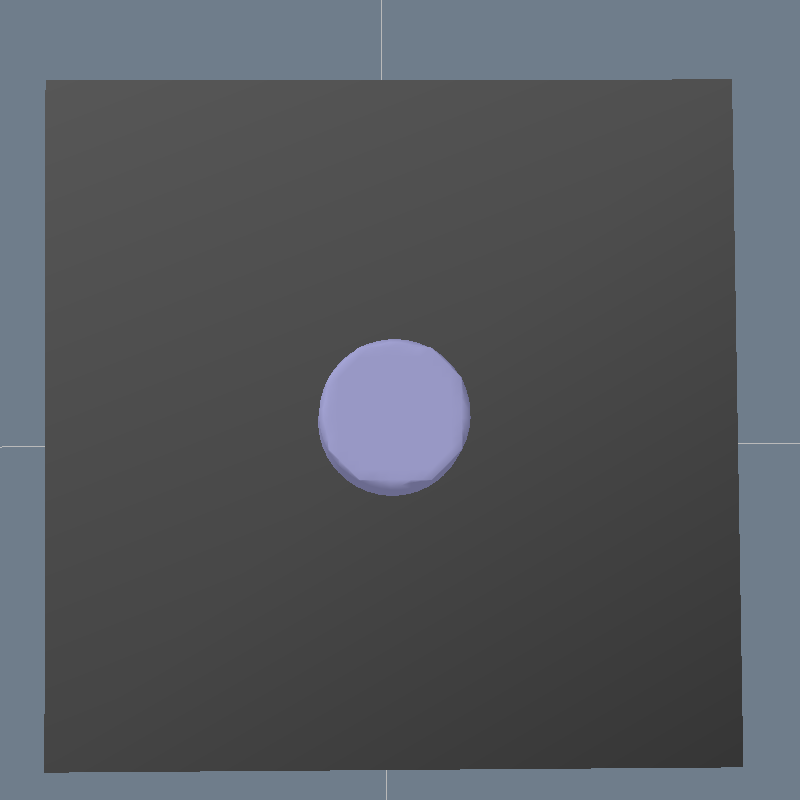}
	\caption{Tested object shapes for out-of-distribution shape generalization experiment for pushing scenario.}
	\label{fig:exp:pushing:generalizationShapes}
\end{figure}

In order to do so, we consider 7 different objects, shown in Fig.~\ref{fig:exp:pushing:generalizationShapes}, which are clearly out-of-distribution compared to the training dataset in terms of shape type and topology. 

\subsubsection{Forward Prediction Error}
Tab.~\ref{tab:shapeGeneralizationPushingPredictionError_Q_comparison} shows the forward prediction error in $\Delta q^1$ space for the out-of-distribution shapes.
To obtain these results, we apply random pushes on the out-of-distribution shapes with pushers of different sizes in the simulator and record the ground-truth rigid transformations $\Delta q^1$ of the pushed object.

\begin{table}
	\centering
	\begin{tabular}{ccccc}
		\hline
		& & $x$ [mm] & $y$ [mm] & $\alpha$ [$^\circ$]\\
		\hline
		\multirow{2}{1.5cm}{SDF} & $\phi$ & 2.3 $\pm$ 2.2 & 2.3 $\pm$ 2.3 & 0.81 $\pm$  0.87\\
		& $\phi(\cdot;¸ I)$ (with image encoder) & 2.4 $\pm$ 2.3 & 2.4 $\pm$ 2.4 & 0.85 $\pm$ 0.92 \\
		\hline
		\multirow{3}{1.5cm}{occupancy measure} & $\psi_s$ with $b = 100$ & 3.0  $\pm$ 3.2 & 3.2 $\pm$ 3.3 & 0.99 $\pm$ 1.06 \\
		& $\psi_s$ with $b = 1000$ & 3.2 $\pm$ 3.5 & 3.3 $\pm$ 3.5 & 1.02 $\pm$ 1.06 \\
		& $\psi$ & 3.0 $\pm$ 3.3 & 3.4 $\pm$ 3.6 & 1.06 $\pm$ 1.09 \\
		\hline
		\multirow{2}{1.5cm}{point-cloud} & pointnet & 4.9 $\pm$ 4.5 & 5.1 $\pm$ 4.7 & 0.99 $\pm$ 0.97 \\
		& pointnet++ & 4.5 $\pm$ 4.9 & 4.6 $\pm$ 4.8 & 0.77 $\pm$ 0.78 \\
		\hline
	\end{tabular}
	\caption{Shape generalization experiment. Comparison of $\Delta q^1$ prediction error between models learned on SDF, occupancy measure and point-cloud object representations.}
	\label{tab:shapeGeneralizationPushingPredictionError_Q_comparison}
\end{table}

In terms of absolute numbers, the SDF based model significantly outperforms the other object representations.

Comparing these results with the forward prediction error evaluated on the on-distribution evaluation dataset (see Tab.~\ref{tab:pushingPredictionError_Q_comparison}), one can see that the one-step forward prediction error in $x,y$-transformation increases by only 9.5\% for the proposed model learned on top of SDF object representations (9\% with the learned image conditioned SDF), compared to 48.5\% for a model learned with pointnet (66.7\% increase for pointnet++).
For the occupancy measure, the increase is about 14\% (depending on the exact variant), which is not as much as for the point-cloud based models, but also more than with SDFs.

This shows that models learned in SDF space generalize better (and well in terms of absolute numbers) to out-of-distribution shapes.

\subsubsection{Planning and Execution Performance}
\begin{figure}
	\if\generatePlots1
		\input{plots/outOfDistribution}
	\else
		\includegraphics{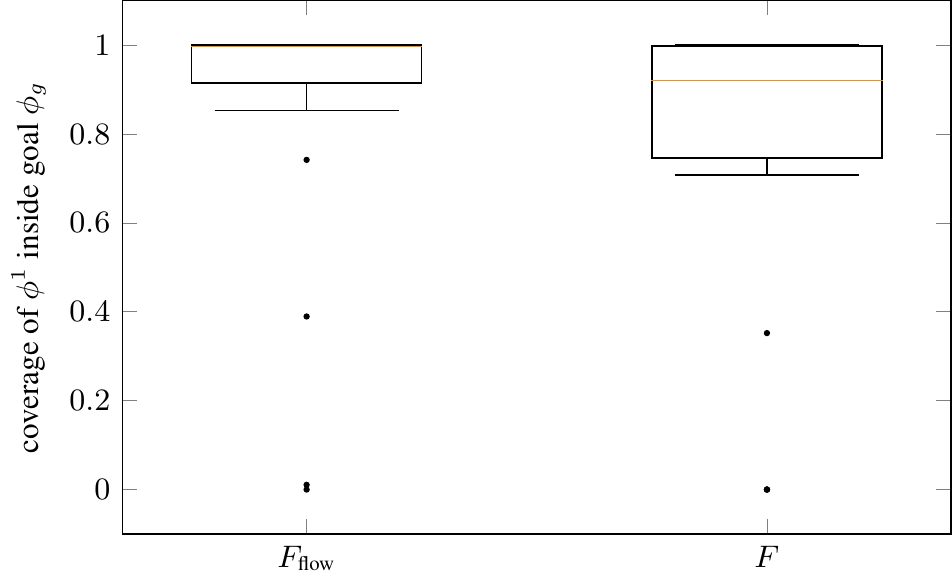}
	\fi
	\caption{Pushing performance on scenes containing out-of-distribution shapes in terms of the amount of $\phi^1$ that is inside of the goal region at the end of the execution. A value of 1 means that the object is fully contained in the goal region.}
	\label{fig:appendix:pushingGeneralizationShapesPerformance}
\end{figure}

Fig.~\ref{fig:appendix:pushingGeneralizationShapesPerformance} shows the performance when executing the planning results open-loop in the simulator for each of the 7 shapes shown in Fig.~\ref{fig:exp:pushing:generalizationShapes} tested in 3 different environments (i.e.\ different start configurations and goal region locations), leading to 21 experiments in total.
No changes in the methodology were required to achieve these results.
The initialization options (cf.\ sec.~\ref{sec:appendix:pushingInitialization}) are also the exact same as for the other experiments.

With a median coverage of 99.7\% of the amount of $\phi^1$ that is inside of the goal region at the end of the execution with the learned $F_\text{flow}$, the model generalizes very well to out-of-distribution scenes.  
Not surprisingly, the median coverage with the direct $F$ of 92.0\% is lower, since there we ask $F$ to not only predict the change of the input SDF $\phi^1_{t-1}$ to $\phi^1_t$, but the whole $\phi^1_t$ directly, which is more challenging especially for out-of-distribution shapes.

This experiment shows that the learned model is capable of generalizing quite out-of-distribution not only with respect to its prediction error, but also when utilized within the planning framework.

\subsection{Generalization to three Interacting Objects}\label{sec:appendix:threeObjects}
\begin{figure}
	\centering
	\includegraphics[trim={1cm 1cm 2cm 2cm}, clip, width=3.3cm]{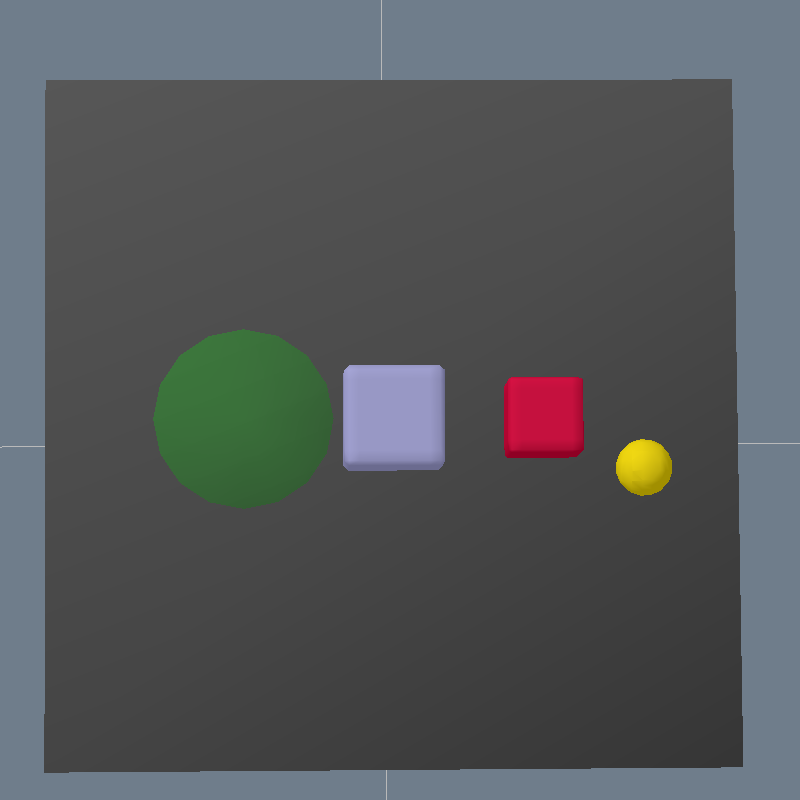}
	\includegraphics[trim={1cm 1cm 2cm 2cm}, clip, width=3.3cm]{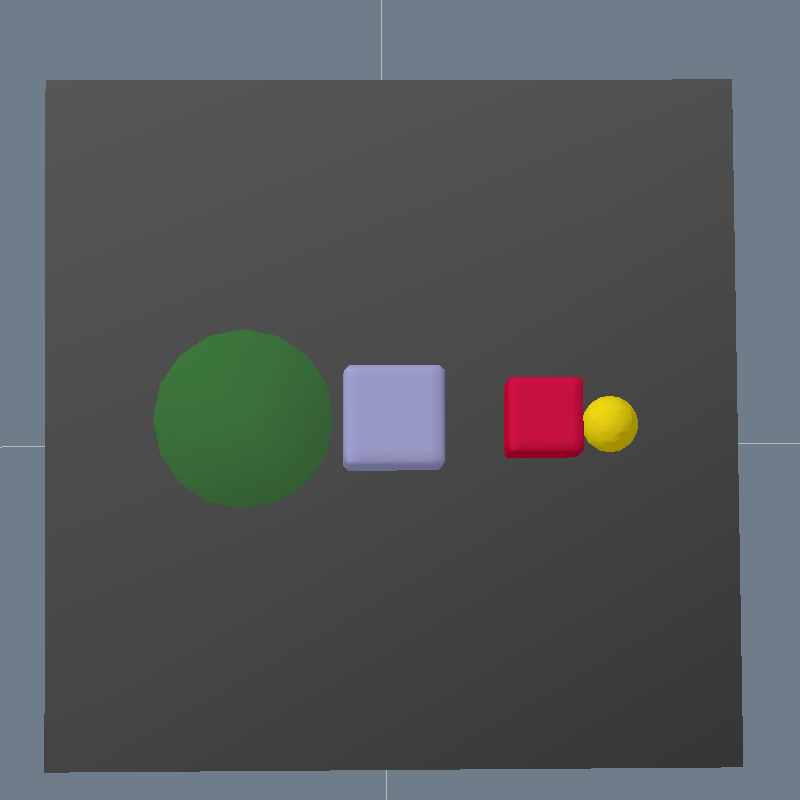}
	\includegraphics[trim={1cm 1cm 2cm 2cm}, clip, width=3.3cm]{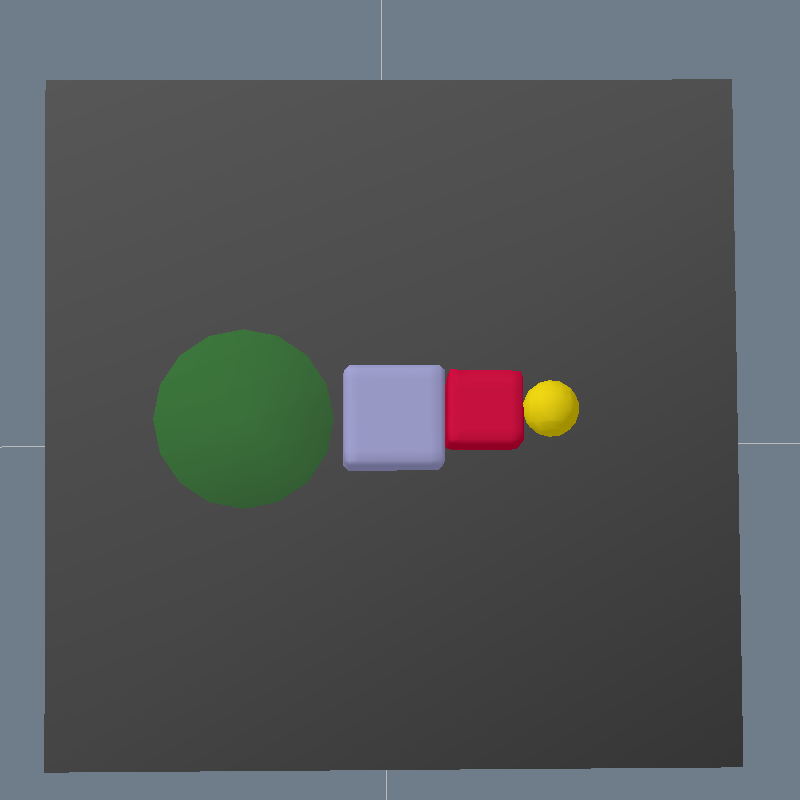}
	\includegraphics[trim={1cm 1cm 2cm 2cm}, clip, width=3.3cm]{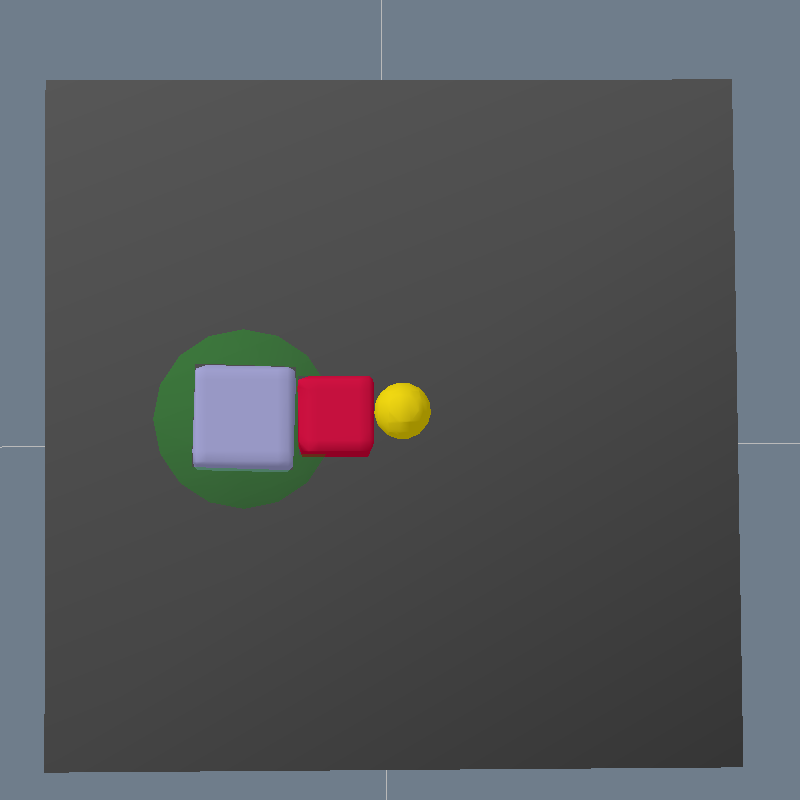}
	\caption{Generalization experiment to three objects that have to interact to solve the task. The yellow pusher is moved such that it pushes the red box such that the red box pushes the blue object to the green goal region. No new model has been learned for this experiment, it can be solved with the model trained on the interaction between two objects.}
	\label{fig:appendix:threeObjects}
\end{figure}

The pushing dynamics model we propose and learn is object-centric, i.e.\ it models the interaction between two objects.
Compared to other approaches that model the dynamics on a scene level as in \cite{xu2020learning}, we can train on only two objects and then generalize to situations where not only more objects are in the scene, but also multiple objects have to interact to solve the task, without having to relearn a new model.
In Fig.~\ref{fig:appendix:threeObjects} we show such a scenario where three objects have to interact to solve the task.

The same learned dynamics model constraint $H_F$ is active here twice, once between the yellow sphere and the red box for the whole length of the trajectory.
Second between the red box and the blue box for the last phase of the trajectory.
Note that we here ask for additional generalization, since in the training data only spheres have been used as the pusher object, but now the model has to predict the dynamics for two boxes that interact.

While the capabilities of the framework are important to be applied to scenes with more objects than being trained on, a case where two objects push one other object simultaneously cannot be realized directly with a model learned on only pair-interactions.
However, our general formalism could handle this case as well with a model of the form
$\phi^1_t = F(\phi^1_{t-1}, \phi^2_{t-1}, \phi^2_{t}, \phi^3_{t-1}, \phi^3_{t})$, which would require relearning such a model.
Nevertheless, this experiment again shows the advantages of our framework of both being object-centric and embedded into an optimization problem.

\subsection{Generalization to Scenes with Obstacles}\label{sec:appendix:obstacles}
\begin{figure}
	\centering
	\includegraphics[trim={1cm 1cm 2cm 2cm}, clip, width=3.3cm]{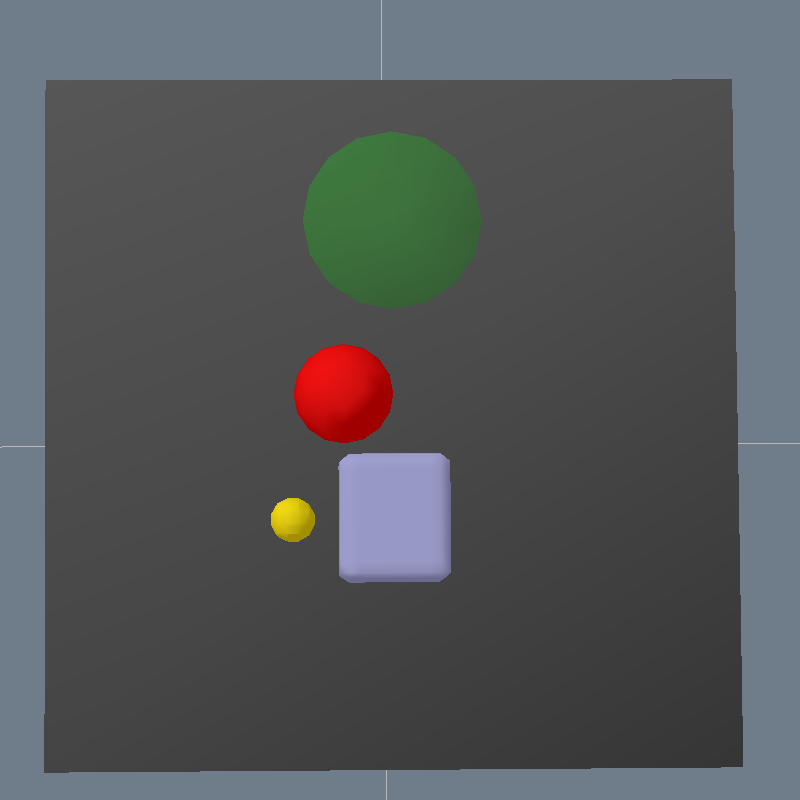}
	\includegraphics[trim={1cm 1cm 2cm 2cm}, clip, width=3.3cm]{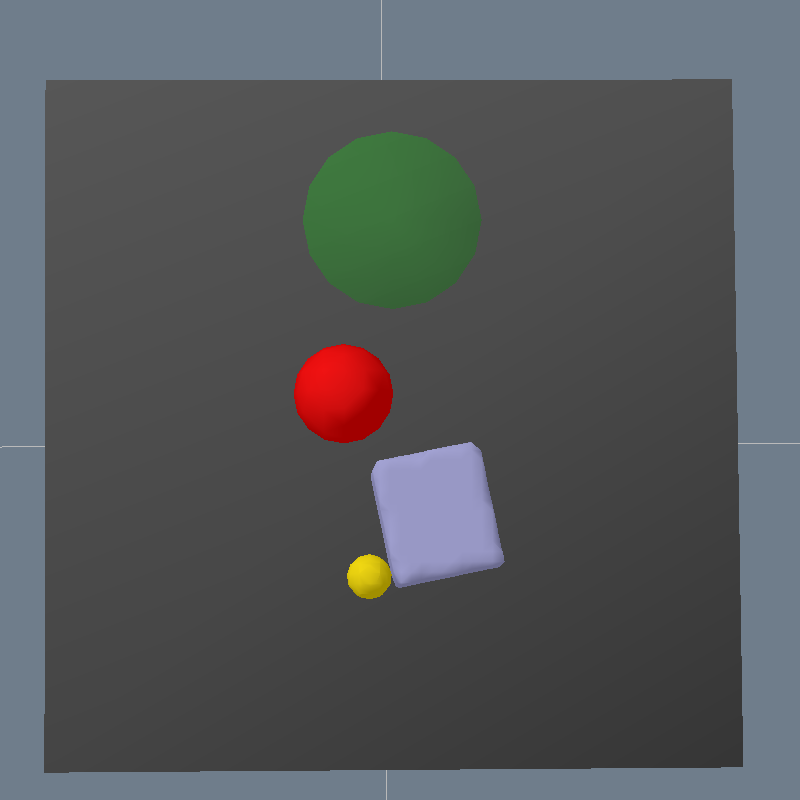}
	\includegraphics[trim={1cm 1cm 2cm 2cm}, clip, width=3.3cm]{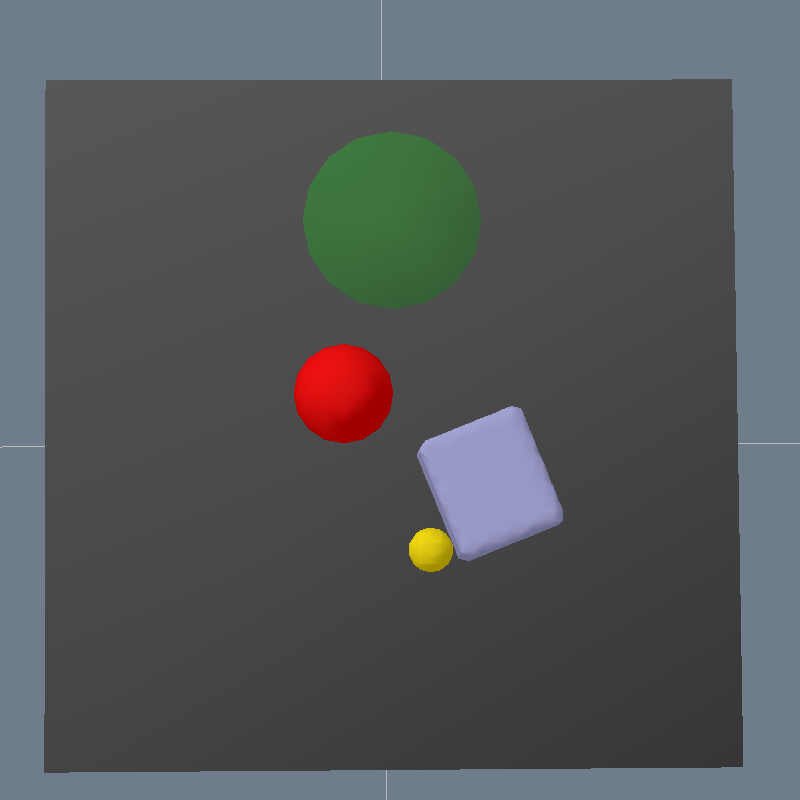}
	\includegraphics[trim={1cm 1cm 2cm 2cm}, clip, width=3.3cm]{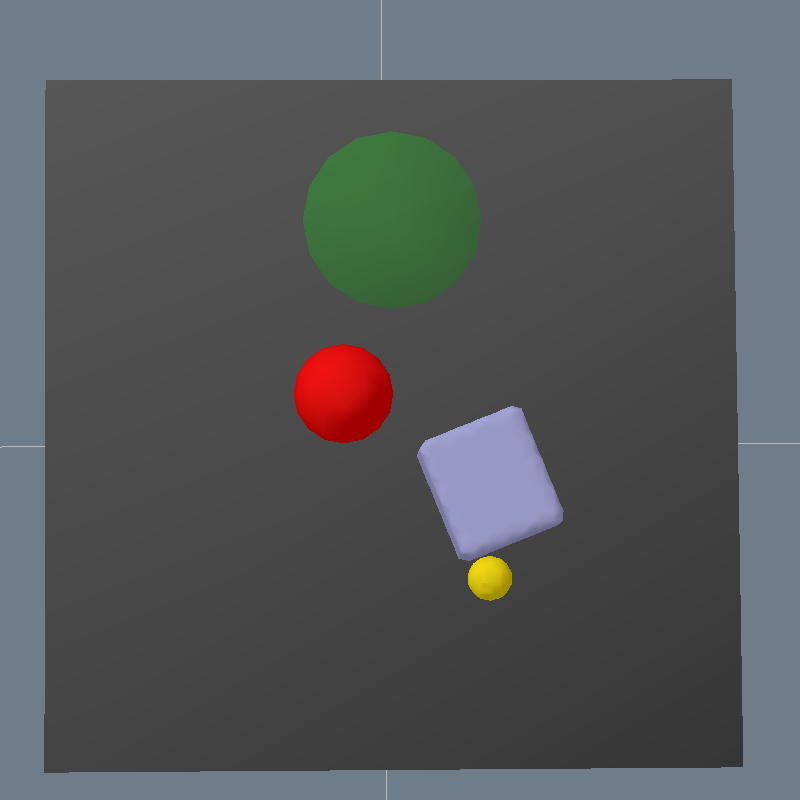}
	\includegraphics[trim={1cm 1cm 2cm 2cm}, clip, width=3.3cm]{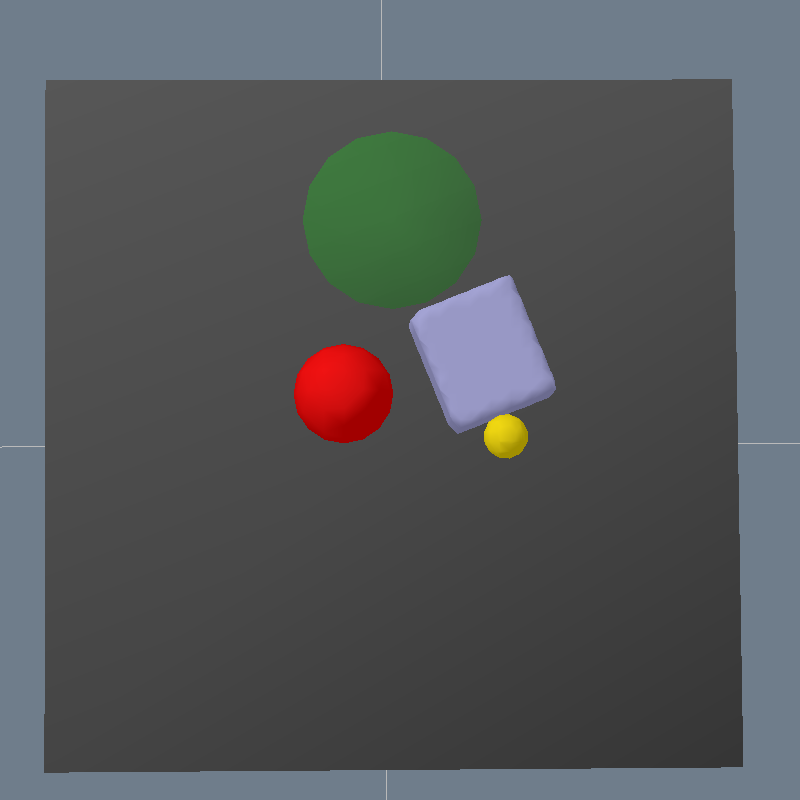}
	\includegraphics[trim={1cm 1cm 2cm 2cm}, clip, width=3.3cm]{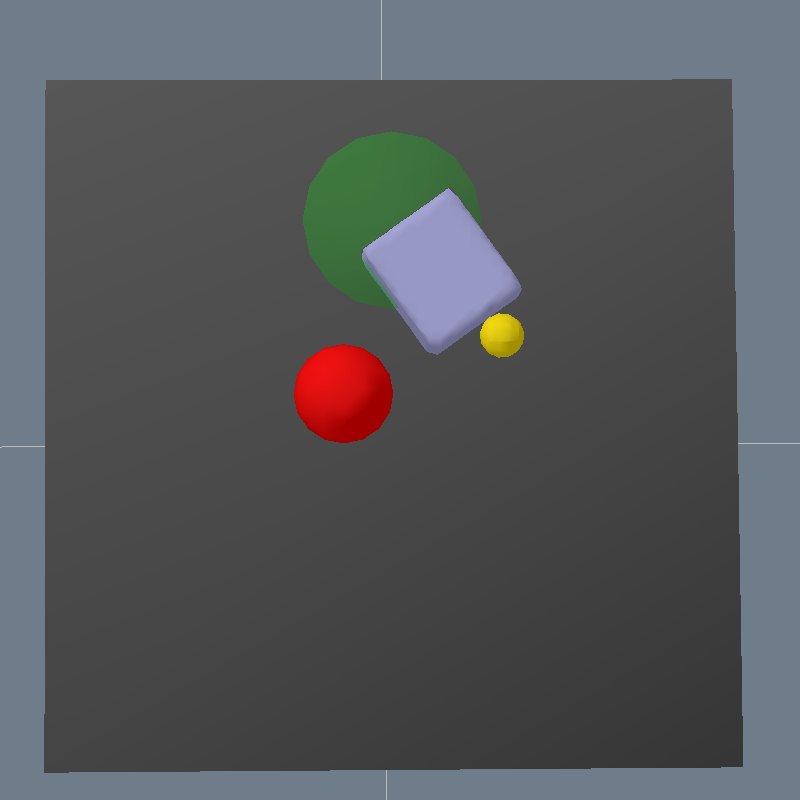}
	\includegraphics[trim={1cm 1cm 2cm 2cm}, clip, width=3.3cm]{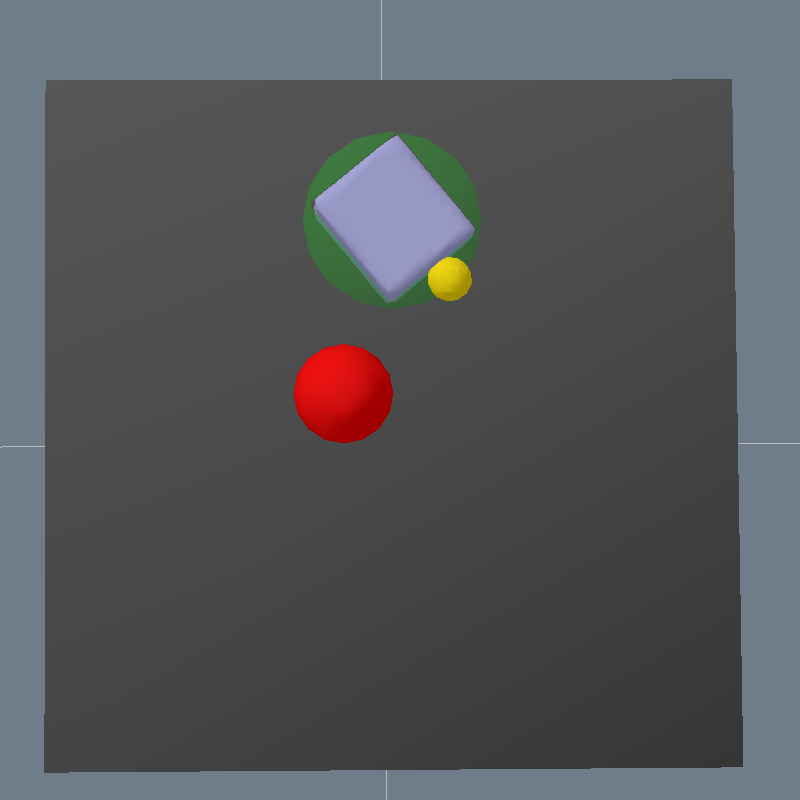}
	\caption{Generalization experiment to a scene that contains an obstacle (red). Three push phases have been determined to solve the task.}
	\label{fig:appendix:obstacle}
\end{figure}

As mentioned in the last section, due to the (learned) functionals being object-centric, we can apply a model that has been trained on scenes containing only two objects to scenarios with more objects.
Fig.~\ref{fig:appendix:obstacle} shows a scenario where there is an obstacle in the scene between the object that should be pushed and the goal region.
The optimization problem then tries to find a pushing trajectory consistent with the learned dynamics model while trying to avoid collision with the obstacle.
The found solution for the scene in Fig.~\ref{fig:appendix:obstacle} consists of three push phases.
There are pair-wise collision constraint functionals $H_\text{coll}$ between the objects in the scene.

Scaling this to scenes that contain many obstacles, the non-convexity of collision avoidance becomes an issue.
However, this is \emph{not} primarily caused by the fact that we represent objects as SDFs and that we plan with learned models on top of them. 
The same issue applies for mesh-based representations and analytic dynamic models.

\section{Ablation Study for Pushing Scenario}\label{sec:appendix:ablationPushing}
\begin{figure}
	\if\generatePlots1
		\input{plots/ablation}
	\else
		\includegraphics{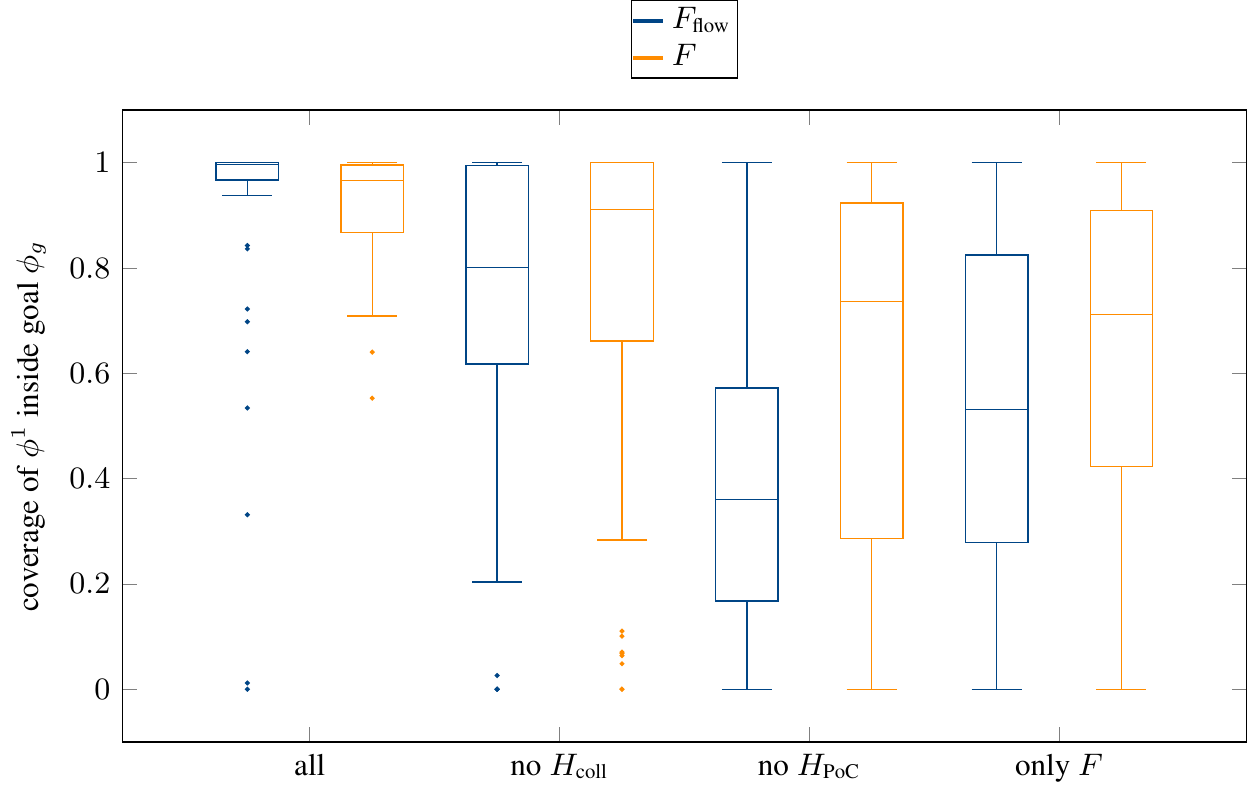}
	\fi
	\caption{Ablation study for pushing experiment: removal of additional constraints $H_\text{coll}$ and $H_\text{PoC}$.}
	\label{fig:pushingAblation}
\end{figure}

In sec.~\ref{sec:exp:pushing:evalPlanning} and sec.~\ref{sec:appendix:additionalDetailsPushing} we describe the additional objectives $H_\text{PoC}$ to encourage the trajectory to establish contact and $H_\text{coll}$ to avoid collisions (cf.\ sec.~\ref{sec:TaskFuncs} and sec.~\ref{sec:appendix:contactEstablishment}, too).
Here we investigate the importance of these, also in comparison with using $F$ and $H_g$ as the only constraints.
Fig.~\ref{fig:pushingAblation} shows the performance on a dataset of 20 box pushing scenarios, both for $F$ (orange) and $F_\text{flow}$ (blue), where we remove $H_\text{coll}$, $H_\text{PoC}$ or both (only $F$).
The main reason for the significantly decreased performance when not using $H_\text{PoC}$ (only $F$, no $H_\text{PoC}$) is that, in some cases, the pusher is not moved at all by the optimization problem.
A forward model alone, or more precisely its gradients, simply does not contain enough information for long-horizon tasks to succeed.
Therefore, $H_\text{PoC}$ helps the optimization problem to establish contact, where then the model locally provides sufficient information to solve the task.
$H_\text{PoC}$ only models that there should be contact, and the exact contact locations are then subject to the model.
However, there is still a large number of cases where planning/execution works even without $H_\text{PoC}$, since the SDF based dynamics model $F$, compared to point-cloud or occupancy measure based models, still provides somehow useful gradients, even without the help of $H_\text{PoC}$.
When removing the collision constraint, the performance also drops.
This is caused mainly by the fact that the optimized trajectory without the collision constraint sometimes is in collision for a very short moment, which then leads to a failed open-loop execution.

\section{Pushing Experiment with (Multiple) Robots}\label{sec:appendix:robots}
A major advantage of our proposed framework compared to other approaches is that the class of models we propose is object-centric, i.e.\ they model the interaction between the pusher and the object directly, instead of abstract actions.
This allows us to directly integrate the model in optimization problems for more complex scenarios, for example where robots should push or otherwise interact with the objects, although the training process did not involve any robots.
The learned model $F$ acts as a constraint $H_F$ on the possible object/pusher trajectories during the phase of the motion where the constraint is active.

\begin{figure}
	\centering
	\includegraphics[trim={0cm 0cm 0cm 0cm}, clip, width=4cm]{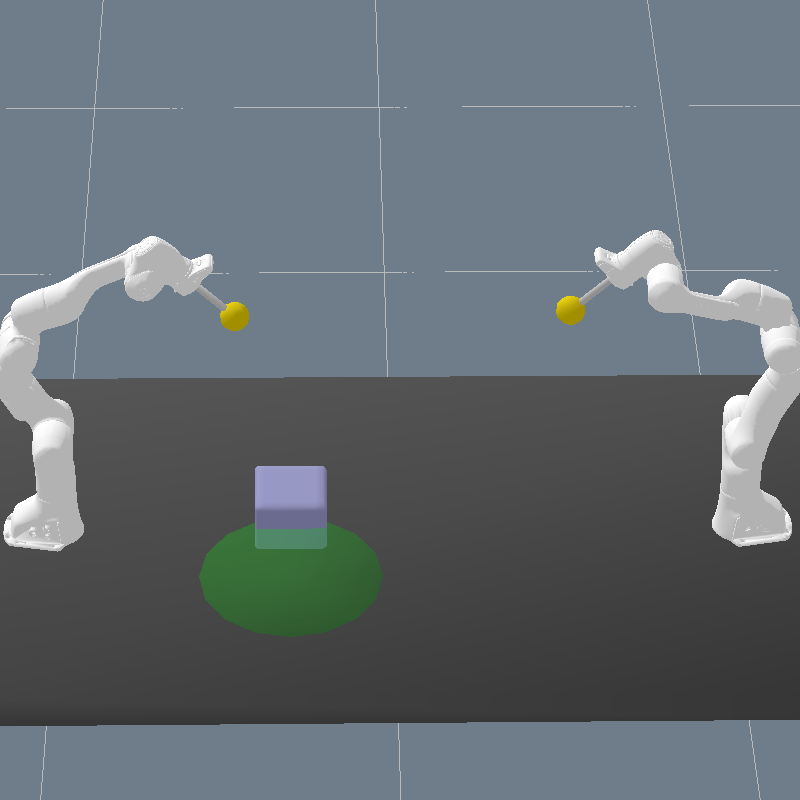}
	\includegraphics[trim={0cm 0cm 0cm 0cm}, clip, width=4cm]{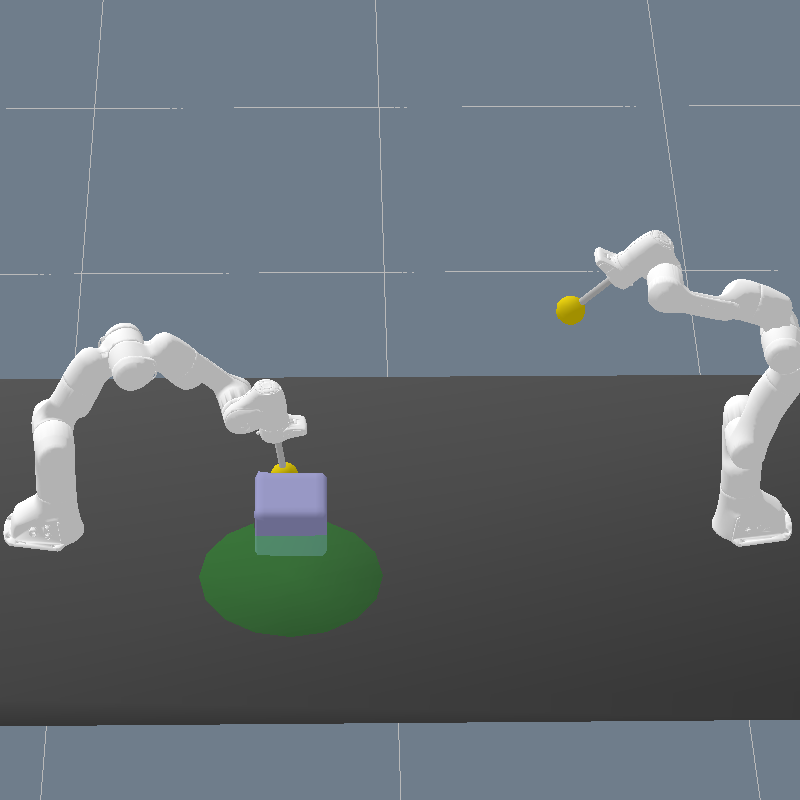}
	\includegraphics[trim={0cm 0cm 0cm 0cm}, clip, width=4cm]{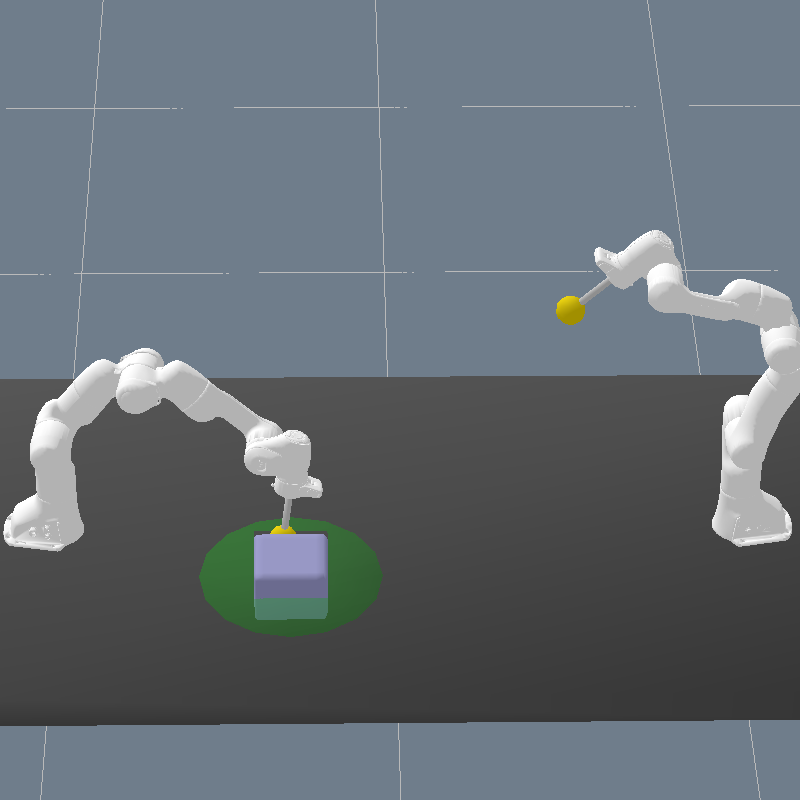}
	\caption{Pushing scenario with robots. Here the goal region (green) is located such that the left robot arm can push the object to the goal region.}
	\label{fig:appendix:oneRobot}
\end{figure}

\begin{figure}
	\centering
	\includegraphics[trim={0cm 0cm 0cm 0cm}, clip, width=4cm]{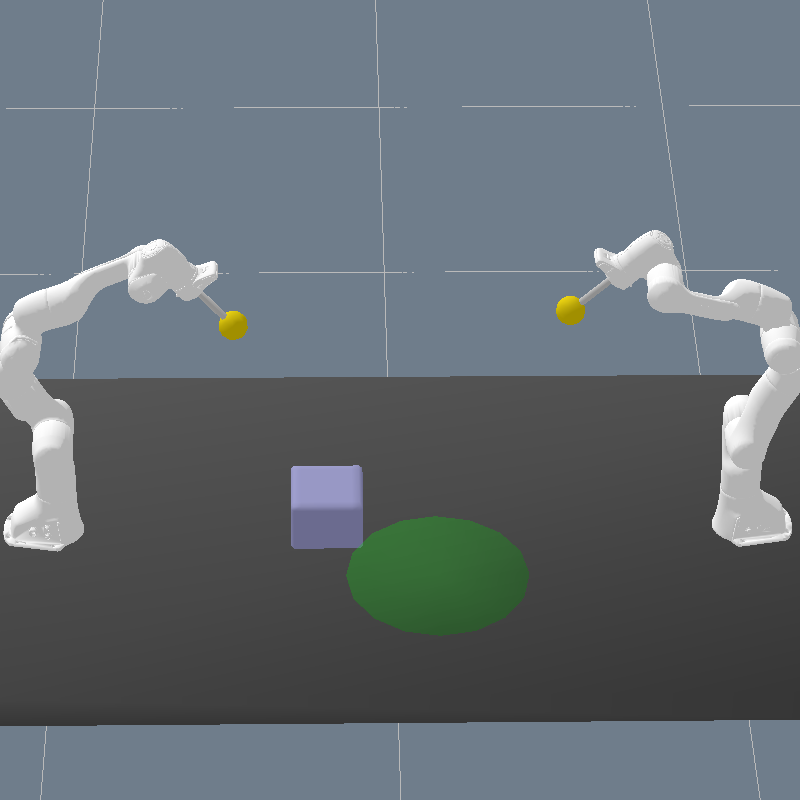}
	\includegraphics[trim={0cm 0cm 0cm 0cm}, clip, width=4cm]{images/pushing/twoRobots_2}
	\includegraphics[trim={0cm 0cm 0cm 0cm}, clip, width=4cm]{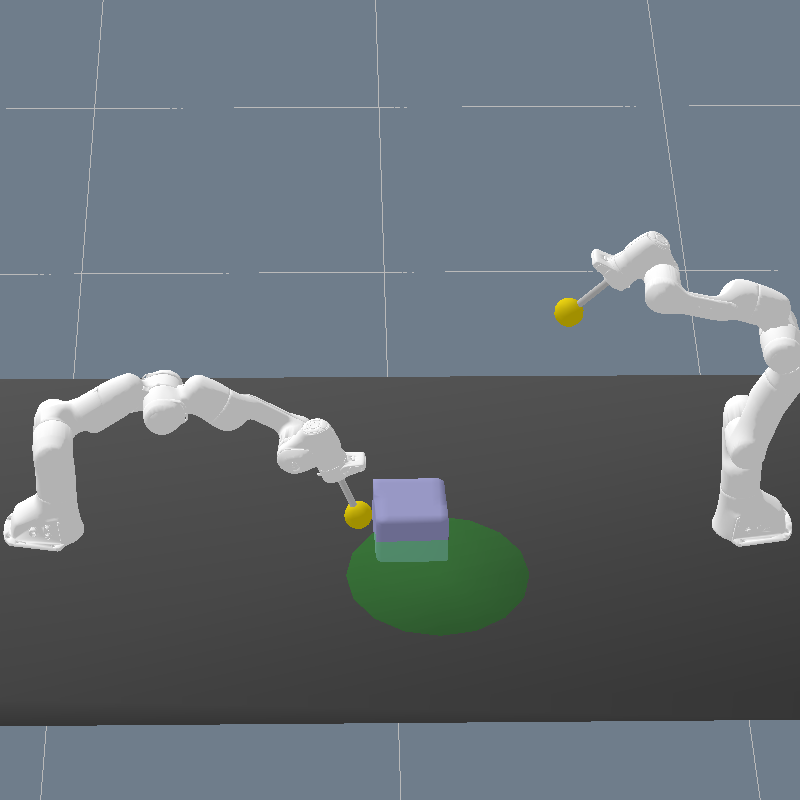}\\[0.2cm]
	\includegraphics[trim={0cm 0cm 0cm 0cm}, clip, width=4cm]{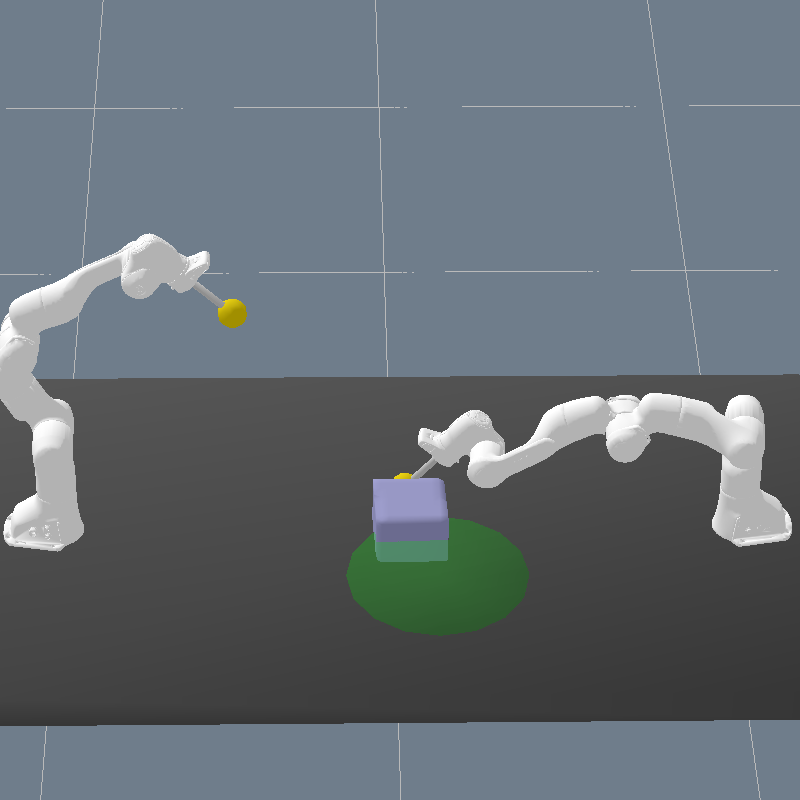}
	\includegraphics[trim={0cm 0cm 0cm 0cm}, clip, width=4cm]{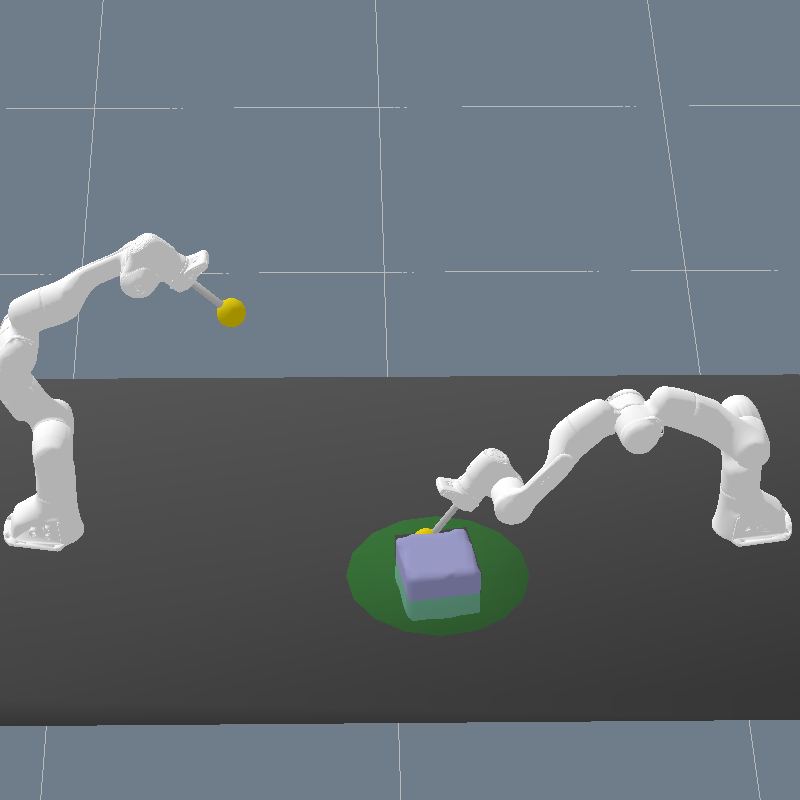}
	\caption{Pushing scenario with robots. Here the goal region (green) is located such that the left robot arm cannot push the object to the goal region. Hence, our framework finds a solution where both robot arms collaborate to push the object into the goal region. The movements of the robot arms are optimized jointly.}
	\label{fig:appendix:twoRobots}
\end{figure}

In Fig.~\ref{fig:appendix:oneRobot} and Fig.~\ref{fig:appendix:twoRobots}, we show two different scenarios where two robots are in the scene.
At the end-effectors of each robot a sphere is mounted, which serves as the pusher object.
The goal is to push the light blue object to two different goal regions (green).
In Fig.~\ref{fig:appendix:oneRobot}, the goal region is located such that the left robot arm can push the object directly into the goal region.
In contrast, for the scenario shown in Fig.~\ref{fig:appendix:twoRobots}, the left robot arm cannot push the object into the goal region, since its kinematic limits do not allow for the necessary movements.
Therefore, our framework finds a solution where both robot arms are involved in the pushing maneuver. 
There is no intermediate goal location specified or similar, the whole motions of the robots are optimized jointly.
The discrete decisions of the optimization problem \eqref{eq:opt} now involve which robot arm to use at which phase of the motion, which implies when and between which objects the dynamic model constraint is active.

\begin{figure}
	\centering
	\includegraphics[trim={0cm 0cm 0cm 0cm}, clip, width=4cm]{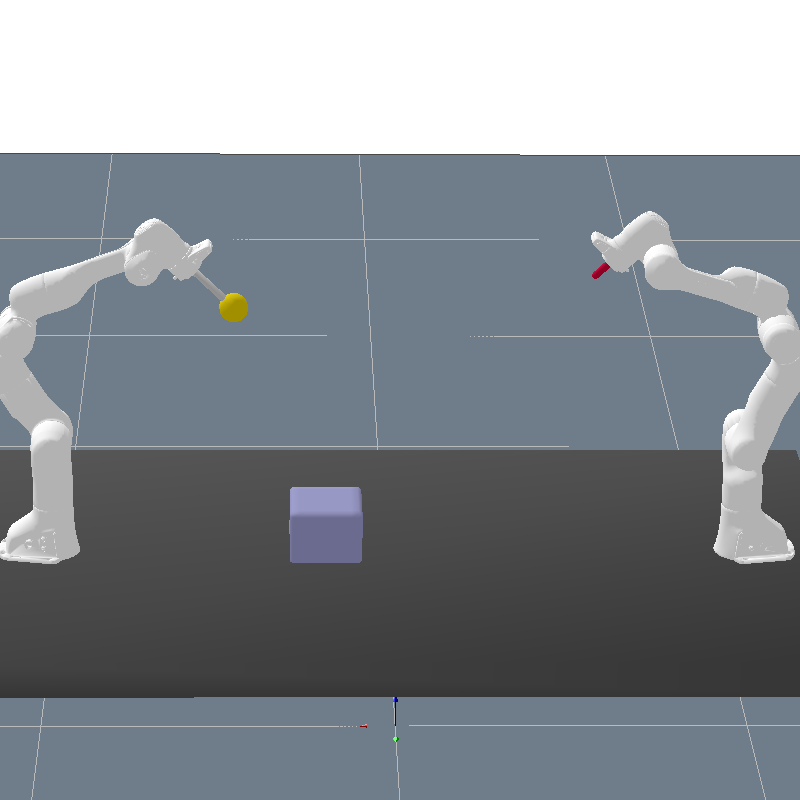}
	\includegraphics[trim={0cm 0cm 0cm 0cm}, clip, width=4cm]{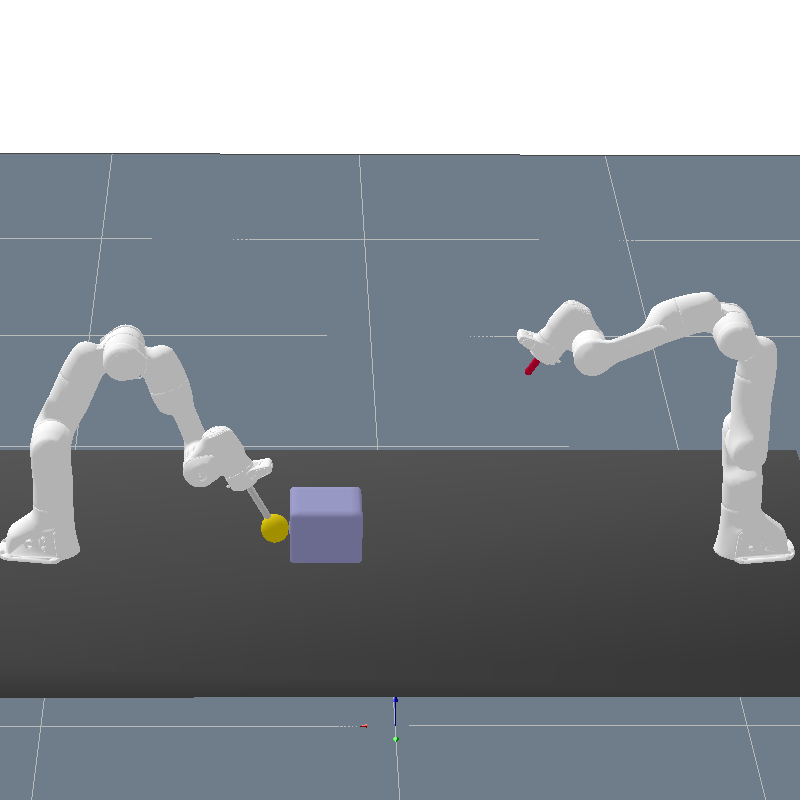}
	\includegraphics[trim={0cm 0cm 0cm 0cm}, clip, width=4cm]{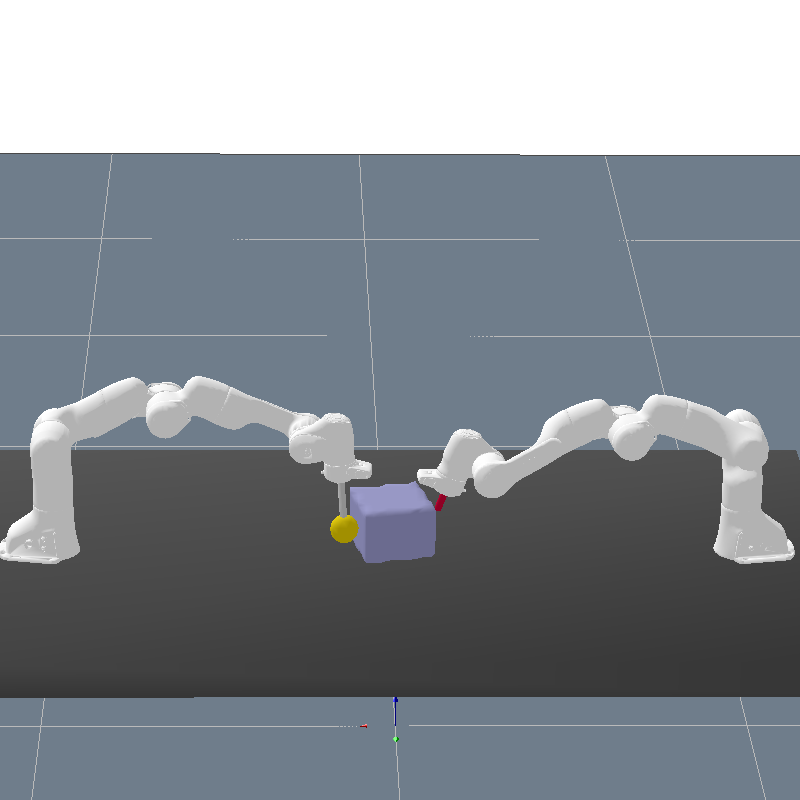}
	\caption{Scenario where the goal is that the right robot arm should touch the object, which is out of reach for the right arm. Therefore, our proposed framework plans a trajectory where the left arm pushes the object towards the right arm until it can reach it. Note how both arms move at the same time, since the optimization problem plans the robot/object motions jointly.}
	\label{fig:appendix:twoRobotsTouch}
\end{figure}

Goal regions are not the only way a goal can be specified.
In Fig.~\ref{fig:appendix:twoRobotsTouch}, a scenario is shown where the goal is that the right robot arm should touch the object.
Since the object is out of reach for the right robot arm, our framework plans a trajectory where the left robot arm pushes the object towards the right robot arm until it is able to touch the object.
Note that in this case there is no notion of a goal region, goal position or similar.
The fact that the object is moved in the direction of the other robot to the right is found by the optimizer trying to find a trajectory that is globally and jointly consistent with all constraints.
The goal constraint in this case is the contact establishment functional $H_\text{PoC}$ (see sec.~\ref{sec:TaskFuncs:PoC} and \eqref{eq:PoCOpt}) between the object and the end-effector of the right robot arm (both represented with SDFs).

Note that in none of these cases, we had to relearn a new model or change the methodology.
The optimization problem now optimizes over both the joint angles of the robots and the rigid transformations of the SDF representing the object.
Since the pusher object is attached to the end-effectors of the robots, the SDF of the pusher object is transformed in space via the movements of the robots.
The dynamic model constraint then makes the motions consistent with the learned physical model and the goal.

These experiments show the versatility and advantages of our problem formulation.

\section{Loss Function for Kinematic Success Model}\label{sec:appendix:lossKinematicSuccess}
To account for the fact that we want to use $H$ as a constraint, we use the loss function
\begin{align}
L(\theta) = \sum_{i=1}^{n} y^iH\!\left(\left(\phi^j\right)_{j\in\mathcal{I}_i}; \theta\right)^2 + (1-y^i)\exp\left(-H\!\left(\left(\phi^j\right)_{j\in\mathcal{I}_i}; \theta\right)\right)
\end{align}
to train the kinematic success models from sec.~\ref{sec:DSDFuncs:kinematic}.
When $y^i = 1$, this loss function brings the value of $H$ closer to zero, while for $y^i = 0$, the value of $H$ is being pushed up.

\section{Contact Establishment Functional}\label{sec:appendix:contactEstablishment}
The contact establishment functional $H_\text{PoC}$ defined in \eqref{eq:PoCOpt} in sec.~\ref{sec:TaskFuncs:PoC} has the property that, if the two objects are in contact, the global minimum of its optimization problem \eqref{eq:PoCOpt} is zero.
Therefore, to integrate $H_\text{PoC}$ into \eqref{eq:opt} without having nested optimizations, we can add $p\in\mathbb{R}^3$ as a decision variable to \eqref{eq:opt} and replace $H_\text{PoC}$ with the two constraints $\phi_1(p) = 0$ and $\phi_2(p) = 0$.
If these constraints are fulfilled, then a minimizer of \eqref{eq:PoCOpt} is found.
This formulation of $H_\text{PoC}$, as mentioned, does not only need no nested optimizations, but also provides informative gradients for the optimization problem, especially since the gradients of SDFs point towards their zero level set.

Note that \eqref{eq:PoCOpt} alone does not prevent objects from overlapping.
It only models that there exists a point where the distance to both objects is zero at the same time.

\section{Additional Experimental Details for Mug-Hanging Scenario}
\begin{figure}
	\centering
	\includegraphics[trim={8cm 9cm 7cm 6cm}, clip, height=3cm]{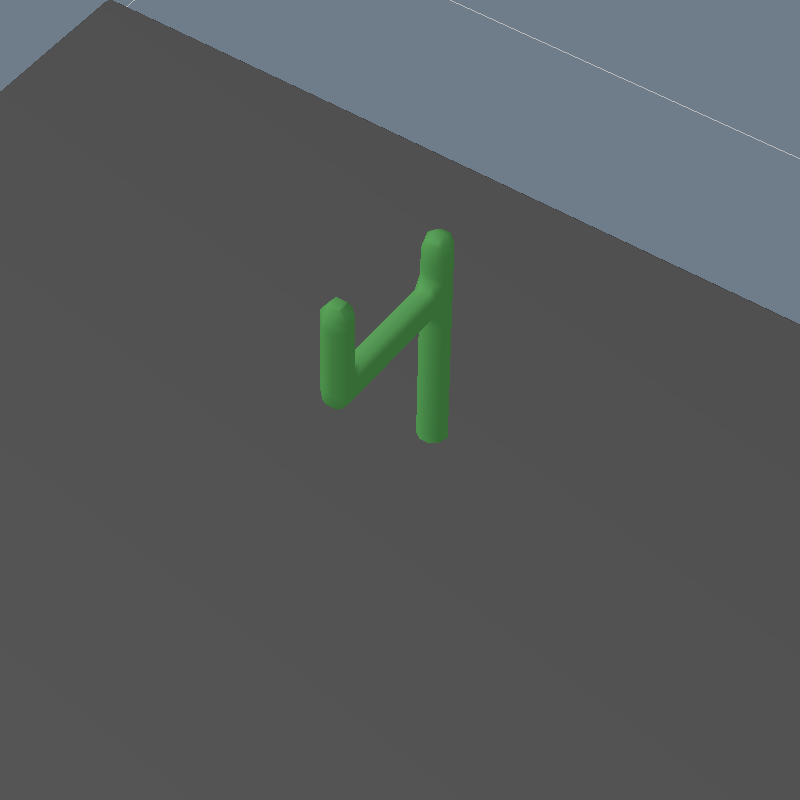}
	\includegraphics[trim={8cm 9cm 7cm 6cm}, clip, height=3cm]{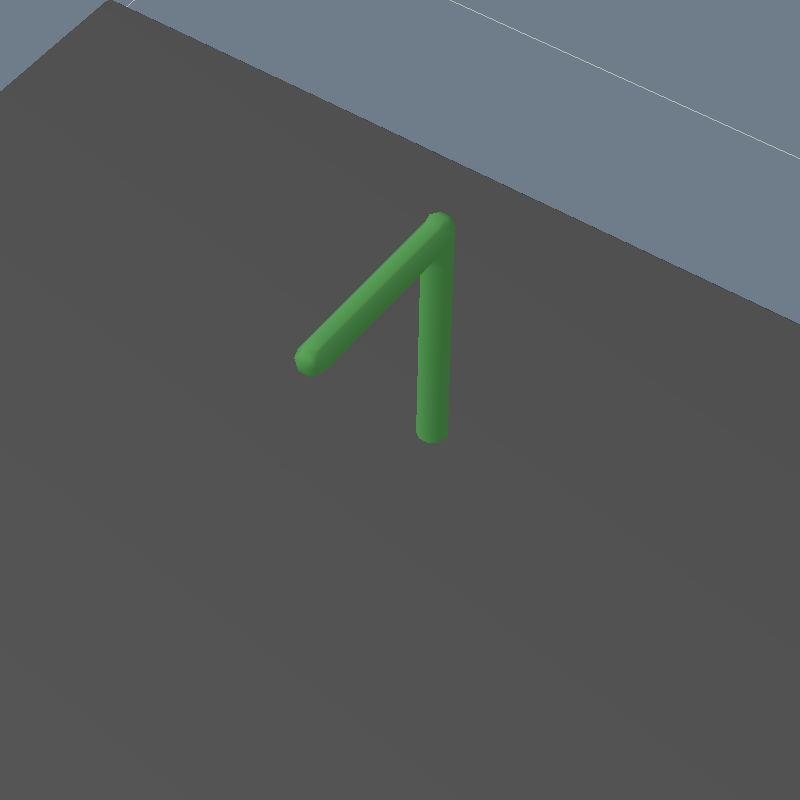}
	\includegraphics[trim={8cm 9cm 7cm 6cm}, clip, height=3cm]{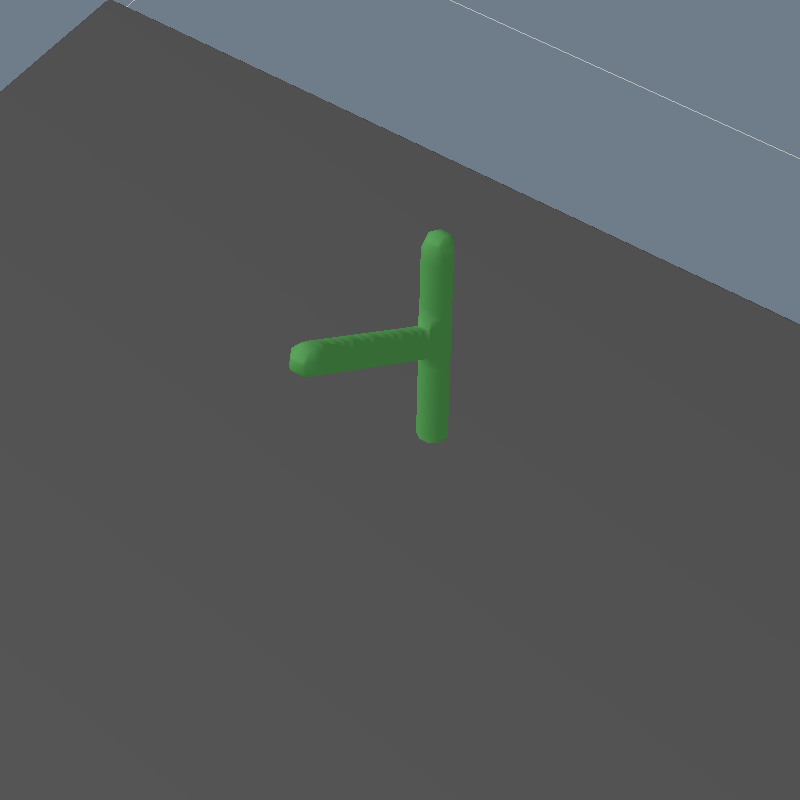}\\[0.2cm]
	\includegraphics[trim={8cm 9cm 7cm 6cm}, clip, height=3cm]{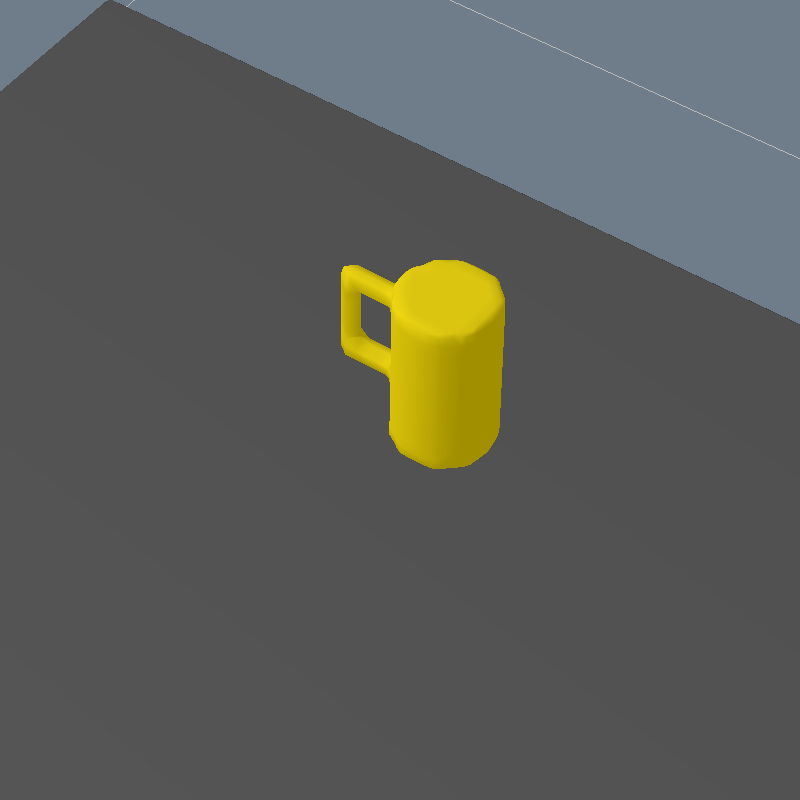}
	\includegraphics[trim={8cm 9cm 7cm 6cm}, clip, height=3cm]{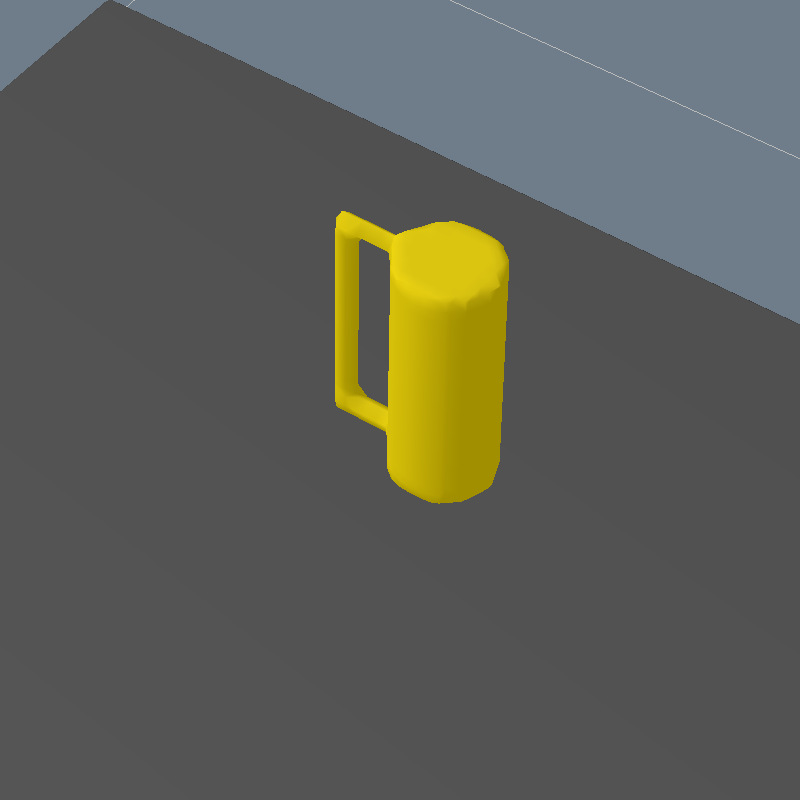}
	\includegraphics[trim={8cm 9cm 7cm 6cm}, clip, height=3cm]{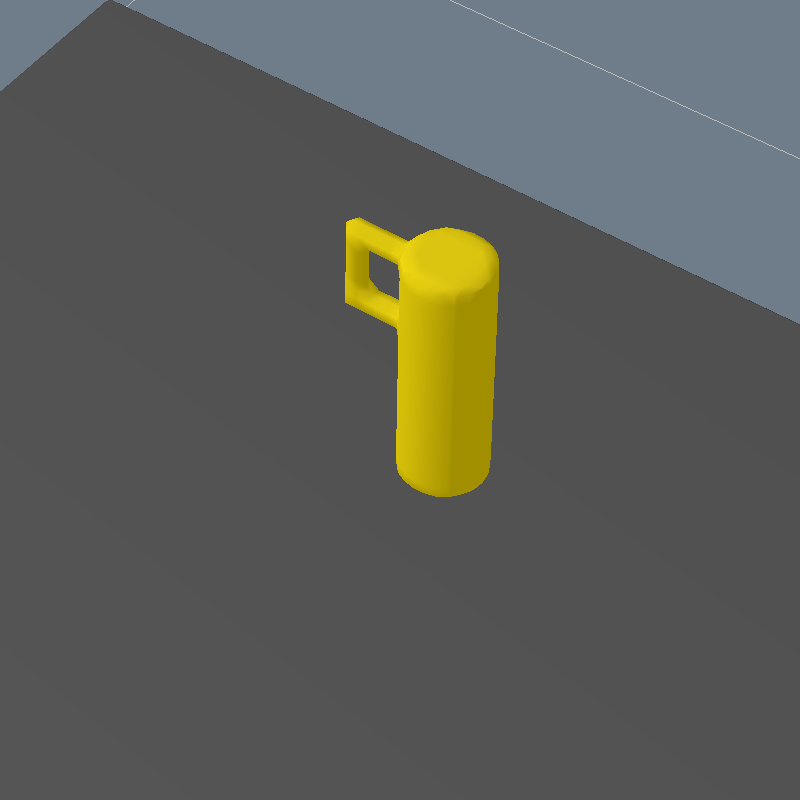}
	\caption{Different mug and hook shapes in evaluation dataset. SDFs meshed with marching cubes.}
	\label{fig:mugs}
\end{figure}

\begin{figure}
	\centering
	\includegraphics[trim={1cm 1cm 1cm 1cm}, clip, width=4.0cm]{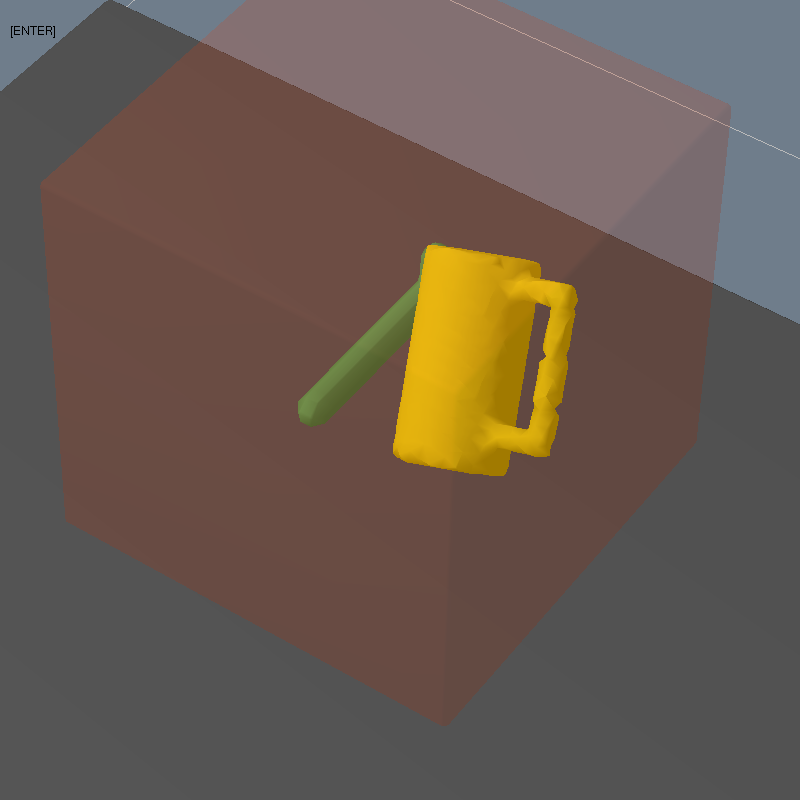}
	\includegraphics[trim={1cm 1cm 1cm 1cm}, clip, width=4.0cm]{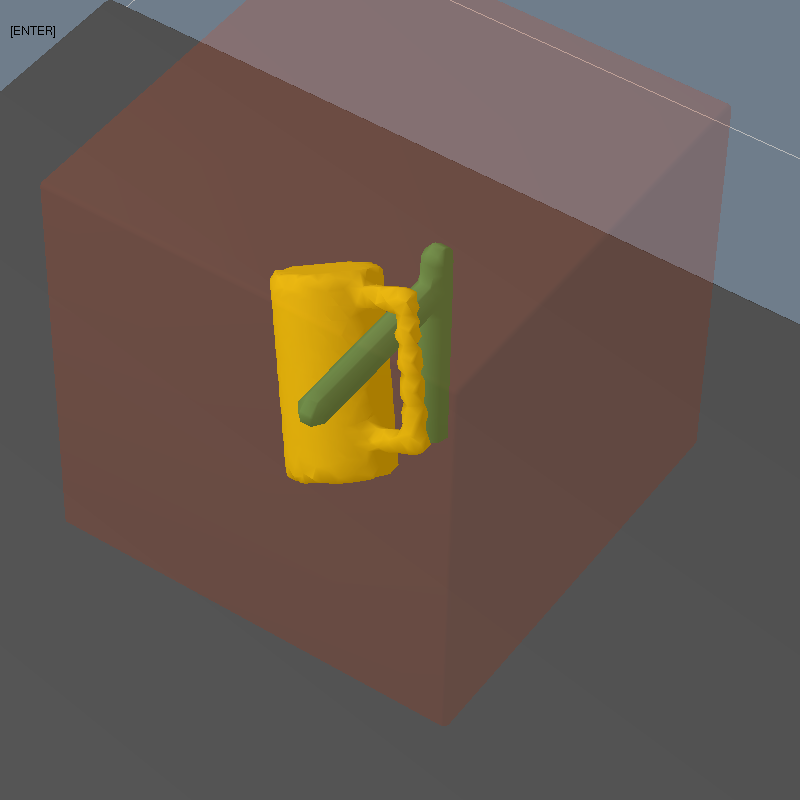}
	\includegraphics[trim={1cm 1cm 1cm 1cm}, clip, width=4.0cm]{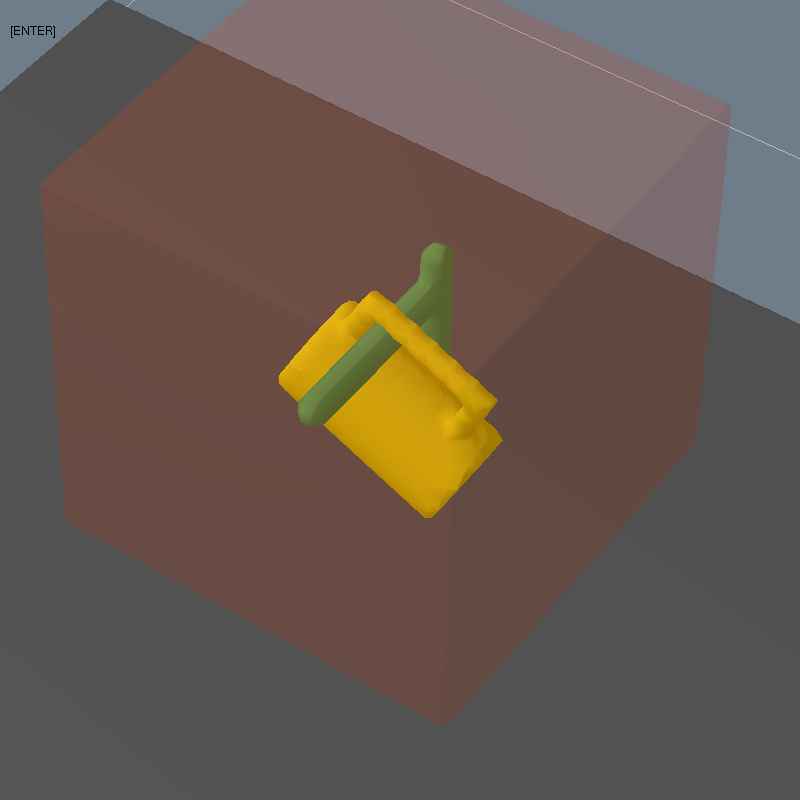}
	\caption{Left: sampled initial configuration from which the optimizer is started. Middle: found solution. Right: configuration after dropping the mug. Transparent red box is $\mathcal{X}_h$.}
	\label{fig:mugsEvolution}
\end{figure}

Fig.~\ref{fig:mugsEvolution} shows the set $\mathcal{X}_h$ (red box) for the functional $H_\text{hang}$. 
The hooks are centered in the $x,y$-plane of $\mathcal{X}_h$, not in $z$-direction (cf.\ also Fig.~\ref{fig:mugs}).
The bounding box $\mathcal{X}_h\in\mathbb{R}^{40\times 40 \times 40}$ in which the SDFs are queried has dimensions of 40 $\times$ 40 $\times$ 40 cm, i.e.\ the resolution is 1 cm.
Note that $\phi$ queried at the bounding box grid points contains the information about the distance to the object surface and hence not only whether there is an object or not at the grid point (compare to the occupancy measure representation).
Therefore, a 1 cm grid resolution turned out to be sufficient in our experiments.

As mentioned in sec.~\ref{sec:exp:mugHanging}, to generate the training data, we uniformly sample the position and orientation of the mugs inside $\mathcal{X}$.
We then run a forward simulation to check if the mug is stable when being dropped from the sampled configuration.
In case the mug has not fallen onto the ground, we apply an impulse to the mug (see supplementary video) to further check the stability.
The dataset $D = \left\{(\phi^1_i(\mathcal{X}_h), \phi^2_i(\mathcal{X}_h), y^i)\right\}_{i=1}^n$
then consists of the evaluated SDFs $\phi^2(\mathcal{X}_h)$ of the hook and $\phi^1(\mathcal{X}_h)$ of the mug in a configuration from which it either leads to a stable configuration ($y^i = 1$) or not ($y^i = 0$) when being dropped.
During training, the functional forms of $\phi^j$ are not needed, only their values on $\mathcal{X}_h$.
When using the model within the optimization problem, then the $\phi^j$ are utilized as functions.
The same impulse is also applied when evaluating the performance of the model.

The radius, height and three parameters of the handle (position relative to the height of the mug, extend in two directions) of the mugs are uniformly sampled.
The hook either consists of two or three parts, whose lengths and angle are sampled uniformly.
See the supplementary video for visualizations of these mugs and hooks in the evaluation dataset.
The training, test, and evaluation dataset utilize the same data generation technique, but all with a different random seed.

The parameter $a$ of \eqref{eq:pairCollision} for the collision constraint $H_\text{coll}$ is $a = 1000$.

\subsection{Network Architecture}\label{sec:appendix:hangingNetworkArchitecture}
The network architecture of $H_\text{hang}$ consists of three 3D convolutional layers, followed by an MLP of 3 dense layers.
The input SDFs (evaluated on $\mathcal{X}_h$) of both the mug and the hook are stacked into a $2\times 40 \times 40 \times 40$-dimensional input per training sample.
The convolutional layers have 3, 5, and 5 output channels with kernel sizes of 3, 5, and 5 and strides of 0, 2, and 2, respectively. The convolutions use ReLU activation functions.
After the convolutions, a fully connected linear layer creates a 200 dimensional feature vector.
This feature vector is followed by an MLP with three layers with hidden size of 300 each and ReLU activations.
The output is 1 dimensional.
The batch size is 32 with a learning rate of 0.0001 utilizing the ADAM optimizer.

\section{Additional Experimental Details for Pushing Scenario}\label{sec:appendix:additionalDetailsPushing}
During data generation, random movements of the pusher biased towards the object are applied.
Every 20th timestep, the SDFs of the object and the pusher at two consecutive timesteps are extracted at $\mathcal{X}_h$, i.e.\ the dataset $D = \left\{\left(\phi^1_t(\mathcal{X}_h), \phi^1_{t-1}(\mathcal{X}_h), \phi^2_{t}(\mathcal{X}_h), \phi^2_{t-1}(\mathcal{X}_h)\right)_i\right\}_{i=1}^n$ consists of these SDF evaluations only. This means that for training, no other information is needed, in particular no actions/velocities/relative transformations etc., just the SDF values queried on the grid points $\mathcal{X}_h$.
For planning with the learned model, the SDFs as functions are required, but not for training.

The bounding box is the 2D set $\mathcal{X}_h \in \mathbb{R}^{140\times 140}$ which covers the whole table.
The table has dimensions of 1.4 m $\times$ 1.4 m, leading to a resolution of 1cm (as with the mug hanging experiment).

The trajectory of rigid transformations in the optimization problem \eqref{eq:opt} is discretized in time by $T = 10$ steps per phase.
If there is a single push phase, then $K=2$ or $K=4$ for two push phases.

When the optimized trajectories are executed in the simulator open-loop, we interpolate linearly with 20 steps between two of the optimized consecutive rigid transformations of $\phi^2$.
This linear interpolation sometimes leads to collisions during the open-loop execution. For example, if two of the optimized pusher configurations are next to a corner of the object, the linear interpolation could collide with the corner, although the optimized configurations are not in collision.

The cost function $c$ penalizes accelerations of the part of $q$ that corresponds to the pusher $\phi^2$.
There are no such cost terms for the motion of $\phi^1$, which is only influenced by the dynamics model functional $H_F$.
In case $H_\text{PoC}$ is added to the optimization problem, there is also an acceleration regularization for $p$.
Since computing accelerations requires at least 3 consecutive timesteps in discrete time, $l = 2$, although the dynamics model $F$ is only quasi static ($l=1$).
The prefix is $q_{-2} = q_{-1} = q_0$.

The parameter $a$ in \eqref{eq:pairCollision} for the collision constraint functional $H_\text{coll}$ is $a = 500$ and $a = 100$ in \eqref{eq:goalRegion} for the goal region functional.

Note that the training data for the forward model $F$ is generated by a physical simulator.
This means a situation where $\phi^1_t$ and $\phi^2_t$ overlap can never be contained in the data.
Hence, there is no reason to believe that the model will predict anything useful in such a region \cite{19-driess-IROS}.
This is another advantage of adding the collision constraint functional $H_\text{coll}$ to the optimization problem, since it can prevent the optimization procedure to query the model in such regions where the model has seen no data.

The test and evaluation set for computing the prediction errors in Tab.~\ref{tab:pushingPredError} and \ref{tab:pushingPredErrorComparison} consists of 597 scenes.
The evaluation scenes for \ref{sec:exp:pushing:evalPlanning} and \ref{sec:exp:pushing:ablation} are yet another dataset.

For the evaluation scenes, we assume the object to be roughly (but not perfectly) in the middle of $\mathcal{X}$.
This has no particular reason. 
Since the functionals are translational invariant, one could also move the bounding box such that the object is roughly centered in it.
The bounding box, however, is never rotated.
Nevertheless, the training data was not biased at all to have objects centered in $\mathcal{X}$, hence also the prediction errors in Tab.~\ref{tab:pushingPredError} are for object positions/orientations in all $\mathcal{X}$.
The initial configuration of the pusher is sampled on a circle around the object.
The goal regions are sampled randomly of a set of 8 positions round the object with added Gaussian noise to their exact center positions. 

\subsection{Network Architecture of $F$ and $F_\text{flow}$}\label{sec:appendix:pushingNetworkArchitecture}
Each SDF $\phi^1_{t-1}$, $\phi^2_{t-1}$, $\phi^2_t$, after being evaluated on $\mathcal{X}_h\in\mathbb{R}^{140\times 140}$, is encoded into a 200 dimensional feature vector by the same encoder with shared weights.
This encoder consists of three 2D convolutional layers and a linear dense layer at the end to produce the 200 dimensional feature vector.
The convolutions have 3, 5, and 5 output channels with kernel sizes of 3, 5, and 5 and strides of 0, 2, and 2, respectively.
The input query point is encoded with one fully connected layer into a 100 dimensional feature vector.
The feature vectors of the 3 encoded SDFs and of the query point encoder are stacked into one 700-dimensional feature vector, which is then processed by an MLP with 2 hidden layers and 300 hidden units each.
The output is one dimensional.
All hidden units use ReLU activations.
The batch size is 32 with a learning rate of 0.0001 utilizing the ADAM optimizer.

\subsection{Initialization of Optimization Problem}\label{sec:appendix:pushingInitialization}
As mentioned in sec.~\ref{sec:exp:pushing:evalPlanning}, directly solving \eqref{eq:opt} on the pushing scenario often leads the optimizer to converge to an infeasible local minimum, since the optimization problem is highly non-convex.
This especially holds true in scenarios where the pusher has to go around the object to achieve the goal.
Such non-convexities of nonlinear trajectory optimization (especially with respect to collision avoidance) are not unique to our approach, but many planning methods that rely on trajectory optimization, even with fully analytical models.

\begin{figure}
	\centering
	\includegraphics[trim={0cm 0cm 0cm 1.5cm}, clip, width=5.0cm]{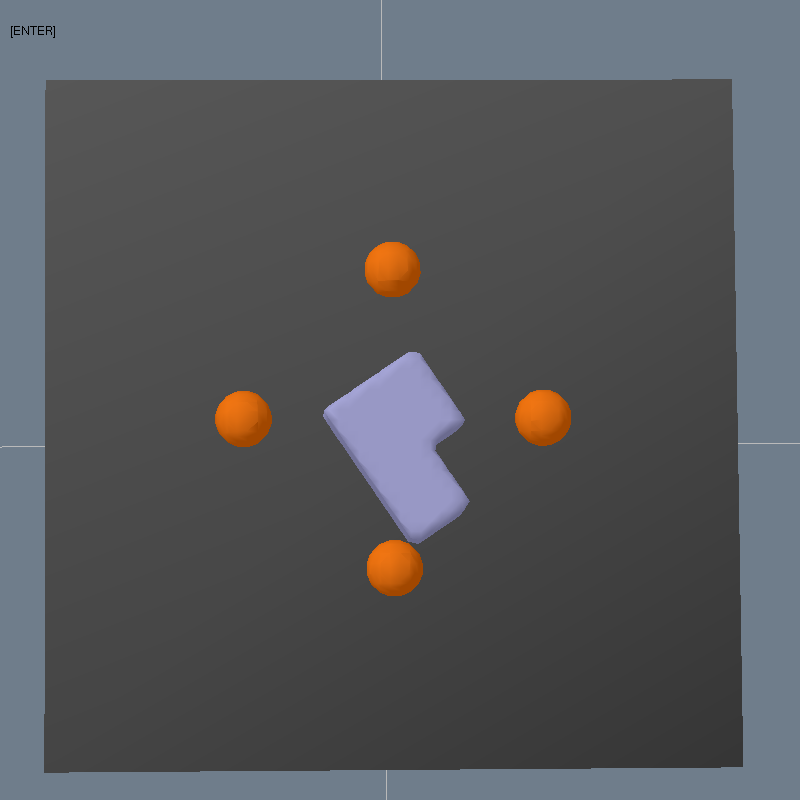}
	\caption{Visualization of the 4 initialization options (orange) for the trajectory optimization problem of the pusher around the object. These initialization options are always the same, independent from the shape/orientation of the object, the goal region position or the initial configuration of the pusher as given by the scene.}
	\label{fig:initializationPushing}
\end{figure}

To mitigate this issue, we initialize the pusher trajectory for the optimization problem, i.e.\ the rigid transformations for $\phi^2$, on four different positions around the object that is being pushed.
Fig.~\ref{fig:initializationPushing} shows these four initialization options in orange.
These initialization options are always the same, no matter of the shape, size, orientation of the object $\phi^1$.
The task planning part of \eqref{eq:opt} now not only decides on the number of pushing phases, but also the initialization of the optimization problem (since the initialization comes from a finite set).
This leads to 4 optimization problems with one push phase and 16 optimization problems with two push phases.
For each scene, we solve all those 20 optimization problems and then choose the one as the solution for the scene where the optimization problem converged best in terms of constraint violations and costs.
Note that we believe that these initialization options are a rather weak prior.
As can be seen in the supplementary video, even though sometimes the pusher has been initialized differently at two different phases, the optimizer sometimes chooses to let the pusher push the object two times on the same face.
Further, these initialization options also do not already establish contact with the object.
For the larger objects in the dataset, depending on the orientation of the object, those initialization options can also sometimes be in collision, which is taking care of during optimization. 
Between the initial given configuration of the pusher in the scene and the chosen initialization points as the initialization of the trajectory optimization problem, the trajectory of rigid transformations of the pusher is interpolated on a circle.
Note that the initial configuration of the pusher has to be distinguished by these initialization options. The first is given by the scene that should be solved, while the latter corresponds to the initial guess for the trajectory of the optimization problem.

\end{document}